\documentclass{article}

\usepackage{microtype}
\usepackage{graphicx}
\usepackage{booktabs} %

\usepackage{hyperref}

\usepackage[accepted]{icml2025}

\usepackage{multirow}
\usepackage{caption}
\usepackage{wrapfig,lipsum, subcaption, changepage, chngcntr}
\usepackage[linesnumbered,ruled,vlined]{algorithm2e}

\usepackage{booktabs, multirow} %
\usepackage{soul}%
\usepackage{changepage,threeparttable} %
\usepackage{arydshln}
\usepackage{amssymb}%
\usepackage{pifont}%

\usepackage{amsmath}
\usepackage{amssymb}
\usepackage{mathtools, enumitem}
\usepackage{amsthm}
\usepackage{xspace}
\usepackage{wrapfig}
\usepackage{capt-of}
\usepackage[export]{adjustbox}
\usepackage{arydshln, placeins, longtable}

\newcommand{\Var}{\mathrm{Var}}

\newcommand{\p}{\mathbb{P}}
\newcommand{\E}{\mathbb{E}}

\newcommand{\ours}{\textsc{LEMoN}\xspace}
\newcommand{\oursfixed}{\textsc{LEMoN\textsubscript{fix}}\xspace}
\newcommand{\oursopt}{\textsc{LEMoN\textsubscript{opt}}\xspace}

\usepackage[capitalize,noabbrev]{cleveref}

\theoremstyle{plain}
\newtheorem{theorem}{Theorem}[section]

\newtheorem{lemma}[theorem]{Lemma}

\theoremstyle{definition}

\theoremstyle{remark}

\usepackage[textsize=tiny]{todonotes}

\icmltitlerunning{\ours: Label Error Detection using Multimodal Neighbors}

\makeatletter
\def\adl@drawiv#1#2#3{%
        \hskip.5\tabcolsep
        \xleaders#3{#2.5\@tempdimb #1{1}#2.5\@tempdimb}%
                #2\z@ plus1fil minus1fil\relax
        \hskip.5\tabcolsep}
\newcommand{\cdashlinelr}[1]{%
  \noalign{\vskip\aboverulesep
           \global\let\@dashdrawstore\adl@draw
           \global\let\adl@draw\adl@drawiv}
  \cdashline{#1}
  \noalign{\global\let\adl@draw\@dashdrawstore
           \vskip\belowrulesep}}
\makeatother

\begin{document}

\twocolumn[
\icmltitle{\ours: Label Error Detection using Multimodal Neighbors}

\icmlsetsymbol{equal}{*}

\begin{icmlauthorlist}
\icmlauthor{Haoran Zhang}{equal,yyy}
\icmlauthor{Aparna Balagopalan}{equal,yyy}
\icmlauthor{Nassim Oufattole}{yyy}
\icmlauthor{Hyewon Jeong}{yyy}
\icmlauthor{Yan Wu}{yyy}
\icmlauthor{Jiacheng Zhu}{yyy}
\icmlauthor{Marzyeh Ghassemi}{yyy}
\end{icmlauthorlist}

\icmlaffiliation{yyy}{Massachusetts Institute of Technology}

\icmlcorrespondingauthor{Haoran Zhang}{haoranz@mit.edu}
\icmlcorrespondingauthor{Aparna Balagopalan}{aparnab@mit.edu}

\icmlkeywords{Machine Learning, ICML}

\vskip 0.3in
]

\printAffiliationsAndNotice{\icmlEqualContribution} %

\begin{abstract}
Large repositories of image-caption pairs are essential for the development of vision-language models. However, these datasets are often extracted from noisy data scraped from the web, and contain many mislabeled instances. In order to improve the reliability of downstream models, it is important to identify and filter images with incorrect captions. However, beyond filtering based on image-caption embedding similarity, no prior works have proposed other methods to filter noisy multimodal data, or concretely assessed the impact of noisy captioning data on downstream training. In this work, we propose, theoretically justify, and empirically validate \ours, a method to identify label errors in image-caption datasets. Our method leverages the multimodal neighborhood of image-caption pairs in the latent space of contrastively pretrained multimodal models to automatically identify label errors. Through empirical evaluations across eight datasets and twelve baselines, we find that \ours outperforms the baselines by over 3\% in label error detection, and that training on datasets filtered using our method improves downstream captioning performance by more than 2 BLEU points over noisy training.
\end{abstract}

\section{Introduction}
\label{sec:intro}

\begin{figure}[t!]
\centering
\includegraphics[width=1.\linewidth]{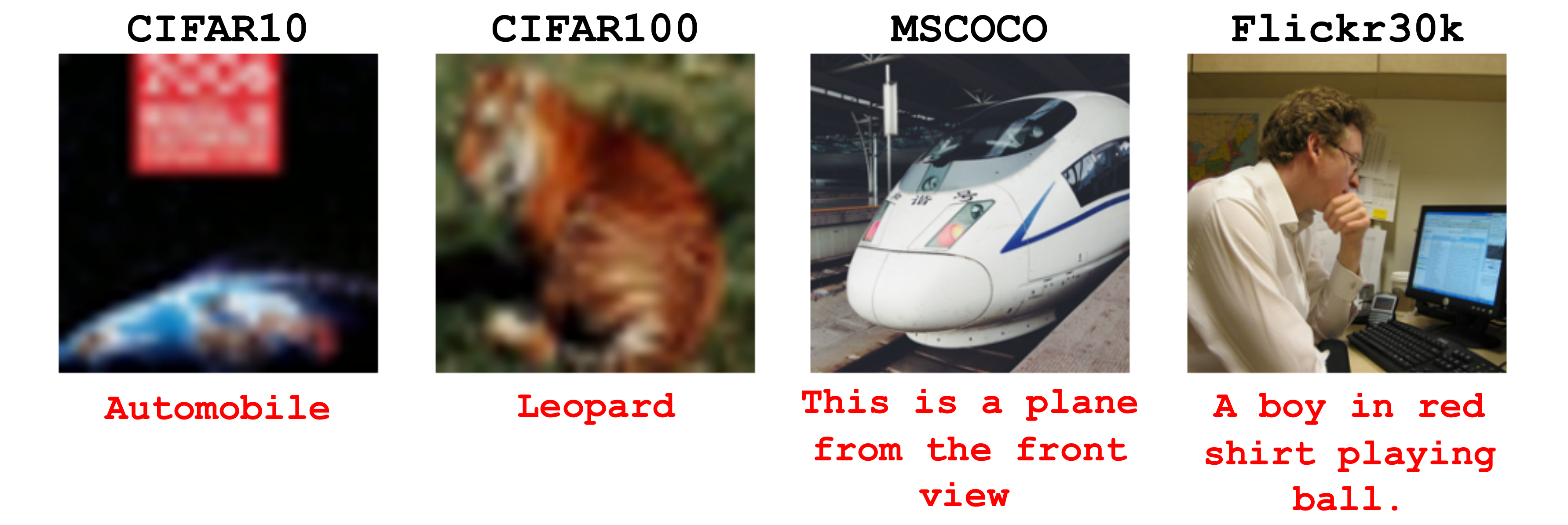}
\caption{Samples from classification and captioning datasets discovered to be mislabeled by our method.}
\label{fig:teaser}
\end{figure}

Machine learning datasets used to train and finetune large vision, language, and vision-language models frequently contain millions of labeled instances~\cite{schuhmann2021laion,li2022blip,wang2022git,changpinyo2021conceptual}. Prior work highlights that some instances in such datasets may be mislabeled~\cite{northcutt2021pervasive,luccioni2023bugs,liao2021we,beyer2020we,plummer2015flickr30k}, as seen in Figure \ref{fig:teaser}.  This is especially problematic in settings such as healthcare, where the reliability of downstream models may depend on the quality of data used for pretraining~\cite{chen2024cross,liu2023tinygsm,longpre2023pretrainer}.

Identifying and correcting label errors in existing datasets at scale would lead to more reliable and accurate models in the real world~\cite{zhu2022detecting, vasudevan2022does,liao2021we, beyer2020we}. However, given the large size of such datasets, manual detection of errors is practically infeasible. This is evidenced by the growth of models trained on noisy data with the web~\cite{li2022blip,wang2022git,liu2024visual}, or with model generated pseudo-labels~\cite{menghini2023enhancing,lai2023padclip}.

Machine learning (ML) based approaches to automatically identifying label errors have also been proposed in prior work~\cite{pleiss2020identifying,swayamdipta2020dataset,liang2023combating, bahri2020deep, zhu2022detecting, northcutt2021confident}. However, we identify two critical limitations: (1) the majority of such works are \emph{unimodal}: i.e., they only utilize image-based representations in detection strategies, and (2) many of the best-performing approaches depend on having access to a model already trained on the downstream tasks of interest~\cite{pleiss2020identifying,swayamdipta2020dataset}.
We hypothesize that applying a neighborhood-based approach to multimodal representations in the form of image-text pairs can improve label error detection without requiring  task-specific training, which may be costly and/or domain specific for some datasets.

Additionally, a common assumption made in prior works is that each label is one-of-k classes ~\cite{bahri2020deep,zhu2022detecting}. The vast majority of label error detection methods proposed in prior works are hence for \emph{classification} datasets. In contrast, datasets used to train large vision-language models contain natural language labels such as image captions~\cite{li2022blip,li2023blip,wang2022git}. Methods to filter out instances with noisy labels --- e.g., based on the similarity of image and caption representations --- have been utilized in prior work with some success~\cite{li2022blip,kang2023noise} for such datasets. However, to the best of our knowledge, no prior works have proposed or rigorously compared methods to identify errors in datasets with natural language labels, or assessed the impact of detection on downstream tasks like image captioning.

\begin{figure*}[t!]
  \label{fig:labelerror}
  \centering
  \begin{adjustwidth}{-0.in}{-0.in}
  \begin{subfigure}{0.5\textwidth}
  \includegraphics[width=1.0\linewidth]{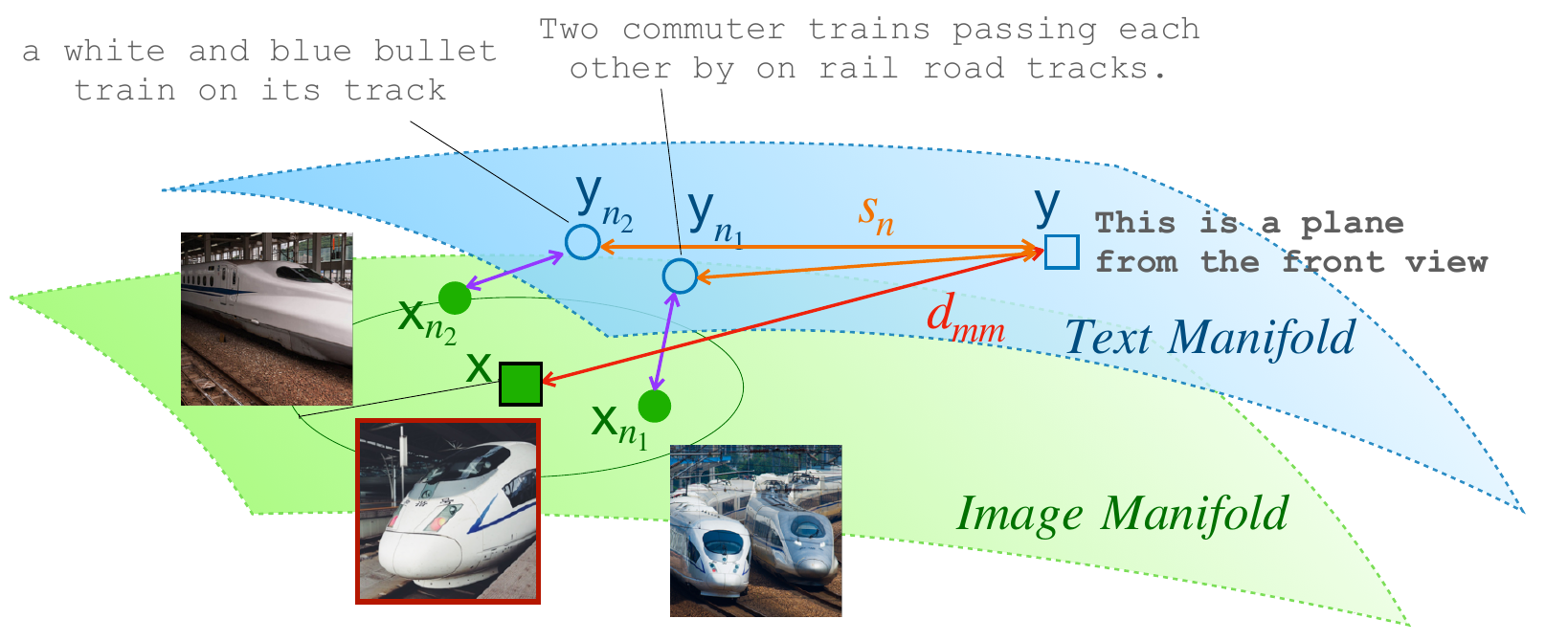}
    \caption{} \label{fig:method_a}
  \end{subfigure}%
  \hfill
   \begin{subfigure}{0.5\textwidth}
  \includegraphics[width=1.0\linewidth]{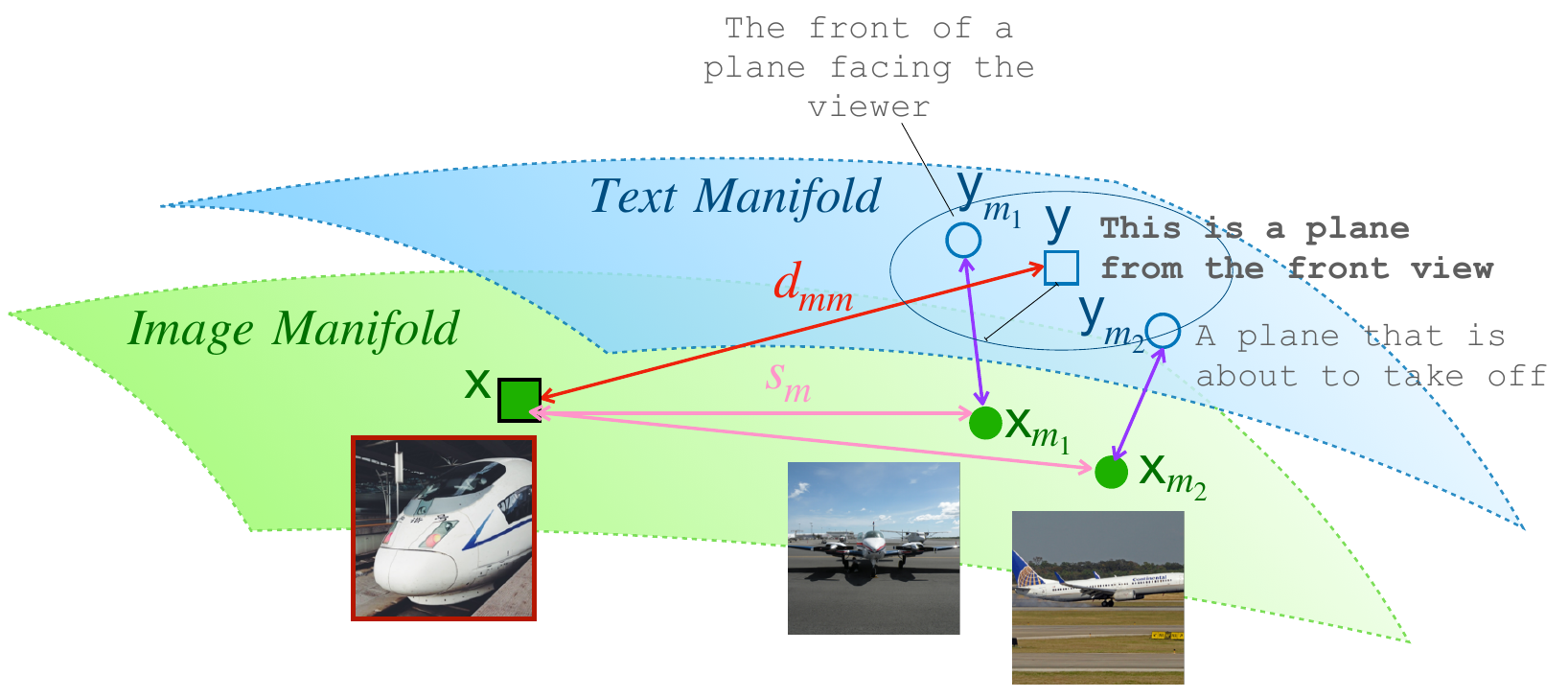}
    \caption{} \label{fig:method_b}
  \end{subfigure}%
  \end{adjustwidth}
  \caption{\textbf{Outline of \ours}, our proposed method for multimodal label error detection. We demonstrate \ours on a real sample from the MSCOCO dataset, where an image of a train ($\mathbf{x}$) is mislabeled as $\mathbf{y}$ = ``\texttt{This is a plane from the front view}''. (a) We compute the simple CLIP similarity $d_{mm}(\mathbf{x}, \mathbf{y})$. We then find the nearest neighbors of $\mathbf{x}$ in the image space ($\mathbf{x}_{n_j}$) and compute the distance between the corresponding texts and $\mathbf{y}$ to compute the score component $s_n$. (b) To compute the score component $s_m$, we find the nearest neighbors of $\mathbf{y}$ in the text space ($\mathbf{y}_{m_k}$), and compute the distance between the corresponding images and $\mathbf{x}$. }\label{fig:method}
\end{figure*}

In this work, we propose \ours\ -- \textbf{L}abel \textbf{E}rror detection using \textbf{M}ultim\textbf{o}dal \textbf{N}eighbors, a method for multimodal label error detection which can be applied to image-text pairs in datasets such as MSCOCO~\cite{lin2014microsoft}. While prior techniques utilize unimodal neighbors for label error detection, \ours leverages multi-modal neighborhoods derived using contrastively pretrained vision-language models such as Contrastive Language-Image Pretraining (CLIP)~\cite{radford2021learning}.
Specifically, in addition to considering pairwise image-text distances, we also retrieve nearest neighbors in the image and text space as illustrated in Figure~\ref{fig:method}. This is motivated by the rich neighborhood geometry in the joint embedding space of multimodal models~\cite{liang2022mind,schrodi2024two}.  We then compute distance scores with neighbors in each modality and combine these into a single score measuring the likelihood of a label error, with the intuition that higher discordance (or higher distance) with neighbors indicates a higher chance of label error. We validate the utility of these scores across eight datasets, including one in a healthcare setting, and compare to over ten baselines.

Our key contributions and findings are as follows\footnote{Code: \url{https://github.com/MLforHealth/LEMoN}}: 
\begin{itemize}
    \itemsep0em 
    \item We propose \ours, a novel, theoretically justified multimodal method capable of detecting label errors in large image-caption datasets (Section~\ref{sec:method}). 
    \item We show that \ours outperforms all training-free baselines for label error detection in three out of four classification datasets by up to 3.4\% AUROC, and in three out of four captioning datasets by up to 6.3\% AUROC (Section~\ref{sec:our_perf_clf}). %
    \item We demonstrate that \ours improves performance on downstream classification and captioning models by filtering out data predicted to be label errors. (Section~\ref{sec:filtering}).
    \item Finally, we verify that the predictions generated by \ours are meaningful through a real world analysis of \ours on existing datasets without known label errors  (Section~\ref{sec:real_world}).

\end{itemize}

\section{Related Works}
\paragraph{Label Noise Detection}

Noisy and incorrect labels~\cite{beyer2020we} in training data may lead to decreased or ``destabilized''~\cite{northcutt2021confident,luccioni2023bugs} performance on downstream tasks~\cite{chen2023understanding,northcutt2021pervasive}. Two orthogonal approaches can be taken to reduce the adverse effects of such labels: developing methods to learn in the presence of label errors~\cite{cui2020label,natarajan2013learning,huang2023nlip} referred to as ``noise robust" training, and/or detecting and filtering out instances with label errors~\cite{zhu2024vdc}. In this work, we focus on the latter direction. This approach may be preferable, as identifying mislabeled samples is more flexible, with applications beyond just removing these samples for downstream model training. For example, accurate label error detection can uncover systematic errors or biases in datasets~\cite{rottmann2023automated}, and these insights can then be used to guide higher-quality data collection practices~\cite{bernhardt2022active}. This is especially important for practitioners releasing datasets intended for model evaluation~\cite{northcutt2021pervasive, schubert2024identifying}.

Prior approaches~\cite{swayamdipta2020dataset,bahri2020deep,pleiss2020identifying,northcutt2021confident,liang2023combating, wu2020topological, kim2021fine} for automatic label error detection include relying on the training dynamics of task-specific downstream models~\cite{swayamdipta2020dataset} and neighborhood-based strategies~\cite{bahri2020deep,grivas2020not}. Some of these techniques are fully supervised \citep{northcutt2021confident, chen2023understanding} or unsupervised  \citep{pleiss2020identifying, swayamdipta2020dataset, grivas2020not, bahri2020deep}, use pre-trained generative models \cite{gertz2024potential} or are fully training-free approaches \citep{zhu2022detecting, liang2023combating}. Previous approaches for label error detection closest to this work includes deep k-nearest neighbor (deep k-NN) methods using k-NN entropy on vector space embeddings \cite{bahri2020deep, grivas2020not} and SimiFeat \citep{zhu2022detecting} which employs a local neighborhood-based voting or ranking for noise identification. In contrast to these methods, our work enhances label noise detection by harnessing information across \emph{multiple data modalities}, such as image and text.

\paragraph{Contrastive Learning}
Contrastive learning is a representation learning method, relying on positive and negative pairs of data instances \cite{chen2020simple, misra2020self, balestriero2023cookbook} to learn an embedding space. The core idea is to embed similar data points (positive pairs) closer together than dissimilar data points (negative pairs)~\cite{schroff2015facenet, sohn2016improved, oord2018representation}.  In this work, we primarily utilize pre-trained models that use the CLIP loss (where the pre-training objective is to predict which text caption is paired with which image) for jointly embedding image and text data~\cite{radford2021learning}.

\paragraph{Image Captioning}
The goal of image captioning is to describe a given image~\cite{fu2024noise} in natural language.
Prior approaches for caption generation have included supervised training of end-to-end models from scratch~\cite{wang2022end,
lin2022swinbert,
hu2023exploiting,xu2015show,fu2024noise}. More recently, vision-language models pretrained on large datasets of noisy image-caption pairs extracted from the web~\cite{li2022blip,li2023blip,wang2022git} -- such as CC12M~\cite{changpinyo2021conceptual} -- have been utilized for captioning. Some of the pretraining tasks include image-text contrastive learning, image-text matching, and/or retrieval~\cite{li2022blip}, as well as general purpose text generation conditioned on an input image~\cite{wang2022git}. Given that datasets for training such large models are noisy~\cite{kang2023noise}, several steps have been utilized in prior work to filter out noisy captions during training. The most common strategy involves computing the similarity between representations of the image and caption text using another pretrained model (e.g., CLIP) prior to training~\cite{kang2023noise}. Another approach in training the BLIP~\cite{li2022blip} model is to synthetically generate noisy captions and train a classifier to distinguish between high quality captions and noisy captions with a cross-entropy loss~\cite{li2022blip}. To the best of our knowledge, no previous work has conducted a comprehensive comparison of various strategies for label error detection in captioning datasets. %

\paragraph{Multimodal Neighborhood Methods} 
Previous studies~\cite{lisupervision, thomas2020preserving, huang2024neighbor, liang2022mind, cai2023semantic} have examined the geometry of neighborhood spaces in multimodal models, often with the goal of improving representation learning~\cite{huang2024neighbor, lisupervision} or retrieval~\cite{thomas2020preserving, thomas2022emphasizing}. The closest related work is~\citet{thomas2022emphasizing}, where the authors use the semantic neighborhood of multimodal models to identify samples with high semantic diversity using text-based neighbors of neighbors. Importantly, the objective of their work is different from ours, which leads their proposed discrepancy and diversity scores to lack signal for label error in our setting. We further clarify this in Appendix \ref{app:compare_with_baseline}, and empirically compare against their discrepancy score as a baseline. %
We believe our work is the first to use multimodal neighbors for label error detection. %

\section{\ours: Label Error Detection using Multimodal Neighbors}
\label{sec:method}

\label{sec:clip}

 We are given a dataset $\mathcal{D} = \{(\mathbf{x}, \mathbf{y})_{i=1}^N\}$ consisting of two modalities $\mathbf{x} \in \mathcal{X}$ and $\mathbf{y} \in \mathcal{Y}$. For example, $\mathcal{X}$ may represent the set of all natural images, and $\mathcal{Y}$ may represent the set of all English text, or a restricted subset such as $\{\texttt{cat}, \texttt{dog}, ...\}$. We assume the existence of, but not access to, an oracle $f^*: \mathcal{X} \times \mathcal{Y} \rightarrow \{0, 1\}$, which is able to assign a binary mislabel indicator $z_i = f^*(\mathbf{x}_i, \mathbf{y}_i)$ to each sample in $\mathcal{D}$. Here, $z_i = 1$ indicates that the sample is mislabeled, and $z_i = 0$ indicates that the sample is correctly labeled. Our goal is to output a score $s \in \mathbb{R}$ with some model $s := f(\mathbf{x}, \mathbf{y})$  
 such that: $$\text{AUROC} = \E_{\substack{(\mathbf{x}, \mathbf{y}) \sim \p(\cdot | z = 1) , (\mathbf{x}', \mathbf{y}') \sim \p(\cdot | z = 0)}}[\mathbf{1}_{f(\mathbf{x}, \mathbf{y}) \geq f(\mathbf{x}', \mathbf{y}')}]$$

is maximized. Prior works have alternatively aimed to maximize the F1 score, optimizing over a threshold $t$:
$$\text{F1} = \max_{t \in \mathbb{R}} \frac{2 \cdot \p(z = 1 | s \geq t )\cdot \p(s \geq t | z = 1  )}{\p(z = 1 | s \geq t ) + \p(s \geq t | z = 1  )}$$

Here, we build on prior work for label error detection in unimodal data~\cite{bahri2020deep, zhu2022detecting} and propose a method for $f$ based on nearest neighbors, summarized in Figure \ref{fig:method}. Suppose we have a query sample $(\mathbf{x}, \mathbf{y})$\footnote{One could take, for any $i$, $(\mathbf{x}, \mathbf{y}) := (\mathbf{x}, \mathbf{y})_i, D' := D \setminus \{(\mathbf{x}, \mathbf{y})_i\}$}. Define $B(\mathbf{x}, r) := \{x' \in \mathcal{X}: d_{\mathcal{X}}(\mathbf{x}, \mathbf{x}') \leq r \}$, the ball of radius $r$ around $\mathbf{x}$, and $B(\mathbf{y}, r)$ similarly. Let $r_k(\mathbf{x}) := \inf\{ r: | B(\mathbf{x}, r) \cap \mathcal{D}| \geq k \} $, the minimum radius required to encompass at least $k$ neighbors. Then, we define $\{\mathbf{x}_{n_1},\mathbf{x}_{n_2},...,\mathbf{x}_{n_k}\} := B(\mathbf{x}, r_k(\mathbf{x})) \cap \mathcal{D}$ the top $k$ nearest neighbors of $\mathbf{x}$, and $\{\mathbf{y}_{m_1},\mathbf{y}_{m_2},...,\mathbf{y}_{m_k}\} := B(\mathbf{y}, r_k(\mathbf{y})) \cap \mathcal{D}$ the top $k$ nearest neighbors of $\mathbf{y}$\footnote{We will use a subscript $n_j$ to index nearest neighbors in $\mathcal{X}$, and subscript $m_j$ for neighbors in $\mathcal{Y}$.}. We assume that the neighbors are sorted in order of ascending distance, e.g. $d_\mathcal{X}(\mathbf{x}, \mathbf{x}_{n_2}) \geq d_\mathcal{X}(\mathbf{x}, \mathbf{x}_{n_1})$. 

If $\mathcal{Y}$ is a small discrete set, we could choose $d(\mathbf{y}, \mathbf{y}') = \mathbf{1}_{\mathbf{y} = \mathbf{y}'}$. If $\mathcal{X}$ or $\mathcal{Y}$ are unstructured or high dimensional, we assume access to multimodal encoders $h_\theta = (h^\mathcal{X}_\theta, h^\mathcal{Y}_\theta)$, where $h^\mathcal{X}_\theta : \mathcal{X} \rightarrow \mathbb{R}^d$ and $h^\mathcal{Y}_\theta : \mathcal{Y} \rightarrow \mathbb{R}^d$. Here, $h_{\theta}$ may be a CLIP model~\cite{radford2021learning} trained on a large internet corpus, or, as we show later, it may be sufficient to train $h_{\theta}$ from scratch only on $\mathcal{D}$. Then, we could naturally use simple distance metrics in the embedding space, such as the cosine distance $d_\mathcal{X}(\mathbf{x}, \mathbf{x'}) =  d_{\text{cos}}(h^{\mathcal{X}}_{\theta}(\mathbf{x}), h^{\mathcal{X}}_{\theta}(\mathbf{x}')) = 1 - \frac{h^{\mathcal{X}}_{\theta} (\mathbf{x})^T h^{\mathcal{X}}_{\theta} (\mathbf{x}')}{||h^{\mathcal{X}}_{\theta} (\mathbf{x})||_2 \cdot ||h^{\mathcal{X}}_{\theta} (\mathbf{x}')||_2}$. Our proposed score is the linear combination of three terms:
\begin{equation}
    s = f(\mathbf{x}, \mathbf{y}) = d_{mm}(\mathbf{x}, \mathbf{y}) + \beta s_n (\mathbf{x}, \mathbf{y}, \mathcal{D}) + \gamma s_m(\mathbf{x}, \mathbf{y}, \mathcal{D}),
    \label{eq:our_score}
\end{equation}
where $\beta , \gamma \geq 0$ are hyperparameters. 
Here, $d_{mm} (\mathbf{x}, \mathbf{y}) := d_{\text{cos}}(h_{\theta}^{\mathcal{X}}(\mathbf{x}), h_{\theta}^{\mathcal{Y}}(\mathbf{y}))$ is the multimodal distance, which has been shown empirically to provide a meaningful signal in prior label error detection work \cite{liang2023combating, kang2023noise}. We thus use this distance as the basis, and augment it with two additional terms based on nearest neighbors:
\vspace{-2mm}
\begin{equation} \label{eq:sn}
\resizebox{0.5\textwidth}{!}{
    $s_n(\mathbf{x}, \mathbf{y}, \mathcal{D}) = \frac{1}{k} \displaystyle\sum_{j=1}^k d_{\mathcal{Y}} (\mathbf{y}, \mathbf{y}_{n_j}) e^{-\tau_{1, n}  d_{\mathcal{X}} (\mathbf{x}, \mathbf{x}_{n_j}) } e^{-\tau_{2, n} d_{mm} (\mathbf{x}_{n_j},  \mathbf{y}_{n_j}) }$,
    }
\end{equation}
where $(\mathbf{x}_{n_j}, \mathbf{y}_{n_j}) \in \mathcal{D}$, and $\tau_{1,n}, \tau_{2,n} \geq 0$ are hyperparameters. This corresponds to finding the nearest neighbors of $\mathbf{x}$ in $\mathcal{X}$ space, then averaging the distance between their \textit{corresponding} modality in $\mathcal{Y}$ and $\mathbf{y}$. We weight this average with two additional terms. The $\tau_{1, n}$ term corresponds to downweighting neighbors which are far from $\mathbf{x}$. Intuitively, this is useful when $k$ is too large for $\mathbf{x}$ and not all neighbors are relevant, and can be thought of as an adaptive $k$. The $\tau_{2, n}$ term corresponds to downweighting neighbors which are themselves likely to be mislabeled. If $(\mathbf{x}_{n_j}, \mathbf{y}_{n_j})$ is itself mislabeled, then $d_{\mathcal{Y}}(\mathbf{y},  \mathbf{y}_{n_j})$ would contribute an erroneous signal to whether $(\mathbf{x}, \mathbf{y})$ is mislabeled, and we thus want to downweight those instances.

The third term is analogous to $s_n$, but uses neighbors of $\mathbf{y}$:
\vspace{-2mm}
\begin{equation} \label{eq:sm}
\resizebox{0.5\textwidth}{!}{
    $s_m(\mathbf{x}, \mathbf{y}, \mathcal{D}) = \frac{1}{k} \displaystyle\sum_{j=1}^k d_{\mathcal{X}} (\mathbf{x}, \mathbf{x}_{m_j}) e^{-\tau_{1, m}  d_{\mathcal{Y}} (\mathbf{y}, \mathbf{y}_{m_j}) } e^{-\tau_{2, m} d_{mm} (\mathbf{x}_{m_j},  \mathbf{y}_{m_j}) }$
    ,}
\end{equation}
where $(\mathbf{x}_{m_j}, \mathbf{y}_{m_j}) \in \mathcal{D}$, and $\tau_{1,m}, \tau_{2,m} \geq 0$ are hyperparameters. Crucially, note that notationally, $\mathbf{x}_{n_j} \neq \mathbf{x}_{m_j}$, and $\mathbf{y}_{n_j} \neq \mathbf{y}_{m_j}$. Specifically, $\mathbf{y}_{n_j}$ corresponds to the $\mathcal{Y}$ modality of nearest neighbors taken in $\mathcal{X}$ space, and $\mathbf{y}_{m_j}$ corresponds to the nearest neighbors of $\mathbf{y}$ taken in $\mathcal{Y}$ space.

We note that our method is a generalization of several prior methods. When $\beta = \gamma = 0$, the method is equivalent to CLIP similarity~\cite{liang2023combating}. When $\beta$ is large, $\tau_{1, n} = \tau_{2, n}  = \gamma = 0$, and $d(\mathbf{y}, \mathbf{y}_{n_j}) = \mathbf{1}_{\mathbf{y} = \mathbf{y}_{n_j}}$, the method is equivalent to Deep kNN~\cite{bahri2020deep}. An algorithm outline and high-level description of the method can be found in Appendix \ref{app:algo}.

Our method contains several hyperparameters: $k, \beta, \gamma, \tau_{1, n}, \tau_{2, n}, \tau_{1. m}$, and $\tau_{2, m}$. When there is a validation set with known mislabel flags, we perform a grid search over $k$, and use numerical optimization methods to search for an optimal value of the remaining hyperparameters which maximize label error detection performance on this set, which we describe further in Section \ref{sec:selection}. We refer to our method in this setting as \oursopt. We will empirically show that only a few hundred labeled validation samples may be sufficient to achieve optimal performance in this setting. 

When there is no labeled validation set available, we will show that our method is fairly robust to these hyperparameter choices, and that choosing a set of reasonable fixed values for these hyperparameters yields nearly comparable results. We refer to our method in this setting as \oursfixed.

\section{Theoretical Analysis}
\label{sec:theoretical_analysis}

We show that our multimodal kNN scores (Equations (\ref{eq:sn}) and (\ref{eq:sm})) provide a signal for label error. Suppose there exists a ``paraphrase function'' $\mathcal{H} : \mathcal{Y} \rightarrow \mathcal{P}(\mathcal{Y})$, where $\mathcal{P}$ denotes the powerset, such that for a particular sample $(x, y)$ with $\mathcal{H}(y) = (\bar{y}_1, \bar{y}_2 ..., )$,  $(x, \bar{y}_i)$ is considered correctly labeled for all $\bar{y}_i \in \mathcal{H}(y)$. Informally, $\mathcal{H}$ outputs the set of all possible captions which correctly describe $x$. Similarly define $\mathcal{J}(x)$, which outputs the set of images with identical semantics as $x$. 

\noindent\textbf{Assumption 1} (Structure of $\mathcal{H}$, $\mathcal{J}$): 
\begin{itemize}
    \item Let $(x', y')$ be an arbitrary sample. If $y' \not\in \mathcal{H}(y)$, then $x' \not\in \mathcal{J}(x)$.
    \item Let $(x', y')$ be an arbitrary mislabeled sample. Then, $\forall y'' \in \mathcal{H}(y')$, $x'' \not\in \mathcal{J}(x')$.
\end{itemize}
\noindent\textbf{Assumption 2} (Distribution of Distances): \textit{Let $(X, Y)$ be a randomly drawn sample.}
\begin{itemize}[noitemsep,topsep=0pt]
    \item $\forall\ X' \not\in \mathcal{J}(X): d_{\mathcal{X}}(X, X') \mathop{\sim}\limits^{\mathrm{iid}} \mathcal{N}(\mu_1, \sigma_1^2)$  .
    \item $\forall\ \bar{X} \in \mathcal{J}(X): d_{\mathcal{X}}(X, \bar{X}) \mathop{\sim}\limits^{\mathrm{iid}} \mathcal{N}(\mu_2, \sigma_2^2)$ .
\end{itemize}

We empirically validate this assumption in Appendix \ref{app:normality_test}. 

Let $N_k(Y) = \{Y_{m_1}, ... , Y_{m_k}\}$ denote the nearest neighbors of $Y$ in the text space. Let $\frac{1}{k}|\mathcal{H}(Y) \cap N_k(Y)| = \zeta_Y$, a random variable. 
Suppose that $\frac{1}{k}|\{i: (X_{m_i}, Y_{m_i}) \text{ is mislabeled}\}| = p$ is constant for all samples in the support of $(X, Y)$.

Let $S_m(X, Y) = \frac{1}{k} \sum_{Y_{m_i} \in N_k(Y)} d_{\mathcal{X}}(X, X_{m_i}) $, which is identical to the proposed Equation (\ref{eq:sm}) with $\tau_1 = \tau_2 = 0$. 
\begin{theorem}[AUROC of kNN Score] Let $(X, Y)$ be a randomly selected correctly labeled sample,  and $(X', Y')$ a randomly selected incorrectly labeled sample. Under Assumptions 1 and 2: 
\begin{align*}
    \mathbb{P}( S_m(X', Y')  > S_m(X, Y)) &= 1 - \Phi(\frac{-\mu}{\sigma})
\end{align*}
where $\mu = \E[\zeta_{Y}] (1-p) (\mu_1 - \mu_2 ), \sigma =$ \resizebox{0.5\textwidth}{!}{$\left(\frac{ \E[\zeta_Y] (1-p) \sigma_2^2 + (2 - \E[\zeta_Y] (1-p) ) \sigma_1^2}{k} + \Var(\zeta_Y) (1-p)^2 (\mu_2 - \mu_1)^2 \right)^{1/2}$}, and $\Phi$ is the Gaussian CDF. 
\end{theorem}

\noindent This provides an expression for the detection AUROC of the score $S_m$. The same expression can be derived for $S_n$ by symmetry.

\begin{lemma}[Non-random Signal of kNN Score] \label{lemma:bad} If the following three conditions hold: (1) $p < 1$, (2) $\E[\zeta_{Y}] > 0$, (3) $\mu_1 > \mu_2$. Then, $\mathbb{P}(S_m(X', Y') > S_m(X, Y) ) > 0.5$.
\end{lemma}

\noindent Under these mild conditions, $S_m$, our proposed multimodal neighborhood score, provides a better than random signal at detecting mislabeled samples. The proof can be found in Appendix \ref{app:thm_proof}. We additionally provide a theorem showing that embedding models trained via the contrastive multimodal objective are natural noisy label detectors in Appendix \ref{app:thm2}.

\section{Experiments}
\label{sec:exps}
\subsection{Datasets}
\label{sec:datasets}
We evaluate our method using eight datasets, as shown in Table \ref{tab:datasets}. Four datasets (\texttt{cifar10},  \texttt{cifar100}, \texttt{stanfordCars}, \texttt{miniImageNet}) are label error detection datasets from the classification setting. 
The four remaining datasets are image captioning datasets. For \texttt{mscoco} and \texttt{flickr30k}, we use the Karpathy split~\cite{karpathy2015deep}. The remaining datasets were randomly split into: training or reference set for the label detection method (80\%), validation set for hyperparameter selection (10\%), and test set for performance evaluation (10\%). 

\begin{table*}[!htp]\centering

\caption{Classification and captioning datasets. $n$ is the number of samples. In the main paper, results shown are for the bolded noise type with 40\% noise level for synthetic noise. Results for remaining settings can be found in the appendices. }
\resizebox{0.95\textwidth}{!}{  
\begin{tabular}{lrrrcccc}\toprule
\textbf{Dataset} &\multicolumn{3}{c}{\textbf{$n$}} &\multicolumn{2}{c}{\textbf{Domain}} &\textbf{Noise Types} \\ \cmidrule(l){2-4} \cmidrule(l){5-6}
&\textbf{Train} &\textbf{Validation} &\textbf{Test} &\textbf{Image} &\textbf{Text} & \\\midrule
\texttt{cifar10} &40,000 &5,000 &5,000 &Natural images&Object labels &\{\textbf{\textit{human}} ~\cite{wei2021learning}, \textit{sym., asym}.\} \\
\texttt{cifar100} &40,000 &5,000 &5,000 &Natural images &Object labels &\{\textbf{\textit{human}}~\cite{wei2021learning},\textit{ sym}., \textit{asym}.\} \\
\texttt{miniImageNet}~\citep{jiang2020beyond} & 49,419 & 24,710 & 24,710 & Natural images & Object labels & \{\textbf{\textit{real}}\} \\
\texttt{stanfordCars}~\citep{jiang2020beyond} & 13,501 & 6,751 & 6,752 & Car images & Car year and model & \{\textbf{\textit{real}}\} \\
\texttt{mscoco}~\cite{lin2014microsoft} &82,783 &5,000 &5,000 &Natural images &Captions &\{\textbf{\textit{cat.}}, \textit{noun}, \textit{ random}\} \\
\texttt{flickr30k}~\cite{young2014image} &29,000 &1,014 &1,000 &Natural images &Captions &\{\textbf{\textit{noun}}, \textit{random}\} \\
\texttt{mmimdb}~\cite{arevalo2017gated} &15,552 &2,608 &7,799 &Movie Posters &Plot summaries &\{\textbf{\textit{cat.}}, \textit{noun},  \textit{random}\} \\
\texttt{mimiccxr}~\cite{johnson2019mimic} & 368,909& 2,991& 5,159&Chest X-rays &Radiology reports &\{\textbf{\textit{cat.}},\textit{ random}\} \\
\bottomrule
\end{tabular}
}
\label{tab:datasets}
\end{table*}

\subsubsection{Noise Types}

 In \texttt{cifar10} and \texttt{cifar100}, we utilize a dataset collected in prior work \cite{wei2021learning} with human mislabels (\textit{human}). We also follow prior work~\cite{zhu2022detecting} in experimenting with class symmetric (\textit{sym.}) and class asymmetric (\textit{asym.}) synthetic noise. For \texttt{stanfordCars} and \texttt{miniimagenet}, we use datasets from ~\citet{jiang2020beyond},
which contain noise from real-world (\textit{real}) web annotators .
 
 For the four captioning datasets, we devise several ways to inject synthetic noise of prevalence $p$. The simplest way is to randomly select $p$ fraction (\textit{random}) of the samples and assign their text modality to be that of another random caption. In datasets where additional metadata is available (\texttt{mscoco}: object category, \texttt{mmimdb}: genre of movie, \texttt{mimiccxr}: disease label), we can randomly swap the caption with that of another sample from the same category (\textit{cat}). Finally, in all captioning datasets except \texttt{mimiccxr}, we tag each token of each caption with its part-of-speech using SpaCy~\cite{spacy2}, and then randomly assign a selected sample's text modality to be from another sample with at least one noun in common (\textit{noun}). Dataset processing details are in Appendix \ref{sec:data_processing}.

These noise types are intended to simulate an array of realistic label corruptions in the real world. As such, the resulting synthetic dataset may not have an exact noise level $p$, as e.g. a randomly selected caption may actually be correct for the image, as well as due to noise in the base datasets, which we explore in Section \ref{sec:real_world}. Unless otherwise stated, results shown in the main paper are for the bolded noise type in Table \ref{tab:datasets}, with 40\% synthetic noise. Additional results for other noise types can be found in Appendix \ref{app:add_results}. 

\subsection{Model Selection and Evaluation}
\label{sec:selection}
We run \ours on each dataset, using the training split of each dataset to compute nearest neighbors. In classification datasets, we use the discrete metric $d_{\mathcal{Y}}(\mathbf{y}, \mathbf{y'}) = \mathbf{1}_{\mathbf{y} = \mathbf{y'}} $. In all other cases and for $d_{\mathcal{X}}$, we utilize cosine or euclidean distance computed in the embedding space of a pretrained CLIP model, selecting the best distance metric on the validation set for \oursopt, and keeping the distance as the cosine distance for \oursfixed. In \texttt{mimiccxr}, we use BiomedCLIP (ViT-B/16) \cite{zhang2023biomedclip}, and we use OpenAI CLIP ViT-B/32 \cite{radford2021learning} for all other datasets. A full list of hyperparameters for our method and the baselines are in Appendix~\ref{sec:hparams}.

For \oursopt, we select the hyperparameter combination that maximizes F1 on a labeled validation set. We report the AUROC, macro-averaged AUPRC, and F1 for this model. For \oursfixed, we fix the hyperparameters at the following reasonable values: $k = 30$, $\beta = \gamma = 5$, $\tau_{1, n} = \tau_{1, m} = 0.1$, and $\tau_{2, n} = \tau_{2, m} = 5$. We report AUROC and AUPRC, as the F1 requires additional information to compute a threshold for the score.
We recognize that access to such a validation set as in \oursopt may be unrealistic, but we will empirically show that (1) our method is fairly robust to selection of these hyperparameters, (2) only a few hundred labeled samples may be sufficient to select these hyperparameters, (3) using \oursfixed with the fixed  hyperparameter setting described above achieves nearly comparable results, and (4) hyperparameters optimized on a dataset with synthetic noise may transfer well to real datasets. 

We repeat each experiment three times, using a different random seed for the noise sampling (for \textit{human} and \textit{real} noise, we use a different random data split). Performance metrics shown are test-set results averaged over these three runs, with error bounds corresponding to one standard deviation.

\paragraph{Baselines}
\label{sec:baselines}

We compare our method versus previous state-of-the-art in both the classification and captioning settings. We additionally adapt several baselines from the classification setting to the captioning setting. We briefly list the baselines here, and a detailed description is in Appendix \ref{sec:baseline_methods_app}. 

In the classification setting, we experiment with the following baselines which require training a classifier on the particular dataset:  \textbf{AUM}~\cite{pleiss2020identifying}, \textbf{Datamap}~\cite{swayamdipta2020dataset}, and \textbf{Confident Learning} \cite{northcutt2021confident}, and the following baselines which do not require classifier training: \textbf{Deep k-NN}~\cite{bahri2020deep}, \textbf{SimiFeat} \cite{zhu2022detecting}-Voting and Ranking, discrepancy in the image space (\textbf{Discrepancy}) ($\Upsilon^{DIS}_X$ from \citet{thomas2022emphasizing}), \textbf{CLIP Similarity}~\cite{kang2023noise}, and \textbf{CLIP Logits}~\cite{liang2023combating,feng2024clipcleaner}.

\begin{table*}[tb]
\centering
\caption{Label error detection performance across classification datasets. We separate AUM, Datamap, and Confident learning, as they require training a classifier from scratch. Bold denotes best score within each training approach. A full version of this table with AUPRC can be found in Appendix \ref{app:clf_label_error_detection.}.}
\resizebox{0.9\textwidth}{!}{  
\begin{tabular}{lcrrrrrrrr}\toprule
\textbf{Method} & \textbf{Training-Free} &\multicolumn{2}{c}{\texttt{cifar10}} &\multicolumn{2}{c}{\texttt{cifar100}} &\multicolumn{2}{c}{\texttt{miniImageNet}} &\multicolumn{2}{c}{\texttt{stanfordCars}} \\
\cmidrule(lr){3-4}\cmidrule(lr){5-6}\cmidrule(lr){7-8}\cmidrule(lr){9-10}
& & \textbf{AUROC} & \textbf{F1} & \textbf{AUROC} & \textbf{F1} & \textbf{AUROC} & \textbf{F1} & \textbf{AUROC} & \textbf{F1} \\
\midrule
AUM & \multirow{3}{*}{\ding{55}}
    & \textbf{98.3} (0.1) & \textbf{92.9} (0.1)
    & \textbf{92.3} (0.3) & \textbf{81.1} (0.3)
    & 83.1 (0.2) & 68.3 (0.4)
    & 70.4 (2.3) & 47.2 (3.1) \\
Datamap &
    & 98.2 (0.1) & 92.2 (0.5)
    & 91.8 (0.3) & 80.8 (0.5)
    & \textbf{85.1} (0.3) & \textbf{70.6} (0.2)
    & \textbf{72.2} (1.7) & \textbf{50.4} (2.1) \\
Confident &
    & 89.6 (1.4) & 88.2 (1.7)
    & 78.6 (0.4) & 73.7 (0.5)
    & 59.5 (0.7) & 37.7 (1.5)
    & 60.7 (0.3) & 39.9 (0.6) \\
\cdashlinelr{1-10}
CLIP Logits & \multirow{8}{*}{\ding{51}}
    & 95.5 (0.2) & 86.2 (0.6)
    & 84.9 (0.7) & 72.0 (0.9)
    & \textbf{90.0} (0.2) & \textbf{77.1} (0.2)
    & 68.8 (0.1) & 47.3 (0.5) \\
CLIP Sim.\ &
    & 92.2 (0.2) & 82.3 (0.3)
    & 80.8 (0.9) & 68.7 (1.1)
    & 89.3 (0.2) & 76.1 (0.4)
    & 69.8 (0.4) & 46.6 (0.5) \\
Simifeat-V &
    & 90.9 (0.1) & 88.4 (0.5)
    & 79.3 (0.4) & 72.8 (0.7)
    & 68.2 (0.3) & 55.1 (0.6)
    & 63.4 (1.3) & 43.3 (1.5) \\
Simifeat-R &
    & 90.7 (0.2) & 88.2 (0.3)
    & 79.6 (0.2) & 73.3 (0.3)
    & 68.1 (0.2) & 54.8 (0.5)
    & 63.6 (1.2) & 43.5 (1.6) \\
Discrepancy &
    & 77.1 (1.9) & 66.4 (2.2)
    & 66.0 (1.5) & 58.9 (0.8)
    & 79.5 (0.2) & 64.0 (0.1)
    & 65.7 (0.3) & 44.3 (0.7) \\
Deep k-NN &
    & 97.8 (0.1) & 91.4 (0.6)
    & 87.4 (0.3) & 75.7 (0.3)
    & 83.2 (0.2) & 68.5 (0.2)
    & 71.5 (0.6) & 49.1 (0.6) \\
\oursfixed (Ours) &
    & 97.7 (0.2) & 90.9 (0.1)
    & 88.9 (0.7) & 75.4 (0.6)
    & 89.5 (0.2) & 74.7 (0.2)
    & 72.6 (0.7) & 47.7 (2.0) \\
\oursopt (Ours) &
    & \textbf{98.1} (0.0) & \textbf{92.0} (0.2)
    & \textbf{90.8} (0.0) & \textbf{78.4} (0.0)
    & \textbf{90.0} (0.4) & 76.9 (0.2)
    & \textbf{73.1} (0.5) & \textbf{51.3} (0.5) \\
\bottomrule
\end{tabular}

}
\label{tab:clf_label_error_detection}
\end{table*}

In the captioning setting, we compare our method with  \textbf{LLaVA}~\cite{liu2024visual} prompting and \textbf{CapFilt} \cite{li2022blip}. We note that the latter can be viewed as an oracle for natural image captioning, as it has been trained in a supervised manner on \textit{clean} \texttt{mscoco} data.
\textbf{CLIP Similarity}~\cite{kang2023noise}, \textbf{Discrepancy}~\cite{thomas2022emphasizing},  and \textbf{Datamap}~\cite{swayamdipta2020dataset} can also be used directly in this setting. Next, to adapt classification baselines to captioning, we embed the captions using the corresponding CLIP text encoder, and then use K-means clustering to assign the text caption into one of 100 clusters. We then apply \textbf{Deep k-NN}~\cite{bahri2020deep} and \textbf{Confident Learning} \cite{northcutt2021confident}, using the cluster ID as the discretized class. Finally, we adapt \textbf{VDC} \cite{zhu2024vdc} to the captioning setting using open source models. Further details can be found in Appendix \ref{sec:baseline_methods_app}.

\section{Results}

\begin{table*}[!htbp]\centering
\caption{Label error detection performance on captioning datasets. We separate CapFilt as it has been trained on \textit{clean} \texttt{mscoco} data. Bold denotes best (highest) score. A full version of this table with AUPRC can be found in Appendix \ref{sec:captioning_synth}.}
\resizebox{0.88\textwidth}{!}{%
\begin{tabular}{lrrrrrrrr}\toprule
\textbf{Method} &\multicolumn{2}{c}{\texttt{flickr30k}} &\multicolumn{2}{c}{\texttt{mscoco}} &\multicolumn{2}{c}{\texttt{mmimdb}} &\multicolumn{2}{c}{\texttt{mimiccxr}} \\
\cmidrule(lr){2-3}\cmidrule(lr){4-5}\cmidrule(lr){6-7}\cmidrule(lr){8-9}
 & \textbf{AUROC} & \textbf{F1} & \textbf{AUROC} & \textbf{F1} & \textbf{AUROC} & \textbf{F1} & \textbf{AUROC} & \textbf{F1} \\
\midrule
LLaVA                     & 79.3 (0.8) & 65.0 (1.1) & 80.3 (0.1) & 74.9 (0.3) & 58.4 (0.2) & 58.5 (0.1) & 53.9 (0.5) & 57.0 (0.1) \\
Datamap                   & 52.7 (1.5) & 50.4 (1.8) & 68.9 (0.8) & 60.3 (1.2) & 54.0 (0.3) & 57.2 (0.1) & 50.2 (1.2) & 57.0 (0.1) \\
Discrepancy               & 73.0 (0.6) & 59.0 (0.3) & 72.7 (0.3) & 62.5 (0.3) & 57.8 (0.4) & 57.4 (0.2) & 60.0 (0.7) & 57.2 (0.1) \\
VDC                       & 92.9 (1.1) & 81.1 (1.6) & 94.1 (0.2) & 86.3 (0.4) & 80.5 (0.3) & 69.3 (0.6) & 50.8 (0.4) & 57.0 (0.1) \\
Deep k-NN                 & 71.1 (0.4) & 59.2 (0.8) & 76.6 (0.4) & 67.5 (0.8) & 61.2 (0.4) & 58.3 (0.4) & 62.9 (0.4) & 59.2 (0.1) \\
Confident                 & 63.1 (0.9) & 54.0 (0.9) & 71.5 (0.5) & 66.5 (0.5) & 52.8 (1.1) & 51.8 (1.8) & 61.8 (0.3) & 58.1 (0.6) \\
CLIP Sim.                 & \textbf{94.8} (0.5) & \textbf{84.2} (0.9) & 93.8 (0.2) & 84.5 (0.4) & 85.1 (0.3) & 72.7 (0.6) & 64.1 (0.4) & 59.2 (0.0) \\
\oursfixed (Ours)         & 93.6 (0.2) & -- & 92.0 (0.1) & -- & 84.3 (0.3) & -- & 66.3 (0.4) & -- \\
\oursopt (Ours)           & 94.5 (0.2) & 83.6 (1.4) & \textbf{95.6} (0.2) & \textbf{87.0} (0.2) & \textbf{86.0} (0.1) & \textbf{73.5} (0.3) & \textbf{70.4} (1.6) & \textbf{61.1} (0.8) \\
\cdashlinelr{1-9}
CapFilt (Supervised Training) & 98.6 (0.1) & 93.1 (0.7) & 99.3 (0.0) & 95.4 (0.4) & 82.7 (0.7) & 71.3 (0.3) & 49.7 (0.3) & 57.0 (0.0) \\
\bottomrule
\end{tabular}}
\label{tab:captioning}
\end{table*}

\subsection{\ours Outperforms Baselines on Label Error Detection}
\label{sec:our_perf}
\paragraph{Classification}
\label{sec:our_perf_clf}
In Table \ref{tab:clf_label_error_detection}, we show the performance of \ours against the baselines for label error detection on four classification datasets. We find that our method performs on-par with, or outperforms, existing baselines which do not require classifier training on all classification datasets. Two downstream-task specific approaches (AUM and Datamap) outperform most training-free models (particularly on \texttt{cifar100}), but \ours performs comparably and even outperforms them in two datasets. Similar results are also observed on the two synthetic error types (see Appendix Table~\ref{tab:label_error_synth_clf}). We find that \oursfixed performs almost comparably with \oursopt, and still beats almost all baselines.

\paragraph{Captioning}
\label{sec:our_perf_capt}

In Table~\ref{tab:captioning}, we find that our method outperforms existing neighborhood and similarity-based baselines on three datasets. In two datasets, our model underperforms the oracle  (CapFilt) which has been pre-trained on \textit{clean} captions from the \texttt{mscoco} dataset. %
Results for synthetic error types show similar trends (see Appendix~\ref{sec:captioning_synth}).

\paragraph{Label Error Detection Performance Across Noise Levels}  In Figure \ref{fig:vary_noise}, we show the performance of \ours versus the CLIP similarity baseline on \texttt{mscoco} and \texttt{mmimdb}, varying the level of the synthetic noise. We find that \ours performs better uniformly across noise levels.

\paragraph{Size of Labeled Validation Set} In Appendix Figure \ref{fig:vary_val_size}, we examine how varying the size of the labeled validation set impacts the performance of \oursopt. We find that in all four captioning datasets, having about 100-500 labeled examples is sufficient to tune hyperparameters in \oursopt to outperform \oursfixed. In the three datasets where \oursfixed underperforms the CLIP similarity baseline, we find again that having 100-500 labeled validation samples is sufficient for tuning \oursopt to perform on par with this baseline.

\paragraph{Robustness to Hyperparameters} Here, we test the robustness of our method when there is no labeled validation set available. First, in Appendix \ref{sec:robustness_to_hparams}, we visualize the F1 of the selected score when varying $\beta$ and $\gamma$, keeping all other hyperparameters at their selected optimal values. We find that for most datasets and noise types, there is a reasonably large space of such hyperparameters, bounded away from the origin, which achieves close to optimal performance. 

Next, we compare the performance of \oursopt and \oursfixed with hyperparameters described in Section \ref{sec:selection} across all datasets in Table~\ref{tab:no_hparams_still_good}. We find that when there is no labeled validation set available, using these hyperparameters results in an AUROC drop of only 1.6\% on average (std = 1.3\%), with a worst-case AUROC drop of 4.1\% across all 18 dataset and noise type combinations. Thus, even when a labeled validation set is not available, \oursfixed with reasonable hyperparameter settings is able to outperform most baselines which do use such information.

\subsection{Filtering Mislabeled Data Improves Downstream Performance} 
\label{sec:filtering}

\paragraph{Classification}

\begin{figure}[htbp!]
  \centering
  \includegraphics[width=\linewidth,trim={0.2 0.5 0.2 0.7}]{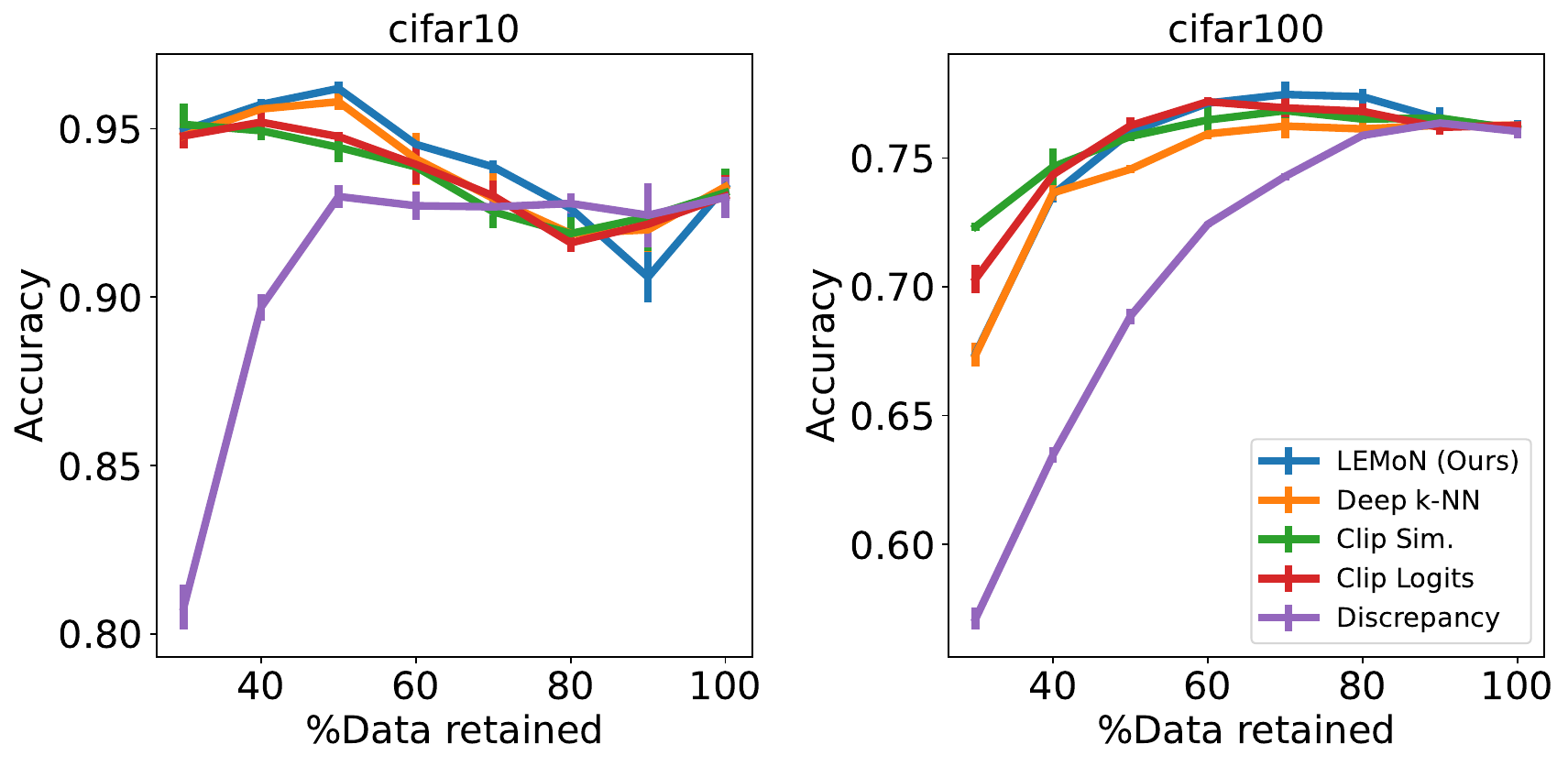}
  \caption{Downstream classification accuracy on \texttt{cifar10} (left) and \texttt{cifar100} (right) with \oursopt with \textit{human} noise versus the baselines. Note that the noise prevalence is 40\% in both datasets.  \label{fig:downstream_class_threshold} }
\end{figure}

To assess the impact of label error detection on the performance of the downstream classification tasks, we filter out samples from the training set with mislabel scores in the top $q$ percentile. We vary $q$, train ViT~\cite{dosovitskiy2020image} models on the filtered dataset, and evaluate the downstream test accuracy using clean data. We compare the performance of \oursopt with all training-free baselines that produce a continuous score (i.e. all except Simifeat and Confident). In Figure \ref{fig:downstream_class_threshold}, we find that training with \oursopt filtered samples leads to the highest accuracy on \texttt{cifar10} and \texttt{cifar100} after removing more than 30\% of the data. Training with \oursopt filtered samples is also on par with baselines on the other datasets (within 0.5\% points of best baseline) as shown in Appendix \ref{app:downstream_clf_acc}. Further, unlike other baselines, LEMoN is consistently in the top-2 best performing methods across all four datasets in terms of downstream accuracy. We also show that filtering data in this manner does not reduce classifier robustness (Appendix \ref{ref:ood_robust}).

\paragraph{Captioning}

\begin{table}[t!]\centering
\caption{Downstream captioning performance when removing 40\% samples with highest mislabel scores. We find that filtering noisy data with \oursopt improves captioning.}
\begin{adjustbox}{width = 0.85\linewidth}
\begin{tabular}{lcrrr}\toprule
\textbf{Dataset}  & \textbf{Method} & \textbf{B@4} &\textbf{CIDER} &\textbf{ROUGE}\\\midrule

\multirow{4}{*}{\texttt{flickr30k}} & No Filtering &27.9 (0.7) &	64.7 (0.3)&	49.3 (0.3)	\\
 & CLIP Sim.& \textbf{29.6 (0.5)} &	71.5 (1.3) &	50.6 (0.3)	\\
&\oursopt & 29.5 (0.6)	& \textbf{72.5 (0.8)} & \textbf{51.0 (0.4)}\\ \cdashlinelr{2-5}
  &Clean& 31.0 (0.4)	& 75.0 (0.8)	& 51.9	(0.0)\\

\midrule
\multirow{4}{*}{\texttt{mscoco}} & No Filtering &35.0 (0.1)	& 116.7 (0.4) &	56.3 (0.2)	 \\
 & CLIP Sim.&37.8 (0.1) &	126.5 (0.4)	& 58.3 (0.1)	\\

&\oursopt&\textbf{38.0 (0.2)}	& \textbf{126.7 (0.5)} &	\textbf{58.4 (0.1)}	\\

\cdashlinelr{2-5}
  &Clean&37.9 (0.3) &	127.0 (0.6) &	58.4 (0.1)	\\
\bottomrule
\end{tabular}
\end{adjustbox}
\label{tab:captioning_downstream_results}

\end{table}

 We finetune a pre-trained GenerativeImage2Text (GIT)~\cite{wang2022git} model to generate captions. Given the large size of the model, we use the parameter-efficient Low-Rank Adaptation (LoRA)~\cite{hu2021lora} for all captioning models. We train models with clean data, noisy captions (\textit{No Filtering}), and by filtering data detected as being mislabeled by a label detection method. In Table~\ref{tab:captioning_downstream_results}, we compare results of using either our model or a strong baseline (CLIP Sim.) for filtering data, as measured by the BLEU-4~\cite{papineni2002bleu}, CIDER~\cite{vedantam2015cider}, and ROUGE~\cite{lin2004rouge} scores. In all cases, we filtered out the top 40\% percentile of data predicted to be mislabeled (i.e., equal to the expected prevalence of noisy data). We find that (1) filtering out data predicted to be mislabeled helps recover performance as compared to training on fully clean data along multiple metrics, and (2) our method performs comparably to the baseline in improving downstream results, with some improvements over CLIP Similarity on \texttt{mscoco} and \texttt{flickr30k} (for CIDER and ROUGE, and for BLEU-4 scores on \texttt{mscoco}). However, as training with a fully clean dataset only outperforms training with a fully noisy (i.e. no filtering) dataset by 2-3 BLEU-4 points, the range of potential improvement for any filtering method is limited.

\subsection{Ablations} In Table \ref{tab:ablation}, we show the performance of our method after ablating each component. We find that mislabel detection performance almost decreases monotonically as we remove additional components until we reach the CLIP Similarity baseline. We find that ablating the $\tau_1$ and $\tau_2$ terms results in a performance loss of about $1\%$. In Table \ref{tab:ablation_rebuttal}, we examine the performance of each of the three components of our score and their combinations. We find that $d_{mm}$ is the most critical term. Of the two nearest neighbors terms, we find that $s_n$ (nearest image neighbors) is more important in general, though this is highly dataset dependent, e.g. error detection in \texttt{mmimdb} relies much more on neighbors in the text space than the image space, while the opposite is true for \texttt{mscoco}.

\subsection{External Pretraining May Not Be Required}
\label{sec:no_pretraining}

\paragraph{Medical Images} Thus far, all of the results for \ours (and CLIP Similarity) have utilized CLIP models which have been pretrained on external datasets (e.g. PMC-15M in the case of BiomedCLIP). Here, we examine whether we can achieve comparable performance by pretraining CLIP from scratch \textit{only on the noisy data}. We select \texttt{mimiccxr} as it has the most samples out of all captioning datasets. Similar to CheXzero \cite{tiu2022expert}, we pretrain a CLIP ViT B/16 from scratch on the \texttt{mimiccxr} training set with 40\% noise. We train this model for 10 epochs with a batch size of 64, and do not do any model selection or early stopping. We then apply \ours and the CLIP similarity baseline using this model, for the same noise level and noise type. We present our results in Table \ref{tab:clip_from_scratch}. Surprisingly, we find that pretraining CLIP only on noisy data from MIMIC-CXR actually outperforms BiomedCLIP. This could be attributed to the pretraining domain (chest X-rays and radiology notes) matching the inference domain exactly~\cite{nguyen2022quality}. As an upper bound, we evaluate the same methods using CheXzero \cite{tiu2022expert}, which has been pretrained on \textit{clean} MIMIC-CXR data. We find that, as expected, it far outperforms this baseline. We conclude that, for large noisy datasets, pretraining a CLIP model from scratch could be a viable solution, though pretraining on clean data from the same domain is certainly superior.

\begin{table}[htbp!]
\caption{{Performance of \ours for label error detection versus the CLIP similarity baseline on \texttt{mimiccxr}, when external pretrained models may not be available. BiomedCLIP \cite{zhang2023large} is trained on a large corpus of biomedical image-text pairs. We find that pretraining only on noisy data from MIMIC-CXR outperforms BiomedCLIP, though pretraining on clean \texttt{mimiccxr} data (as in CheXzero \cite{tiu2022expert}) does perform better. }}
\centering
\resizebox{1.0\linewidth}{!}{  
\begin{tabular}{@{}llrrrr@{}}
\toprule
 & & \multicolumn{2}{c}{\textbf{Random Noise}} & \multicolumn{2}{c}{\textbf{Cat. Noise}} \\ 
\cmidrule(l){3-4} \cmidrule(l){5-6}
 & & \textbf{AUROC} & \textbf{F1} & \textbf{AUROC} & \textbf{F1} \\ 
\midrule
\multirow{3}{*}{\textbf{BiomedCLIP}} 
& Clip Sim.       & 66.8 (0.8)          & 60.1 (0.4)     & 64.1 (0.4)          & 59.2 (0.0)   \\
& \oursfixed (Ours) & 69.5 (0.7)          & -              & 66.3 (0.4)          & -           \\
& \oursopt (Ours)   & \textbf{73.7} (1.7) & \textbf{63.5} (0.8)  & \textbf{70.4} (1.6) & \textbf{61.1} (0.8) \\
\midrule
\multirow{3}{*}{\textbf{\begin{tabular}[c]{@{}l@{}}CLIP Pretrain \\  On Noisy Data\end{tabular}}}
& Clip Sim.       & 78.8 (0.1)          & 66.8 (0.3)     & 76.5 (0.5)          & 64.4 (0.5)	   \\
& \oursfixed (Ours) & \textbf{80.5} (0.1) & -              & 77.0 (0.5)          & -           \\
& \oursopt (Ours)   & 80.0 (0.9) & \textbf{67.7} (0.5)
  & \textbf{77.2} (0.8) & \textbf{64.6} (0.3)	 \\
\midrule
\multirow{3}{*}{\textbf{CheXzero}}
& Clip Sim.       & 90.8 (0.2)          & 79.5 (0.2)     & 88.4 (0.6)          & 76.1 (0.7)	   \\
& \oursfixed (Ours) & 91.4 (0.1)          & -              & 88.4 (0.7)          & -           \\
& \oursopt (Ours)   & \textbf{91.8} (0.2)	 & \textbf{80.9} (0.6)  & \textbf{88.5} (0.4)	 & \textbf{76.2} (1.0)	 \\
\bottomrule
\end{tabular}
}
\label{tab:clip_from_scratch}
\end{table}

\paragraph{Web-Scale Corpus} Motivated by this result, we conduct a large scale experiment on the CC3M dataset \cite{changpinyo2021conceptual}, which contains 2.9 million valid URLs to image-caption pairs. We pretrain CLIP from scratch on this dataset, then use this CLIP model to filter samples in the original dataset using \oursfixed and the CLIP similarity baseline. We select the 1 million samples with the lowest mislabel scores from each method, and pretrain another CLIP from scratch on this clean subset. We evaluate the resulting model on zero-shot classification using the VTAB benchmark \cite{zhai2019visual}. We find filtering with LEMoN marginally outperforms the baseline on average zero-shot accuracy, though both underperform pretraining on the full corpus. Full details are in Appendix \ref{app:cc3m_clip}. We additionally conduct an experiment on Datacomp \cite{gadre2024datacomp} in Appendix \ref{app:datacomp}.

\subsection{Real-World Analysis}
\label{sec:real_world}
We conduct a preliminary study of \ours on real datasets without known label errors. We run \oursfixed and the CLIP similarity baseline on \texttt{cifar10}, \texttt{cifar100}, \texttt{flickr30k}, and \texttt{mscoco}. As no labeled validation set is available, we use optimal hyperparameters from models previously run on each dataset with synthetic noise from Section \ref{sec:our_perf} (Appendix \ref{sec:opt_hparams_real_app}). For each dataset, we select the top 200 images from the validation and test splits with the highest mislabel scores. We then manually annotated each sample to determine whether it was mislabeled. Crucially, during labeling, images were randomly selected, so the labeler is unaware of whether the candidate image originated from the baseline or our method. We find that our method outperforms the baseline for every dataset (Table \ref{tab:real_world}). Examples of real-world mislabels are also in Figures~\ref{fig:teaser} and \ref{fig:teaser_full}. We present a further comparison of our identified error sets in \texttt{cifar10} and \texttt{cifar100} with crowd-sourced labels \cite{northcutt2021pervasive} in Appendix~\ref{app:cl_real_world}.

\begin{table}[htbp!]
\caption{We manually label 200 images from real-world datasets that each method identifies as the most likely to be mislabeled and show the percentage (\%) of times where it is actually mislabeled. Numbers in parentheses are 95\% confidence intervals from a binomial proportion.}
\centering
\resizebox{0.55\linewidth}{!}{  
\begin{tabular}{@{}rrr@{}}
\toprule
\multicolumn{1}{l}{} & \textbf{CLIP Sim.}  & \textbf{Ours}       \\ \midrule
\texttt{cifar10}              & 5.5 (3.2)  &\textbf{ 10.0} (4.2) \\
\texttt{cifar100}             & 11.0 (4.3) & \textbf{20.5} (5.6) \\
\texttt{flickr30k}            & 32.5 (6.5) &\textbf{ 41.0} (6.8) \\
\texttt{mscoco}              & 19.5 (5.5) & \textbf{25.5} (6.0) \\ \bottomrule
\end{tabular}
}
\label{tab:real_world}
\end{table}

\section{Conclusion}

In this work, we proposed \ours, a method that leverages the neighborhood structure of contrastively pretrained multimodal embeddings to automatically identify label errors in image datasets with natural language text labels.

Our approach is a promising step to automatically detecting and filtering data mislabels at scale. Through experiments on multiple datasets with synthetic and real-world noise, we demonstrated \ours's effectiveness in detecting label errors and improving downstream model performance.

\paragraph{Limitations} Our work has several limitations. As we primarily rely on existing open-sourced datasets, some parts of these datasets may have been used as training data in pretrained models. We specifically chose pretrained models that take care not to include the test sets of such datasets. Further, we run experiments on a real-world healthcare dataset with access controls (\texttt{mimiccxr}) to verify our results. Next, we assume that there exists an oracle binary indicator for whether a sample is mislabeled. As we saw in practice, real-world mislabels contain much more uncertainty and ambiguity, e.g. due to blurry images~\cite{gao2017deep,  basile2021toward,gordon2021disagreement,gordon2022jury}. Evaluating the effectiveness of our score as a measure of this uncertainty is an area of future work.

\section*{Acknowledgments}
This work was supported in part by a National Science Foundation (NSF) 22-586 Faculty Early Career Development Award (\#2339381), a Gordon \& Betty Moore Foundation award, and a Google Research Scholar award. We would like to thank Walter Gerych and Olawale Salaudeen for their valuable feedback.

\section*{Impact Statement}

In this work, we have studied the identification and removal of samples deemed to be mislabels. We recognize that not all such samples are truly erroneous, and that these false positives may not be uniformly distributed across the space of images and text. For example, automated detectors such as \ours may flag legitimate but atypical examples, such as rare concepts, under-represented languages or dialectic phrases, and images from minority classes or groups. Removal of these samples may then lead to downstream models with certain biases. To alleviate these issues, we encourage practitioners to use \ours and similar tools to surface candidates for expert review, rather than as an unsupervised pruning tool. We view the ability to manually examine such samples as a strength of the filtering approach, over loss-based approaches, for learning under label noise.

\bibliography{bibfile}

\begin{thebibliography}{89}
\providecommand{\natexlab}[1]{#1}
\providecommand{\url}[1]{\texttt{#1}}
\expandafter\ifx\csname urlstyle\endcsname\relax
  \providecommand{\doi}[1]{doi: #1}\else
  \providecommand{\doi}{doi: \begingroup \urlstyle{rm}\Url}\fi

\bibitem[Arevalo et~al.(2017)Arevalo, Solorio, Montes-y G{\'o}mez, and Gonz{\'a}lez]{arevalo2017gated}
Arevalo, J., Solorio, T., Montes-y G{\'o}mez, M., and Gonz{\'a}lez, F.~A.
\newblock Gated multimodal units for information fusion.
\newblock \emph{arXiv preprint arXiv:1702.01992}, 2017.

\bibitem[Bahri et~al.(2020)Bahri, Jiang, and Gupta]{bahri2020deep}
Bahri, D., Jiang, H., and Gupta, M.
\newblock Deep k-nn for noisy labels.
\newblock In \emph{International Conference on Machine Learning}, pp.\  540--550. PMLR, 2020.

\bibitem[Balestriero et~al.(2023)Balestriero, Ibrahim, Sobal, Morcos, Shekhar, Goldstein, Bordes, Bardes, Mialon, Tian, et~al.]{balestriero2023cookbook}
Balestriero, R., Ibrahim, M., Sobal, V., Morcos, A., Shekhar, S., Goldstein, T., Bordes, F., Bardes, A., Mialon, G., Tian, Y., et~al.
\newblock A cookbook of self-supervised learning.
\newblock \emph{arXiv preprint arXiv:2304.12210}, 2023.

\bibitem[Basile et~al.(2021)Basile, Cabitza, Campagner, and Fell]{basile2021toward}
Basile, V., Cabitza, F., Campagner, A., and Fell, M.
\newblock Toward a perspectivist turn in ground truthing for predictive computing.
\newblock \emph{arXiv preprint arXiv:2109.04270}, 2021.

\bibitem[Bernhardt et~al.(2022)Bernhardt, Castro, Tanno, Schwaighofer, Tezcan, Monteiro, Bannur, Lungren, Nori, Glocker, et~al.]{bernhardt2022active}
Bernhardt, M., Castro, D.~C., Tanno, R., Schwaighofer, A., Tezcan, K.~C., Monteiro, M., Bannur, S., Lungren, M.~P., Nori, A., Glocker, B., et~al.
\newblock Active label cleaning for improved dataset quality under resource constraints.
\newblock \emph{Nature communications}, 13\penalty0 (1):\penalty0 1161, 2022.

\bibitem[Beyer et~al.(2020)Beyer, H{\'e}naff, Kolesnikov, Zhai, and Oord]{beyer2020we}
Beyer, L., H{\'e}naff, O.~J., Kolesnikov, A., Zhai, X., and Oord, A. v.~d.
\newblock Are we done with imagenet?
\newblock \emph{arXiv preprint arXiv:2006.07159}, 2020.

\bibitem[Cai et~al.(2023)Cai, Qiu, Chen, Zhang, and Chen]{cai2023semantic}
Cai, S., Qiu, L., Chen, X., Zhang, Q., and Chen, L.
\newblock Semantic-enhanced image clustering.
\newblock In \emph{Proceedings of the AAAI conference on artificial intelligence}, volume~37, pp.\  6869--6878, 2023.

\bibitem[Changpinyo et~al.(2021)Changpinyo, Sharma, Ding, and Soricut]{changpinyo2021conceptual}
Changpinyo, S., Sharma, P., Ding, N., and Soricut, R.
\newblock Conceptual 12m: Pushing web-scale image-text pre-training to recognize long-tail visual concepts.
\newblock In \emph{Proceedings of the IEEE/CVF conference on computer vision and pattern recognition}, pp.\  3558--3568, 2021.

\bibitem[Chen et~al.(2023)Chen, Wang, Shah, Tao, Wei, Xie, Sugiyama, and Raj]{chen2023understanding}
Chen, H., Wang, J., Shah, A., Tao, R., Wei, H., Xie, X., Sugiyama, M., and Raj, B.
\newblock Understanding and mitigating the label noise in pre-training on downstream tasks.
\newblock \emph{arXiv preprint arXiv:2309.17002}, 2023.

\bibitem[Chen et~al.(2024)Chen, Gallifant, Gao, Moreira, Munch, Muthukkumar, Rajan, Kolluri, Fiske, Hastings, et~al.]{chen2024cross}
Chen, S., Gallifant, J., Gao, M., Moreira, P., Munch, N., Muthukkumar, A., Rajan, A., Kolluri, J., Fiske, A., Hastings, J., et~al.
\newblock Cross-care: Assessing the healthcare implications of pre-training data on language model bias.
\newblock \emph{arXiv preprint arXiv:2405.05506}, 2024.

\bibitem[Chen et~al.(2020)Chen, Kornblith, Norouzi, and Hinton]{chen2020simple}
Chen, T., Kornblith, S., Norouzi, M., and Hinton, G.
\newblock A simple framework for contrastive learning of visual representations.
\newblock In \emph{International conference on machine learning}, pp.\  1597--1607. PMLR, 2020.

\bibitem[Cui et~al.(2020)Cui, Zhang, and Ji]{cui2020label}
Cui, Z., Zhang, Y., and Ji, Q.
\newblock Label error correction and generation through label relationships.
\newblock In \emph{Proceedings of the AAAI Conference on Artificial Intelligence}, volume~34, pp.\  3693--3700, 2020.

\bibitem[Dai et~al.(2023)Dai, Li, Li, Tiong, Zhao, Wang, Li, Fung, and Hoi]{dai2023instructblip}
Dai, W., Li, J., Li, D., Tiong, A., Zhao, J., Wang, W., Li, B., Fung, P., and Hoi, S.
\newblock Instructblip: Towards general-purpose vision-language models with instruction tuning. arxiv 2023.
\newblock \emph{arXiv preprint arXiv:2305.06500}, 2, 2023.

\bibitem[Diffenderfer et~al.(2021)Diffenderfer, Bartoldson, Chaganti, Zhang, and Kailkhura]{diffenderfer2021winning}
Diffenderfer, J., Bartoldson, B., Chaganti, S., Zhang, J., and Kailkhura, B.
\newblock A winning hand: Compressing deep networks can improve out-of-distribution robustness.
\newblock \emph{Advances in neural information processing systems}, 34:\penalty0 664--676, 2021.

\bibitem[Dosovitskiy et~al.(2020)Dosovitskiy, Beyer, Kolesnikov, Weissenborn, Zhai, Unterthiner, Dehghani, Minderer, Heigold, Gelly, et~al.]{dosovitskiy2020image}
Dosovitskiy, A., Beyer, L., Kolesnikov, A., Weissenborn, D., Zhai, X., Unterthiner, T., Dehghani, M., Minderer, M., Heigold, G., Gelly, S., et~al.
\newblock An image is worth 16x16 words: Transformers for image recognition at scale.
\newblock In \emph{International Conference on Learning Representations}, 2020.

\bibitem[Dubey et~al.(2024)Dubey, Jauhri, Pandey, Kadian, Al-Dahle, Letman, Mathur, Schelten, Yang, Fan, et~al.]{dubey2024llama}
Dubey, A., Jauhri, A., Pandey, A., Kadian, A., Al-Dahle, A., Letman, A., Mathur, A., Schelten, A., Yang, A., Fan, A., et~al.
\newblock The llama 3 herd of models.
\newblock \emph{arXiv preprint arXiv:2407.21783}, 2024.

\bibitem[Feng et~al.()Feng, Tzimiropoulos, and Patras]{feng2024clipcleaner}
Feng, C., Tzimiropoulos, G., and Patras, I.
\newblock Clipcleaner: Cleaning noisy labels with clip.
\newblock In \emph{ACM Multimedia 2024}.

\bibitem[Fu et~al.(2024)Fu, Song, Zhou, and Yang]{fu2024noise}
Fu, Z., Song, K., Zhou, L., and Yang, Y.
\newblock Noise-aware image captioning with progressively exploring mismatched words.
\newblock In \emph{Proceedings of the AAAI Conference on Artificial Intelligence}, volume~38, pp.\  12091--12099, 2024.

\bibitem[Gadre et~al.(2024)Gadre, Ilharco, Fang, Hayase, Smyrnis, Nguyen, Marten, Wortsman, Ghosh, Zhang, et~al.]{gadre2024datacomp}
Gadre, S.~Y., Ilharco, G., Fang, A., Hayase, J., Smyrnis, G., Nguyen, T., Marten, R., Wortsman, M., Ghosh, D., Zhang, J., et~al.
\newblock Datacomp: In search of the next generation of multimodal datasets.
\newblock \emph{Advances in Neural Information Processing Systems}, 36, 2024.

\bibitem[Gao et~al.(2017)Gao, Xing, Xie, Wu, and Geng]{gao2017deep}
Gao, B.-B., Xing, C., Xie, C.-W., Wu, J., and Geng, X.
\newblock Deep label distribution learning with label ambiguity.
\newblock \emph{IEEE Transactions on Image Processing}, 26\penalty0 (6):\penalty0 2825--2838, 2017.

\bibitem[Gertz et~al.(2024)Gertz, Dratsch, Bunck, Lennartz, Iuga, Hellmich, Persigehl, Pennig, Gietzen, Fervers, et~al.]{gertz2024potential}
Gertz, R.~J., Dratsch, T., Bunck, A.~C., Lennartz, S., Iuga, A.-I., Hellmich, M.~G., Persigehl, T., Pennig, L., Gietzen, C.~H., Fervers, P., et~al.
\newblock Potential of gpt-4 for detecting errors in radiology reports: Implications for reporting accuracy.
\newblock \emph{Radiology}, 311\penalty0 (1):\penalty0 e232714, 2024.

\bibitem[Ghasemi \& Zahediasl(2012)Ghasemi and Zahediasl]{ghasemi2012normality}
Ghasemi, A. and Zahediasl, S.
\newblock Normality tests for statistical analysis: a guide for non-statisticians.
\newblock \emph{International journal of endocrinology and metabolism}, 10\penalty0 (2):\penalty0 486, 2012.

\bibitem[Gordon et~al.(2021)Gordon, Zhou, Patel, Hashimoto, and Bernstein]{gordon2021disagreement}
Gordon, M.~L., Zhou, K., Patel, K., Hashimoto, T., and Bernstein, M.~S.
\newblock The disagreement deconvolution: Bringing machine learning performance metrics in line with reality.
\newblock In \emph{Proceedings of the 2021 CHI Conference on Human Factors in Computing Systems}, pp.\  1--14, 2021.

\bibitem[Gordon et~al.(2022)Gordon, Lam, Park, Patel, Hancock, Hashimoto, and Bernstein]{gordon2022jury}
Gordon, M.~L., Lam, M.~S., Park, J.~S., Patel, K., Hancock, J., Hashimoto, T., and Bernstein, M.~S.
\newblock Jury learning: Integrating dissenting voices into machine learning models.
\newblock In \emph{Proceedings of the 2022 CHI Conference on Human Factors in Computing Systems}, pp.\  1--19, 2022.

\bibitem[Grattafiori et~al.(2024)Grattafiori, Dubey, Jauhri, Pandey, Kadian, Al-Dahle, Letman, Mathur, Schelten, Vaughan, et~al.]{grattafiori2024llama}
Grattafiori, A., Dubey, A., Jauhri, A., Pandey, A., Kadian, A., Al-Dahle, A., Letman, A., Mathur, A., Schelten, A., Vaughan, A., et~al.
\newblock The llama 3 herd of models.
\newblock \emph{arXiv preprint arXiv:2407.21783}, 2024.

\bibitem[Grivas et~al.(2020)Grivas, Alex, Grover, Tobin, and Whiteley]{grivas2020not}
Grivas, A., Alex, B., Grover, C., Tobin, R., and Whiteley, W.
\newblock Not a cute stroke: analysis of rule-and neural network-based information extraction systems for brain radiology reports.
\newblock In \emph{Proceedings of the 11th international workshop on health text mining and information analysis}, pp.\  24--37, 2020.

\bibitem[Hendrycks \& Dietterich(2018)Hendrycks and Dietterich]{hendrycks2018benchmarking}
Hendrycks, D. and Dietterich, T.
\newblock Benchmarking neural network robustness to common corruptions and perturbations.
\newblock In \emph{International Conference on Learning Representations}, 2018.

\bibitem[Honnibal \& Montani(2017)Honnibal and Montani]{spacy2}
Honnibal, M. and Montani, I.
\newblock {spaCy 2}: Natural language understanding with {B}loom embeddings, convolutional neural networks and incremental parsing.
\newblock To appear, 2017.

\bibitem[Hu et~al.(2021)Hu, Wallis, Allen-Zhu, Li, Wang, Wang, Chen, et~al.]{hu2021lora}
Hu, E.~J., Wallis, P., Allen-Zhu, Z., Li, Y., Wang, S., Wang, L., Chen, W., et~al.
\newblock Lora: Low-rank adaptation of large language models.
\newblock In \emph{International Conference on Learning Representations}, 2021.

\bibitem[Hu et~al.(2023)Hu, Cavicchioli, and Capotondi]{hu2023exploiting}
Hu, J.~C., Cavicchioli, R., and Capotondi, A.
\newblock Exploiting multiple sequence lengths in fast end to end training for image captioning.
\newblock In \emph{2023 IEEE International Conference on Big Data (BigData)}, pp.\  2173--2182. IEEE, 2023.

\bibitem[Huang et~al.(2024)Huang, He, Wang, Chen, Li, Feng, Wang, and Zhu]{huang2024neighbor}
Huang, B., He, F., Wang, Q., Chen, H., Li, G., Feng, Z., Wang, X., and Zhu, W.
\newblock Neighbor does matter: Global positive-negative sampling for vision-language pre-training.
\newblock In \emph{ACM Multimedia 2024}, 2024.

\bibitem[Huang et~al.(2023)Huang, Long, Han, Xu, Liang, Xu, and Liang]{huang2023nlip}
Huang, R., Long, Y., Han, J., Xu, H., Liang, X., Xu, C., and Liang, X.
\newblock Nlip: Noise-robust language-image pre-training.
\newblock In \emph{Proceedings of the AAAI Conference on Artificial Intelligence}, volume~37, pp.\  926--934, 2023.

\bibitem[Jiang et~al.(2020)Jiang, Huang, Liu, and Yang]{jiang2020beyond}
Jiang, L., Huang, D., Liu, M., and Yang, W.
\newblock Beyond synthetic noise: Deep learning on controlled noisy labels.
\newblock In \emph{International conference on machine learning}, pp.\  4804--4815. PMLR, 2020.

\bibitem[Johnson et~al.(2019)Johnson, Pollard, Greenbaum, Lungren, Deng, Peng, Lu, Mark, Berkowitz, and Horng]{johnson2019mimic}
Johnson, A.~E., Pollard, T.~J., Greenbaum, N.~R., Lungren, M.~P., Deng, C.-y., Peng, Y., Lu, Z., Mark, R.~G., Berkowitz, S.~J., and Horng, S.
\newblock Mimic-cxr-jpg, a large publicly available database of labeled chest radiographs.
\newblock \emph{arXiv preprint arXiv:1901.07042}, 2019.

\bibitem[Kang et~al.(2023)Kang, Mun, Lee, and Roh]{kang2023noise}
Kang, W., Mun, J., Lee, S., and Roh, B.
\newblock Noise-aware learning from web-crawled image-text data for image captioning.
\newblock In \emph{Proceedings of the IEEE/CVF International Conference on Computer Vision}, pp.\  2942--2952, 2023.

\bibitem[Karpathy \& Fei-Fei(2015)Karpathy and Fei-Fei]{karpathy2015deep}
Karpathy, A. and Fei-Fei, L.
\newblock Deep visual-semantic alignments for generating image descriptions.
\newblock In \emph{Proceedings of the IEEE conference on computer vision and pattern recognition}, pp.\  3128--3137, 2015.

\bibitem[Kim et~al.(2021)Kim, Ko, Choi, Yun, et~al.]{kim2021fine}
Kim, T., Ko, J., Choi, J., Yun, S.-Y., et~al.
\newblock Fine samples for learning with noisy labels.
\newblock \emph{Advances in Neural Information Processing Systems}, 34:\penalty0 24137--24149, 2021.

\bibitem[Lai et~al.(2023)Lai, Vesdapunt, Zhou, Wu, Huynh, Li, Fu, and Chuah]{lai2023padclip}
Lai, Z., Vesdapunt, N., Zhou, N., Wu, J., Huynh, C.~P., Li, X., Fu, K.~K., and Chuah, C.-N.
\newblock Padclip: Pseudo-labeling with adaptive debiasing in clip for unsupervised domain adaptation.
\newblock In \emph{Proceedings of the IEEE/CVF International Conference on Computer Vision}, pp.\  16155--16165, 2023.

\bibitem[Li et~al.(2022)Li, Li, Xiong, and Hoi]{li2022blip}
Li, J., Li, D., Xiong, C., and Hoi, S.
\newblock Blip: Bootstrapping language-image pre-training for unified vision-language understanding and generation.
\newblock In \emph{International conference on machine learning}, pp.\  12888--12900. PMLR, 2022.

\bibitem[Li et~al.(2023)Li, Li, Savarese, and Hoi]{li2023blip}
Li, J., Li, D., Savarese, S., and Hoi, S.
\newblock Blip-2: Bootstrapping language-image pre-training with frozen image encoders and large language models.
\newblock In \emph{International conference on machine learning}, pp.\  19730--19742. PMLR, 2023.

\bibitem[Li et~al.(2021)Li, Liang, Zhao, Cui, Ouyang, Shao, Yu, and Yan]{lisupervision}
Li, Y., Liang, F., Zhao, L., Cui, Y., Ouyang, W., Shao, J., Yu, F., and Yan, J.
\newblock Supervision exists everywhere: A data efficient contrastive language-image pre-training paradigm.
\newblock In \emph{International Conference on Learning Representations}, 2021.

\bibitem[Liang et~al.(2023)Liang, Zhu, Shi, and Yang]{liang2023combating}
Liang, C., Zhu, L., Shi, H., and Yang, Y.
\newblock Combating label noise with a general surrogate model for sample selection.
\newblock \emph{arXiv preprint arXiv:2310.10463}, 2023.

\bibitem[Liang et~al.(2022)Liang, Zhang, Kwon, Yeung, and Zou]{liang2022mind}
Liang, V.~W., Zhang, Y., Kwon, Y., Yeung, S., and Zou, J.~Y.
\newblock Mind the gap: Understanding the modality gap in multi-modal contrastive representation learning.
\newblock \emph{Advances in Neural Information Processing Systems}, 35:\penalty0 17612--17625, 2022.

\bibitem[Liao et~al.(2021)Liao, Taori, Raji, and Schmidt]{liao2021we}
Liao, T., Taori, R., Raji, I.~D., and Schmidt, L.
\newblock Are we learning yet? a meta review of evaluation failures across machine learning.
\newblock In \emph{Thirty-fifth Conference on Neural Information Processing Systems Datasets and Benchmarks Track (Round 2)}, 2021.

\bibitem[Lin(2004)]{lin2004rouge}
Lin, C.-Y.
\newblock Rouge: A package for automatic evaluation of summaries.
\newblock In \emph{Text summarization branches out}, pp.\  74--81, 2004.

\bibitem[Lin et~al.(2022)Lin, Li, Lin, Ahmed, Gan, Liu, Lu, and Wang]{lin2022swinbert}
Lin, K., Li, L., Lin, C.-C., Ahmed, F., Gan, Z., Liu, Z., Lu, Y., and Wang, L.
\newblock Swinbert: End-to-end transformers with sparse attention for video captioning.
\newblock In \emph{Proceedings of the IEEE/CVF Conference on Computer Vision and Pattern Recognition}, pp.\  17949--17958, 2022.

\bibitem[Lin et~al.(2014)Lin, Maire, Belongie, Hays, Perona, Ramanan, Doll{\'a}r, and Zitnick]{lin2014microsoft}
Lin, T.-Y., Maire, M., Belongie, S., Hays, J., Perona, P., Ramanan, D., Doll{\'a}r, P., and Zitnick, C.~L.
\newblock Microsoft coco: Common objects in context.
\newblock In \emph{Computer Vision--ECCV 2014: 13th European Conference, Zurich, Switzerland, September 6-12, 2014, Proceedings, Part V 13}, pp.\  740--755. Springer, 2014.

\bibitem[Liu et~al.(2023)Liu, Bubeck, Eldan, Kulkarni, Li, Nguyen, Ward, and Zhang]{liu2023tinygsm}
Liu, B., Bubeck, S., Eldan, R., Kulkarni, J., Li, Y., Nguyen, A., Ward, R., and Zhang, Y.
\newblock Tinygsm: achieving> 80\% on gsm8k with small language models.
\newblock \emph{arXiv preprint arXiv:2312.09241}, 2023.

\bibitem[Liu et~al.(2024)Liu, Li, Wu, and Lee]{liu2024visual}
Liu, H., Li, C., Wu, Q., and Lee, Y.~J.
\newblock Visual instruction tuning.
\newblock \emph{Advances in neural information processing systems}, 36, 2024.

\bibitem[Longpre et~al.(2023)Longpre, Yauney, Reif, Lee, Roberts, Zoph, Zhou, Wei, Robinson, Mimno, et~al.]{longpre2023pretrainer}
Longpre, S., Yauney, G., Reif, E., Lee, K., Roberts, A., Zoph, B., Zhou, D., Wei, J., Robinson, K., Mimno, D., et~al.
\newblock A pretrainer's guide to training data: Measuring the effects of data age, domain coverage, quality, \& toxicity.
\newblock \emph{arXiv preprint arXiv:2305.13169}, 2023.

\bibitem[Loshchilov \& Hutter(2018)Loshchilov and Hutter]{loshchilov2018decoupled}
Loshchilov, I. and Hutter, F.
\newblock Decoupled weight decay regularization.
\newblock In \emph{International Conference on Learning Representations}, 2018.

\bibitem[Luccioni \& Rolnick(2023)Luccioni and Rolnick]{luccioni2023bugs}
Luccioni, A.~S. and Rolnick, D.
\newblock Bugs in the data: How imagenet misrepresents biodiversity.
\newblock In \emph{Proceedings of the AAAI Conference on Artificial Intelligence}, volume~37, pp.\  14382--14390, 2023.

\bibitem[Menghini et~al.(2023)Menghini, Delworth, and Bach]{menghini2023enhancing}
Menghini, C., Delworth, A., and Bach, S.
\newblock Enhancing clip with clip: Exploring pseudolabeling for limited-label prompt tuning.
\newblock \emph{Advances in Neural Information Processing Systems}, 36:\penalty0 60984--61007, 2023.

\bibitem[Misra \& Maaten(2020)Misra and Maaten]{misra2020self}
Misra, I. and Maaten, L. v.~d.
\newblock Self-supervised learning of pretext-invariant representations.
\newblock In \emph{Proceedings of the IEEE/CVF conference on computer vision and pattern recognition}, pp.\  6707--6717, 2020.

\bibitem[Natarajan et~al.(2013)Natarajan, Dhillon, Ravikumar, and Tewari]{natarajan2013learning}
Natarajan, N., Dhillon, I.~S., Ravikumar, P.~K., and Tewari, A.
\newblock Learning with noisy labels.
\newblock \emph{Advances in neural information processing systems}, 26, 2013.

\bibitem[Nguyen et~al.(2022)Nguyen, Ilharco, Wortsman, Oh, and Schmidt]{nguyen2022quality}
Nguyen, T., Ilharco, G., Wortsman, M., Oh, S., and Schmidt, L.
\newblock Quality not quantity: On the interaction between dataset design and robustness of clip.
\newblock \emph{Advances in Neural Information Processing Systems}, 35:\penalty0 21455--21469, 2022.

\bibitem[Northcutt et~al.(2021{\natexlab{a}})Northcutt, Jiang, and Chuang]{northcutt2021confident}
Northcutt, C., Jiang, L., and Chuang, I.
\newblock Confident learning: Estimating uncertainty in dataset labels.
\newblock \emph{Journal of Artificial Intelligence Research}, 70:\penalty0 1373--1411, 2021{\natexlab{a}}.

\bibitem[Northcutt et~al.(2021{\natexlab{b}})Northcutt, Athalye, and Mueller]{northcutt2021pervasive}
Northcutt, C.~G., Athalye, A., and Mueller, J.
\newblock Pervasive label errors in test sets destabilize machine learning benchmarks.
\newblock \emph{arXiv preprint arXiv:2103.14749}, 2021{\natexlab{b}}.

\bibitem[Oord et~al.(2018)Oord, Li, and Vinyals]{oord2018representation}
Oord, A. v.~d., Li, Y., and Vinyals, O.
\newblock Representation learning with contrastive predictive coding.
\newblock \emph{arXiv preprint arXiv:1807.03748}, 2018.

\bibitem[Papineni et~al.(2002)Papineni, Roukos, Ward, and Zhu]{papineni2002bleu}
Papineni, K., Roukos, S., Ward, T., and Zhu, W.-J.
\newblock Bleu: a method for automatic evaluation of machine translation.
\newblock In \emph{Proceedings of the 40th annual meeting of the Association for Computational Linguistics}, pp.\  311--318, 2002.

\bibitem[Pleiss et~al.(2020)Pleiss, Zhang, Elenberg, and Weinberger]{pleiss2020identifying}
Pleiss, G., Zhang, T., Elenberg, E., and Weinberger, K.~Q.
\newblock Identifying mislabeled data using the area under the margin ranking.
\newblock \emph{Advances in Neural Information Processing Systems}, 33:\penalty0 17044--17056, 2020.

\bibitem[Plummer et~al.(2015)Plummer, Wang, Cervantes, Caicedo, Hockenmaier, and Lazebnik]{plummer2015flickr30k}
Plummer, B.~A., Wang, L., Cervantes, C.~M., Caicedo, J.~C., Hockenmaier, J., and Lazebnik, S.
\newblock Flickr30k entities: Collecting region-to-phrase correspondences for richer image-to-sentence models.
\newblock In \emph{Proceedings of the IEEE international conference on computer vision}, pp.\  2641--2649, 2015.

\bibitem[Radford et~al.(2021)Radford, Kim, Hallacy, Ramesh, Goh, Agarwal, Sastry, Askell, Mishkin, Clark, et~al.]{radford2021learning}
Radford, A., Kim, J.~W., Hallacy, C., Ramesh, A., Goh, G., Agarwal, S., Sastry, G., Askell, A., Mishkin, P., Clark, J., et~al.
\newblock Learning transferable visual models from natural language supervision.
\newblock In \emph{International conference on machine learning}, pp.\  8748--8763. PMLR, 2021.

\bibitem[Rennie et~al.(2017)Rennie, Marcheret, Mroueh, Ross, and Goel]{rennie2017self}
Rennie, S.~J., Marcheret, E., Mroueh, Y., Ross, J., and Goel, V.
\newblock Self-critical sequence training for image captioning.
\newblock In \emph{Proceedings of the IEEE conference on computer vision and pattern recognition}, pp.\  7008--7024, 2017.

\bibitem[Ridnik et~al.(2021)Ridnik, Ben-Baruch, Noy, and Zelnik-Manor]{ridnikimagenet}
Ridnik, T., Ben-Baruch, E., Noy, A., and Zelnik-Manor, L.
\newblock Imagenet-21k pretraining for the masses.
\newblock \emph{Proceedings of 35th Conference on Neural Information Processing Systems, Track on Datasets and Benchmarks}, 2021.

\bibitem[Rottmann \& Reese(2023)Rottmann and Reese]{rottmann2023automated}
Rottmann, M. and Reese, M.
\newblock Automated detection of label errors in semantic segmentation datasets via deep learning and uncertainty quantification.
\newblock In \emph{Proceedings of the IEEE/CVF Winter Conference on Applications of Computer Vision}, pp.\  3214--3223, 2023.

\bibitem[Russakovsky et~al.(2015)Russakovsky, Deng, Su, Krause, Satheesh, Ma, Huang, Karpathy, Khosla, Bernstein, et~al.]{russakovsky2015imagenet}
Russakovsky, O., Deng, J., Su, H., Krause, J., Satheesh, S., Ma, S., Huang, Z., Karpathy, A., Khosla, A., Bernstein, M., et~al.
\newblock Imagenet large scale visual recognition challenge.
\newblock \emph{International journal of computer vision}, 115:\penalty0 211--252, 2015.

\bibitem[Schrodi et~al.(2024)Schrodi, Hoffmann, Argus, Fischer, and Brox]{schrodi2024two}
Schrodi, S., Hoffmann, D.~T., Argus, M., Fischer, V., and Brox, T.
\newblock Two effects, one trigger: On the modality gap, object bias, and information imbalance in contrastive vision-language representation learning.
\newblock \emph{arXiv preprint arXiv:2404.07983}, 2024.

\bibitem[Schroff et~al.(2015)Schroff, Kalenichenko, and Philbin]{schroff2015facenet}
Schroff, F., Kalenichenko, D., and Philbin, J.
\newblock Facenet: A unified embedding for face recognition and clustering.
\newblock In \emph{Proceedings of the IEEE conference on computer vision and pattern recognition}, pp.\  815--823, 2015.

\bibitem[Schubert et~al.(2024)Schubert, Riedlinger, Kahl, Kr{\"o}ll, Schoenen, {\v{S}}egvi{\'c}, and Rottmann]{schubert2024identifying}
Schubert, M., Riedlinger, T., Kahl, K., Kr{\"o}ll, D., Schoenen, S., {\v{S}}egvi{\'c}, S., and Rottmann, M.
\newblock Identifying label errors in object detection datasets by loss inspection.
\newblock In \emph{Proceedings of the IEEE/CVF Winter Conference on Applications of Computer Vision}, pp.\  4582--4591, 2024.

\bibitem[Schuhmann et~al.(2021)Schuhmann, Vencu, Beaumont, Kaczmarczyk, Mullis, Katta, Coombes, Jitsev, and Komatsuzaki]{schuhmann2021laion}
Schuhmann, C., Vencu, R., Beaumont, R., Kaczmarczyk, R., Mullis, C., Katta, A., Coombes, T., Jitsev, J., and Komatsuzaki, A.
\newblock Laion-400m: Open dataset of clip-filtered 400 million image-text pairs.
\newblock \emph{arXiv preprint arXiv:2111.02114}, 2021.

\bibitem[Sohn(2016)]{sohn2016improved}
Sohn, K.
\newblock Improved deep metric learning with multi-class n-pair loss objective.
\newblock \emph{Advances in neural information processing systems}, 29, 2016.

\bibitem[Swayamdipta et~al.(2020)Swayamdipta, Schwartz, Lourie, Wang, Hajishirzi, Smith, and Choi]{swayamdipta2020dataset}
Swayamdipta, S., Schwartz, R., Lourie, N., Wang, Y., Hajishirzi, H., Smith, N.~A., and Choi, Y.
\newblock Dataset cartography: Mapping and diagnosing datasets with training dynamics.
\newblock \emph{arXiv preprint arXiv:2009.10795}, 2020.

\bibitem[Thomas \& Kovashka(2020)Thomas and Kovashka]{thomas2020preserving}
Thomas, C. and Kovashka, A.
\newblock Preserving semantic neighborhoods for robust cross-modal retrieval.
\newblock In \emph{Computer Vision--ECCV 2020: 16th European Conference, Glasgow, UK, August 23--28, 2020, Proceedings, Part XVIII 16}, pp.\  317--335. Springer, 2020.

\bibitem[Thomas \& Kovashka(2022)Thomas and Kovashka]{thomas2022emphasizing}
Thomas, C. and Kovashka, A.
\newblock Emphasizing complementary samples for non-literal cross-modal retrieval.
\newblock In \emph{Proceedings of the IEEE/CVF Conference on Computer Vision and Pattern Recognition}, pp.\  4632--4641, 2022.

\bibitem[Tiu et~al.(2022)Tiu, Talius, Patel, Langlotz, Ng, and Rajpurkar]{tiu2022expert}
Tiu, E., Talius, E., Patel, P., Langlotz, C.~P., Ng, A.~Y., and Rajpurkar, P.
\newblock Expert-level detection of pathologies from unannotated chest x-ray images via self-supervised learning.
\newblock \emph{Nature Biomedical Engineering}, 6\penalty0 (12):\penalty0 1399--1406, 2022.

\bibitem[Vasudevan et~al.(2022)Vasudevan, Caine, Gontijo~Lopes, Fridovich-Keil, and Roelofs]{vasudevan2022does}
Vasudevan, V., Caine, B., Gontijo~Lopes, R., Fridovich-Keil, S., and Roelofs, R.
\newblock When does dough become a bagel? analyzing the remaining mistakes on imagenet.
\newblock \emph{Advances in Neural Information Processing Systems}, 35:\penalty0 6720--6734, 2022.

\bibitem[Vedantam et~al.(2015)Vedantam, Lawrence~Zitnick, and Parikh]{vedantam2015cider}
Vedantam, R., Lawrence~Zitnick, C., and Parikh, D.
\newblock Cider: Consensus-based image description evaluation.
\newblock In \emph{Proceedings of the IEEE conference on computer vision and pattern recognition}, pp.\  4566--4575, 2015.

\bibitem[Wang et~al.(2022{\natexlab{a}})Wang, Yang, Hu, Li, Lin, Gan, Liu, Liu, and Wang]{wang2022git}
Wang, J., Yang, Z., Hu, X., Li, L., Lin, K., Gan, Z., Liu, Z., Liu, C., and Wang, L.
\newblock Git: A generative image-to-text transformer for vision and language.
\newblock \emph{Transactions on Machine Learning Research}, 2022{\natexlab{a}}.

\bibitem[Wang et~al.(2022{\natexlab{b}})Wang, Xu, and Sun]{wang2022end}
Wang, Y., Xu, J., and Sun, Y.
\newblock End-to-end transformer based model for image captioning.
\newblock In \emph{Proceedings of the AAAI Conference on Artificial Intelligence}, volume~36, pp.\  2585--2594, 2022{\natexlab{b}}.

\bibitem[Wei et~al.(2021)Wei, Zhu, Cheng, Liu, Niu, and Liu]{wei2021learning}
Wei, J., Zhu, Z., Cheng, H., Liu, T., Niu, G., and Liu, Y.
\newblock Learning with noisy labels revisited: A study using real-world human annotations.
\newblock \emph{arXiv preprint arXiv:2110.12088}, 2021.

\bibitem[Wu et~al.(2020)Wu, Zheng, Goswami, Metaxas, and Chen]{wu2020topological}
Wu, P., Zheng, S., Goswami, M., Metaxas, D., and Chen, C.
\newblock A topological filter for learning with label noise.
\newblock \emph{Advances in neural information processing systems}, 33:\penalty0 21382--21393, 2020.

\bibitem[Xu et~al.(2015)Xu, Ba, Kiros, Cho, Courville, Salakhudinov, Zemel, and Bengio]{xu2015show}
Xu, K., Ba, J., Kiros, R., Cho, K., Courville, A., Salakhudinov, R., Zemel, R., and Bengio, Y.
\newblock Show, attend and tell: Neural image caption generation with visual attention.
\newblock In \emph{International conference on machine learning}, pp.\  2048--2057. PMLR, 2015.

\bibitem[Young et~al.(2014)Young, Lai, Hodosh, and Hockenmaier]{young2014image}
Young, P., Lai, A., Hodosh, M., and Hockenmaier, J.
\newblock From image descriptions to visual denotations: New similarity metrics for semantic inference over event descriptions.
\newblock \emph{Transactions of the Association for Computational Linguistics}, 2:\penalty0 67--78, 2014.

\bibitem[Zhai et~al.(2019)Zhai, Puigcerver, Kolesnikov, Ruyssen, Riquelme, Lucic, Djolonga, Pinto, Neumann, Dosovitskiy, et~al.]{zhai2019visual}
Zhai, X., Puigcerver, J., Kolesnikov, A., Ruyssen, P., Riquelme, C., Lucic, M., Djolonga, J., Pinto, A.~S., Neumann, M., Dosovitskiy, A., et~al.
\newblock The visual task adaptation benchmark.
\newblock 2019.

\bibitem[Zhang et~al.(2023{\natexlab{a}})Zhang, Xu, Usuyama, Bagga, Tinn, Preston, Rao, Wei, Valluri, Wong, et~al.]{zhang2023large}
Zhang, S., Xu, Y., Usuyama, N., Bagga, J., Tinn, R., Preston, S., Rao, R., Wei, M., Valluri, N., Wong, C., et~al.
\newblock Large-scale domain-specific pretraining for biomedical vision-language processing.
\newblock \emph{arXiv preprint arXiv:2303.00915}, 2\penalty0 (3):\penalty0 6, 2023{\natexlab{a}}.

\bibitem[Zhang et~al.(2023{\natexlab{b}})Zhang, Xu, Usuyama, Xu, Bagga, Tinn, Preston, Rao, Wei, Valluri, et~al.]{zhang2023biomedclip}
Zhang, S., Xu, Y., Usuyama, N., Xu, H., Bagga, J., Tinn, R., Preston, S., Rao, R., Wei, M., Valluri, N., et~al.
\newblock Biomedclip: a multimodal biomedical foundation model pretrained from fifteen million scientific image-text pairs.
\newblock \emph{arXiv preprint arXiv:2303.00915}, 2023{\natexlab{b}}.

\bibitem[Zhu et~al.(2022)Zhu, Dong, and Liu]{zhu2022detecting}
Zhu, Z., Dong, Z., and Liu, Y.
\newblock Detecting corrupted labels without training a model to predict.
\newblock In \emph{International conference on machine learning}, pp.\  27412--27427. PMLR, 2022.

\bibitem[Zhu et~al.(2024)Zhu, Zhang, Wei, Wu, and Wu]{zhu2024vdc}
Zhu, Z., Zhang, M., Wei, S., Wu, B., and Wu, B.
\newblock Vdc: Versatile data cleanser based on visual-linguistic inconsistency by multimodal large language models.
\newblock In \emph{The Twelfth International Conference on Learning Representations}, 2024.

\end{thebibliography}
\bibliographystyle{icml2025}

\newpage
\appendix
\onecolumn

\counterwithin{figure}{section}
\counterwithin{table}{section}

\newpage
\section{Theoretical results}
\label{sec:proofs}
\newcommand{\norm}[1]{\left\lVert#1\right\rVert}

\subsection{Proof: Theorem 4.1}
\label{app:thm_proof}
Suppose that $\zeta_Y$ is distributed such that $\text{supp}(k\zeta_Y (1-p)) \subseteq \{0, 1, ..., k\}$.  For a correctly labeled sample $(X, Y)$, we have that $k \zeta_Y (1-p)$ of the neighbors are relevant and have correct labels, and so each contribute $d_{\mathcal{X}}(X, \bar{X})$ to $S_m(X,Y)$, and all remaining samples are either incorrectly labeled, or are not relevant to $Y$, and so each contribute $d_{\mathcal{X}}(X, X')$. Since $S_m(X, Y)$ is the sum of iid Gaussians, it is also a Gaussian, with:

\begin{align*}
    \E[S_m(X, Y)] &= \frac{1}{k} \left( \E[ \E[ d(X, \bar{X}_1) + ... + d(X, \bar{X}_{k \zeta_Y (1-p)}) | \zeta] ] +   \E[ \E[ d(X, X'_1) + ... + d(X, X'_{k - k\zeta_Y(1 - p)}) | \zeta] ] \right) \\
    &= \E[\zeta_Y] (1-p) \mu_2 + (1 - \E[\zeta_Y] (1-p)) \mu_1 \\
    &=  \E[\zeta_Y] (1-p) (\mu_2 - \mu_1) + \mu_1
\end{align*}

\begin{align*}
    \Var[S_m(X, Y)] &= \E[\Var(S_m(X, Y) | \zeta_Y)] + \Var(\E[S_m(X, Y) | \zeta_Y]) \\
    &= \E[\frac{1}{k^2}\Var\left( d(X, \bar{X}_1) + ... + d(X, \bar{X}_{k \zeta_Y (1-p)}) +  d(X, X'_1) + ... + d(X, X'_{k - k\zeta_Y(1 - p)}) | \zeta_Y \right)] \\ & \qquad + \Var(\E[S_m(X, Y) | \zeta_Y]) \\
    &= \E[\frac{1}{k}\left(\zeta_Y (1-p) \sigma_2^2 + (1 - \zeta_Y (1-p)) \sigma_1^2 \right)] + \Var(\zeta_Y (1-p) (\mu_2 - \mu_1) + \mu_1) \\
    &= \frac{1}{k} \left(\E[\zeta_Y](1-p) \sigma_2^2 + (1 - \E[\zeta_Y] (1-p)) \sigma_1^2 \right) + \Var(\zeta_Y) (1-p)^2 (\mu_2 - \mu_1)^2
\end{align*}

Similarly,
\begin{align*}
    S(X', Y') \sim  \mathcal{N}(\mu_1, \frac{\sigma_1^2}{k} )
\end{align*}
Putting it all together:
\begin{align*}
    \mathbb{P}( S_m(X', Y') - S_m(X, Y)  > 0) &= 1 - \Phi(\frac{-\mu}{\sigma})
\end{align*}
Where $\mu = \E[\zeta_Y] (1-p) (\mu_1 - \mu_2 ),\sigma = \sqrt{\frac{1}{k}\left(\E[\zeta_Y] (1-p) \sigma_2^2 + (2 - \E[\zeta_Y] (1-p) ) \sigma_1^2  \right) + \Var(\zeta_Y) (1-p)^2 (\mu_2 - \mu_1)^2} $, and $\Phi$ is the Gaussian CDF. Note that $\Var(\zeta_Y)$ is finite as $\zeta_Y$ is bounded by $[0, 1]$. \\ Setting $\mu > 0$ gives Lemma 4.3.

\subsection{Empirically Validating Assumption 2}
\label{app:normality_test}
\subsubsection{Classification at Dataset Level}
\label{app:assumption_2_1}

\begin{figure*}[t]
  \centering
  \begin{subfigure}{0.48\textwidth}
  \includegraphics[width=1.0\linewidth]{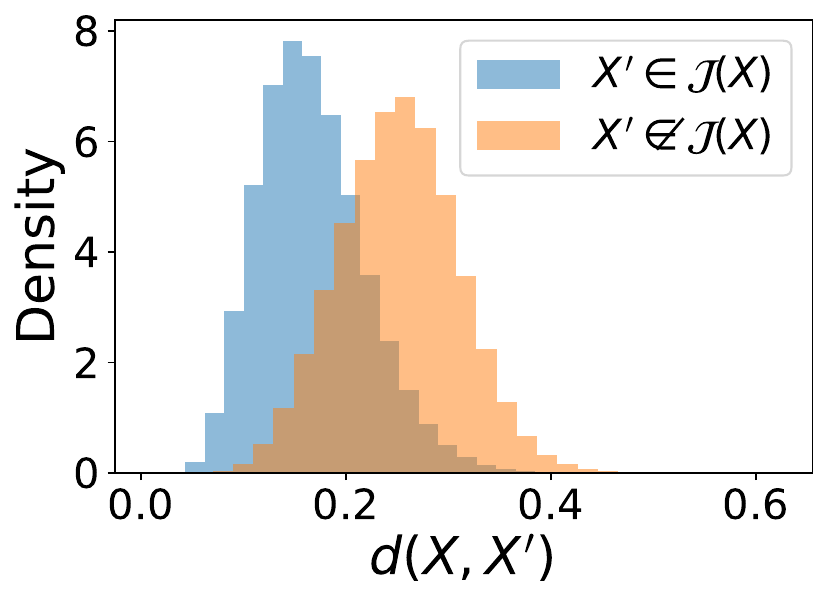}
    \caption{CIFAR-10} 
  \end{subfigure}%
     \hfill
   \begin{subfigure}{0.48\textwidth}
  \includegraphics[width=1.0\linewidth]{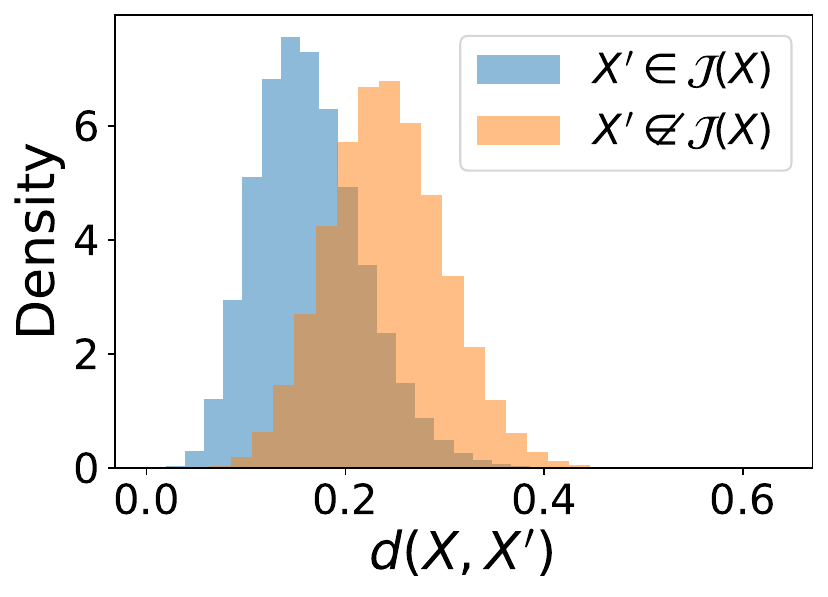}
    \caption{CIFAR-100} 
  \end{subfigure}%
  \caption{Histogram of cosine distances in the CLIP image embedding space}
  \label{fig:normal_dist_plot}
\end{figure*}

To empirically validate Assumption 2, we first utilize the training sets from the original CIFAR-10 and CIFAR-100 datasets. As these are classification datasets, we naturally define $\mathcal{J}$ as: ${x_2 \in \mathcal{J}(x_1) \iff y_1 = y_2}$, i.e. all images with the same label are paraphrases. We encode these images using the image encoder from OpenAI CLIP ViT-B/32 \citep{radford2021learning}, and utilize the cosine distance as $d_\mathcal{X}$. We compute pairwise distance between all 40,000 samples, and categorize these distances into either $x' \in J(x)$ or $x' \not\in J(x)$. We plot a histogram of these distances in Figure \ref{fig:normal_dist_plot}. Visually, both of these distributions appear to be normal, and we also observe that $\mu_1 > \mu_2$ from Lemma \ref{lemma:bad}. We then run a Shapiro–Wilk test on all four distributions to test for normality, randomly subsampling to 100 samples, as the Shapiro-Wilk test is not suitable for large sample sizes~\cite{ghasemi2012normality}. We find that in all four cases, the null hypothesis cannot be rejected ($p > 0.05$), and the test statistics are all greater than 0.97, indicating a high degree of normality.

\subsubsection{Captioning at Per-Sample Level}

\begin{figure*}[htbp]
  \centering
  \includegraphics[width=0.99\linewidth]{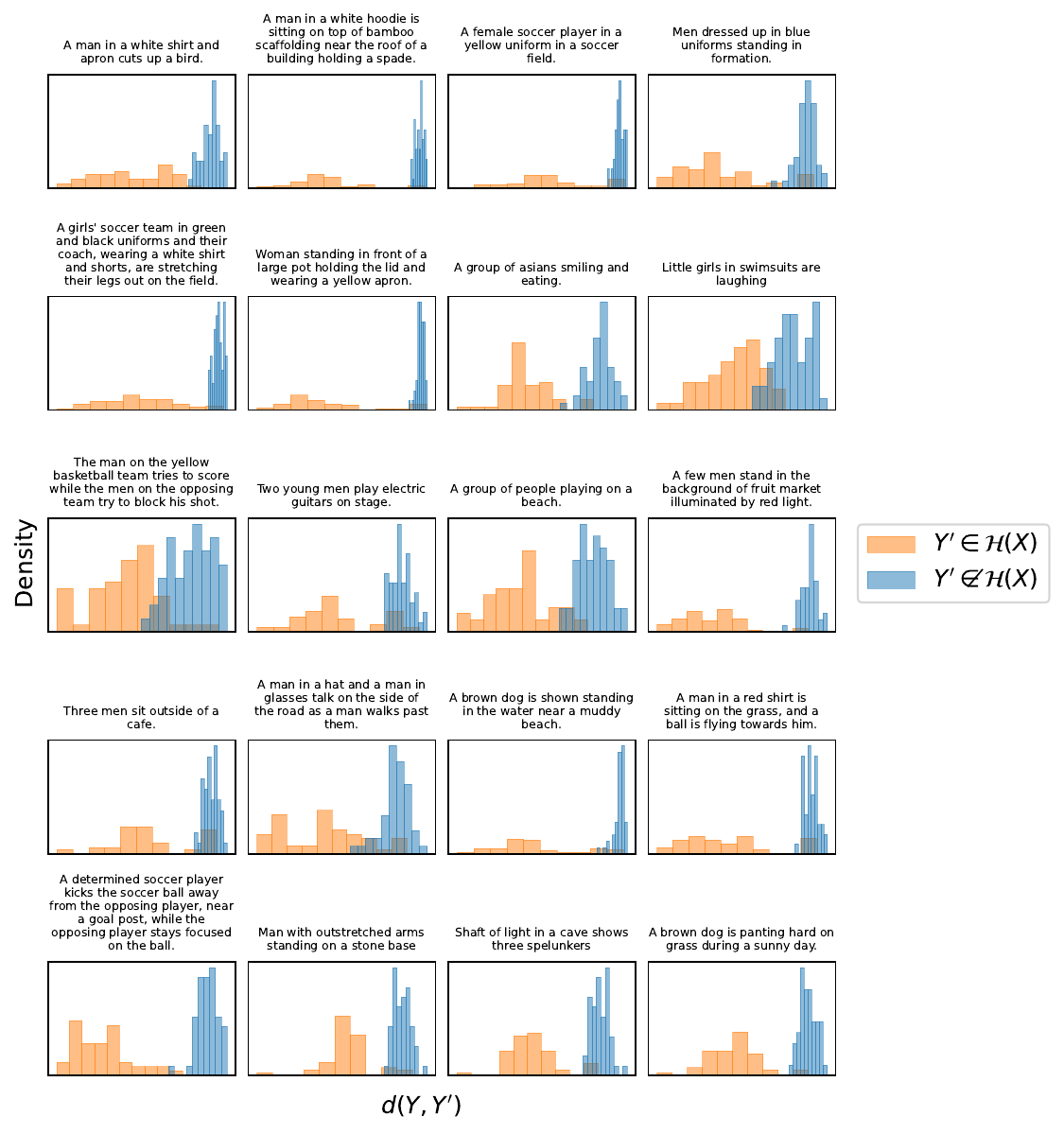}
  \caption{Histogram of log-cosine distances in the CLIP text embedding space}
  \label{fig:normality_real}
\end{figure*}

We verify Assumption 2 at a sample-wise level, for the analagous $\mathcal{H}(Y)$, by conducting an experiment on captions from \texttt{flickr30k}. We select 20 random captions, then prompt Llama 3.1-8B-instruct~\cite{grattafiori2024llama} to generate 50 paraphrasings of each caption (via sampling with temperature of 1), corresponding to 50 samples from  $\mathcal{H}(Y)$ for each caption. For the samples $Y' \not\in \mathcal{H}(Y)$, we randomly select 50 other captions from the dataset. To match the support of the Gaussian, we take the distance function to be the log cosine distance (note that this does not change the ordering of the score across samples). We compute this distance using the text encoder from OpenAI CLIP ViT-B/32~\cite{radford2021learning}, and plot histograms for each caption. We visualize our results in Figure \ref{fig:normality_real}. Running the same Shapiro-Wilk test from Section~\ref{app:assumption_2_1}, we find that of the positive samples, 8/20 are Gaussian, and of the negative samples, 16/20 are Gaussian. Thus, there is some evidence the Gaussianity assumption holds for natural language and complex paraphrase functions.

\subsection{Second Theorem}
\label{app:thm2}
We demonstrate that the embedding models trained via the contrastive multimodal objective are natural noisy label detectors.

\begin{theorem}[Contrastive Multimodal Embedding Models Detect Noisy Labels]
Let $\mathcal{Y} = \mathbb{R}$ and consider a training dataset $\mathcal{D}$. Suppose that $\hat{h}^{\mathcal{X}}_{\theta}: \mathcal{X} \rightarrow \mathbb{R}^d$ is an embedding function, and $\hat{h}^{\mathcal{Y}}_{\theta}: \mathcal{Y} \rightarrow \mathbb{R}^d$ is a Lipschitz continuous embedding function with constant $L_{\mathcal{Y}} > 0$, meaning that for all $y, y' \in \mathcal{Y}$,
$$
\left\| \hat{h}^{\mathcal{Y}}_{\theta}(y) - \hat{h}^{\mathcal{Y}}_{\theta}(y') \right\|_2 \leq L_{\mathcal{Y}} | y - y' |.
$$
For an input $x \in \mathcal{X}$ and its corresponding positive label $y \in \mathcal{Y}$, let $\eta$ be a random variable drawn from a normal distribution: $\eta \sim \mathcal{N}(0, \sigma^2).$
Define a noisy label $y' = y + \eta$. Let $d_{mm}(u, v) = ||u - v||_2$, which is proportional to $\sqrt{d_{cos}(u, v)}$ when $||u||_2 = ||v||_2 = 1$. Then, with probability at least $\delta(\epsilon) = 1 - 2 \Phi\left( -\dfrac{\epsilon}{\sigma} \right)$, where $\epsilon > 0$ and  $\Phi$ is the cumulative distribution function of the standard normal distribution, the following inequality holds:
$$
d_{mm}\left( \hat{h}^{\mathcal{X}}_{\theta}(x),\, \hat{h}^{\mathcal{Y}}_{\theta}(y') \right) \geq  d_{mm}\left( \hat{h}^{\mathcal{X}}_{\theta}(x),\, \hat{h}^{\mathcal{Y}}_{\theta}(y) \right) - L_{\mathcal{Y}}\, \epsilon.
$$

\end{theorem}

  When $L_{\mathcal{Y}} $ is small, this means that the score for the mislabeled sample cannot be much lower than the score for the positive pair with high probability.
Thus, we can see that multimodal embeddings are inherently capable of detecting mislabeled pairs, ensuring the distance between the embeddings of positive pairs is smaller than that of negative pairs. This motivates the use of $d_{mm}$ in \ours and in prior work \cite{kang2023noise, liang2023combating}.

\paragraph{Proof:} Since $\hat{h}^{\mathcal{Y}}_{\theta}$ is Lipschitz continuous with constant $L_{\mathcal{Y}}$, for any $y, y' \in \mathcal{Y}$, we have:
\begin{equation}
    \left\| \hat{h}^{\mathcal{Y}}_{\theta}(y') - \hat{h}^{\mathcal{Y}}_{\theta}(y) \right\|_2 \leq L_{\mathcal{Y}} | y' - y | = L_{\mathcal{Y}} | \eta | \label{eq:t41_proof_1}
\end{equation}

\noindent Let $d_{mm}(u, v) = ||u - v||_2$ be the Euclidean distance. Note that when $||u||_2 = ||v||_2 = 1$ (as in our experiments), we have that $||u-v||_2 =  \sqrt{2 (1 - u^T v)} = \sqrt{2 d_{cos}(u, v)}$, and so the two distances provide the same ordering of scores.
Applying the triangle inequality, we get:
$$
d_{mm}\left( \hat{h}^{\mathcal{X}}_{\theta}(x),\, \hat{h}^{\mathcal{Y}}_{\theta}(y') \right) \geq  d_{mm}\left( \hat{h}^{\mathcal{X}}_{\theta}(x),\, \hat{h}^{\mathcal{Y}}_{\theta}(y) \right) - \left\| \hat{h}^{\mathcal{Y}}_{\theta}(y) 
 - \hat{h}^{\mathcal{Y}}_{\theta}(y') \right\|_2 .
$$

When $| \eta | \leq \epsilon$, and substituting from Equation (\ref{eq:t41_proof_1}), it follows that:
$$
d_{mm}\left( \hat{h}^{\mathcal{X}}_{\theta}(x),\, \hat{h}^{\mathcal{Y}}_{\theta}(y') \right) \geq d_{mm}\left( \hat{h}^{\mathcal{X}}_{\theta}(x),\, \hat{h}^{\mathcal{Y}}_{\theta}(y) \right) - L_{\mathcal{Y}} \epsilon
$$

Since $\eta \sim \mathcal{N}(0, \sigma^2)$, the probability that $| \eta | \leq \epsilon$ is:
$$
P\left( | \eta | \leq \epsilon \right) = 1 - 2 \Phi\left( -\dfrac{\epsilon}{\sigma} \right) = \delta(\epsilon),
$$
where $\Phi$ is the cumulative distribution function of the standard normal distribution.

Thus, with probability at least $\delta(\epsilon)$, we have:
$$
d_{mm}\left( \hat{h}^{\mathcal{X}}_{\theta}(x),\, \hat{h}^{\mathcal{Y}}_{\theta}(y') \right) \geq d_{mm}\left( \hat{h}^{\mathcal{X}}_{\theta}(x),\, \hat{h}^{\mathcal{Y}}_{\theta}(y) \right) - L_{\mathcal{Y}} \epsilon
$$

When $L_{\mathcal{Y}} $ is small, this means that the score for the mislabeled sample cannot be much lower than the score for the positive pair with high probability. 

\section{Comparison with \citet{thomas2022emphasizing}}
\label{app:compare_with_baseline}
The goal of \citet{thomas2022emphasizing} to identify samples with semantic diversity, which is different from our goal of identifying mislabeled examples. As such, their proposed scores (i.e. $\Upsilon^{DIS}$ and $\Upsilon^{DIV}$) may not be effective in identifying mislabeled samples. As an example, consider the score $\Upsilon_Y^{DIS}$, which computes the similarity between the original caption, and the captions of its second-degree neighbors in text-space. Given a particular caption, e.g. ``This is a plane from the front view'' in Figure \ref{fig:method}, it could have second-degree neighbors in text-space that are semantically very similar to this caption (e.g. ``A plane facing the viewer''). However, only computing the distance of these captions in text space does not provide any signal for whether the \textit{image} is correctly paired to the caption. Similarly, the $\Upsilon^{DIV}$ scores also would not necessarily work, as the closeness of neighbors to each other in either modality do not provide a signal for whether the original sample is mislabeled.

However, the score from \citet{thomas2022emphasizing} that would intuitively provide a signal for mislabeling is $\Upsilon_{X}^{DIS}$, which computes second-degree neighbors in text space, then examines similarity between images. This is essentially the sum over $d_\mathcal{X}(\mathbf{x}, \mathbf{x}_{m_j})$ terms in our Equation (\ref{eq:sm}), but using second-degree neighbors instead of nearest neighbors. In addition, our Equation  (\ref{eq:sm}) contains two additional weighting terms (which we show improve label error performance in our ablation experiments). Finally, our proposed score contains the sum of two additional terms, which are not explored in \citet{thomas2022emphasizing}.

We compare the performance of our method against the $\Upsilon_{X}^{DIS}$ score in the main paper, and show performance of all four individual scores and two combined scores from \citet{thomas2022emphasizing} in Appendix \ref{app:r1_score_table}.

\section{\ours Algorithm}
\label{app:algo}
\begin{algorithm}[H]
\caption{\ours: \textbf{L}abel \textbf{E}rror Detection Using \textbf{M}ultim\textbf{o}dal \textbf{N}eighbors}
\label{alg:lemon}
\KwIn{
    Dataset $\mathcal{D} = \{ (\mathbf{x}_i, \mathbf{y}_i) \}_{i=1}^N$, Multimodal encoders $h^\mathcal{X}_\theta$, $h^\mathcal{Y}_\theta$, Distance functions $d_{\mathcal{X}}$, $d_{\mathcal{Y}}$ \\
    Hyperparameters: $k$, $\beta$, $\gamma$, $\tau_{1,n}$, $\tau_{2,n}$, $\tau_{1,m}$, $\tau_{2,m}$
}
\KwOut{
    Scores $\{ s_i \}_{i=1}^N$
}

Cache embeddings $h^\mathcal{X}_\theta(\mathbf{x}_i)$ and $h^\mathcal{Y}_\theta(\mathbf{y}_i)$ for $(\mathbf{x}_i, \mathbf{y}_i) \in \mathcal{D}$  \; 
Cache  
        $d_{mm}(\mathbf{x}_i, \mathbf{y}_i) = 1 - \frac{h^\mathcal{X}_\theta(\mathbf{x}_i) \cdot h^\mathcal{Y}_\theta(\mathbf{y}_i)}{\| h^\mathcal{X}_\theta(\mathbf{x}_i) \|_2 \| h^\mathcal{Y}_\theta(\mathbf{y}_i) \|_2}$
    for $(\mathbf{x}_i, \mathbf{y}_i) \in \mathcal{D}$  
\; 

\For{$i = 1$ \KwTo $N$}{
    Find indices $\{ n_j \}_{j=1}^k$ of $k$ nearest neighbors of $\mathbf{x}_i$ from $\mathcal{D} \setminus \{(\mathbf{x}_i, \mathbf{y}_i)\}$ using $d_{\mathcal{X}}$ \tcp*{$d_{\mathcal{X}}$ can use cached $h_{\theta}^{\mathcal{X}}$}

    Find indices $\{ m_j \}_{j=1}^k$ of $k$ nearest neighbors of $\mathbf{y}_i$ from $\mathcal{D} \setminus \{(\mathbf{x}_i, \mathbf{y}_i)\}$ using $d_{\mathcal{Y}}$ \tcp*{$d_{\mathcal{Y}}$ can use cached $h_{\theta}^{\mathcal{Y}}$}

    Compute $s_{n, i} := \frac{1}{k} \sum_{j=1}^k d_{\mathcal{Y}} (\mathbf{y}_i, \mathbf{y}_{n_j}) e^{-\tau_{1, n}  d_{\mathcal{X}} (\mathbf{x}_i, \mathbf{x}_{n_j}) } e^{-\tau_{2, n} d_{mm} (\mathbf{x}_{n_j},  \mathbf{y}_{n_j}) }$\;

    Compute $s_{m, i} := \frac{1}{k} \sum_{j=1}^k d_{\mathcal{X}} (\mathbf{x}_i, \mathbf{x}_{m_j}) e^{-\tau_{1, m}  d_{\mathcal{Y}} (\mathbf{y}_i, \mathbf{y}_{m_j}) } e^{-\tau_{2, m} d_{mm} (\mathbf{x}_{m_j},  \mathbf{y}_{m_j}) }$\;

    $s_i := d_{mm}(\mathbf{x}_i, \mathbf{y}_i) + \beta s_{n, i}  + \gamma s_{m, i}$
    
}
\Return $\mathbf{s}$\;

\end{algorithm}

For each image-caption pair in the dataset, we first compute how similar the image and caption are to each other using a pre-trained CLIP model ($d_{mm}$), which gives a basic measure of how well they match. To compute $s_m$, we compute the nearest neighbors of the caption among other captions in the dataset. For each neighbor, we look at how similar their corresponding image is to the original image. The intuition is that if a sample is correctly labeled, the image should be similar to images of other samples with similar captions. We weight each neighbor based on how close it is to our original sample and how well-matched the neighboring pairs themselves are. Finally, we repeat this for nearest neighbors in the image space to get $s_n$. \ours is then the weighted sum of these three scores.

\begin{table}[ht]
\centering
\caption{Notation and definitions used in Section \ref{sec:method}.}
\begin{tabular}{cl}
\toprule
\textbf{Symbol/Notation} & \textbf{Meaning} \\ \midrule
$\mathcal{D}$ & Dataset consisting of samples $(\mathbf{x}, \mathbf{y})_{i=1}^N$ \\ 
$\mathbf{x}, \mathcal{X}$ & First modality and its corresponding space (e.g., images) \\
$\mathbf{y}, \mathcal{Y}$ & Second modality and its corresponding space (e.g., text) \\ 
$f^*$ & Oracle function that assigns a binary mislabel indicator $z_i$ \\ 
$z_i$ & Mislabel indicator for sample $i$ ($z_i = 1$ if mislabeled, $z_i = 0$ otherwise) \\
$f(\mathbf{x}, \mathbf{y}) = s$ & Model output score \\
$d_{\mathcal{X}}, d_{\mathcal{Y}}$ & Distance functions in $\mathcal{X}$ and $\mathcal{Y}$ spaces \\
$B(\mathbf{x}, r)$ & Ball of radius $r$ centered at $\mathbf{x}$ in $\mathcal{X}$ space \\
$B(\mathbf{y}, r)$ & Ball of radius $r$ centered at $\mathbf{y}$ in $\mathcal{Y}$ space \\ 
$r_k(\mathbf{x})$ & Radius such that the ball $B(\mathbf{x}, r)$ contains at least $k$ neighbors \\ 
$\mathbf{x}_{n_j}$ & Nearest neighbor $j$ in $\mathcal{X}$ space \\ 
$\mathbf{y}_{m_j}$ & Nearest neighbor $j$ in $\mathcal{Y}$ space \\  
$h_\theta = (h^\mathcal{X}_\theta, h^\mathcal{Y}_\theta)$ & Multimodal encoder mapping $\mathcal{X}$ and $\mathcal{Y}$ to $\mathbb{R}^d$ \\ 
$d_{mm}(\mathbf{x}, \mathbf{y})$ & Multimodal distance between $\mathbf{x}$ and $\mathbf{y}$ \\
$s_n(\mathbf{x}, \mathbf{y}, \mathcal{D})$ & Score component based on $\mathbf{x}$'s neighbors, see Equation (\ref{eq:sn}). \\ 
$s_m(\mathbf{x}, \mathbf{y}, \mathcal{D})$ & Score component based on $\mathbf{y}$'s neighbors, see Equation (\ref{eq:sm}). \\ 
$\beta, \gamma$ & Hyperparameters weighting $s_n$ and $s_m$ \\ 
$\tau_{1, n}, \tau_{2, n}, \tau_{1, m}, \tau_{2, m}$ & Hyperparameters for weighting terms in $s_n$ and $s_m$ \\
$k$ & Number of nearest neighbors \\  \bottomrule
\end{tabular}
\label{tab:notations}
\end{table}

\section{Data Processing}
\label{sec:data_processing}
\subsection{Classification}
We utilize CIFAR10N (\texttt{cifar10}) and CIFAR100N (\texttt{cifar100}) object detection~\cite{zhu2022detecting}  datasets for all classification-based experiments. Each image is associated with a label indicating the primary object present in the image. These datasets contain 50,000 image-label pairs, with a clean and noisy label available per image. The noisy labels are examples of real human errors within the dataset. Further, we also generate synthetically noised labels as described in the main text. All images are resized to 224x224, center cropped, and normalized using mean and standard deviations corresponding to CLIP during the pre-processing stage. These two datasets are released under the Creative Commons Attribution-NonCommercial 4.0 license.  

For \texttt{miniImageNet} and \texttt{stanfordCars}, we use the ``red'' datasets from \citet{jiang2020beyond}, which contain noise from real-world web annotators. We split the full dataset (containing all annotations) into 75\%/12.5\%/12.5\% train/val/test sets, stratifying by the mislabel flag. The annotations are licensed by Google under CC BY 4.0 license, and the images are under CC BY 2.0 license. 

To generate the ``text'' modality for these classification datasets, we utilize the label name correspond to each class. For example, class 0 in \texttt{cifar10} is ``airplane'', and this is the caption associated we associate with all images of that class. In contrast to the caption-based datasets, there will be multiple k-nearest neighbors in the text modality with zero distance (i.e., with the same class label).

\subsection{Captioning}
We preprocess MSCOCO~\cite{lin2014microsoft}  and Flickr30k~\cite{young2014image} by using the Karpathy split~\cite{karpathy2015deep}, and then selecting one random annotation from the ones available. For the MMIMDB dataset~\cite{arevalo2017gated}, we utilize the plot outline as the text, and use the dataset splits provided. For MIMIC-CXR~\cite{johnson2019mimic}, we use all images in the database and the provided data splits, and extract the findings and impression sections from the radiology note for the text modality. Images were normalized and transformed using the same procedure described above.

For downstream captioning, we use the pre-trained tokenizer and image processor corresponding to the pre-trained model (GIT~\cite{wang2022git}) to pre-process image and captions.

Note that \texttt{flickr30k} is available under Flickr terms of use for non-commercial research and/or educational purposes\footnote{\url{https://shannon.cs.illinois.edu/DenotationGraph/}}.  \texttt{mscoco} is available under Creative Commons Attribution 4.0 License. \texttt{mmimdb} is available for personal and non-commercial use\footnote{\url{https://developer.imdb.com/non-commercial-datasets/}}. Finally, \texttt{mimiccxr} is available under the PhysioNet Credentialed Health Data License 1.5.0\footnote{\url{https://physionet.org/content/mimic-cxr/view-license/2.0.0/}}.

\section{Baseline Methods}
\label{sec:baseline_methods_app}

\subsection{Classification}
\subsubsection*{Training-dependent}

\textbf{AUM~\cite{pleiss2020identifying}}: This model assumes access to a classifier that can predict the class that an image likely belongs to. Then, the margin of difference between the prediction probability from the trained classifier for the assigned class and the class with the (next) highest probability is computed and averaged over training epochs. This score is thresholded to identify potential label errors.

\textbf{Datamap~\cite{swayamdipta2020dataset}}: Similar to AUM, this method requires access to a pretrained classifier. In this baseline, it is assumed that instances with label errors are `hard to learn', and thus low confidence in prediction throughout training epochs. To produce a single score, we combine the mean and standard deviation of the probability associated with the assigned class into a single score\footnote{We experimented with different strategies, and the square root of the product of the mean and (1-standard deviation) and (1-mean) and standard deviation led to comparable, high validation F1 scores.}. 

\textbf{Confident Learning} \cite{northcutt2021confident} is designed to identify labeling errors in classification datasets by modeling the relationship between true class labels and noisy ones. It sets thresholds for each true-noisy label pair. Using these thresholds, the model employs predicted class probabilities to rank predictions for each class, filtering out the noisy data. 

\subsubsection*{Training-free}

\textbf{CLIP Logits~\cite{liang2023combating}}: CLIP is used as a zero-shot classifier to obtain the softmax-based probability for the assigned class. This value is then thresholded to identify label errors. Recently, ~\cite{feng2024clipcleaner} used a similar zero-shot prediction jointly with a semi-supervised training approach for learning in the presence of label noise.

\textbf{CLIP Similarity~\cite{kang2023noise}}: The distance (either euclidean or cosine) between image and text embeddings from CLIP are computed and thresholded.

\textbf{Deep k-NN}\cite{bahri2020deep} The proportion of $k$ nearest neighbors\footnote{Note that this score can only take value in $\{0, 1/k, 2/k, ..., 1\}$.} with the same label is computed for each image of interest. Prior works have utilized different representations for obtaining neighbors, including logits and representations from pre-trained~\cite{zhu2022detecting} vision models. We find that pre-trained representations from CLIP outperformed logits from a zero-shot CLIP classifier~\cite{zhu2022detecting}.

\textbf{SimiFeat} \cite{zhu2022detecting} uses nearby features to detect noisy labels under the assumption that local groups of features share clean or noisy labels. \textbf{SimiFeat-V} \cite{zhu2022detecting} uses local voting and \textbf{SimiFeat-R} leverages ranking to detect noisy labels based on HOC estimator. The binary outputs produced are used for all score computations. Note that the difference between Simifeat-V and deep k-NN is in the data processing and augmentation.

\textbf{Discrepancy} \cite{thomas2022emphasizing} finds second-degree nearest neighbors in the text space, then computes the average distance of these neighbors to the original sample in image space. We utilize the same CLIP model to compute semantic distance here as in \ours.

Note that AUM and Datamap use the entire dataset as the reference set for label error detection, whereas SimiFeat uses both the train and test sets as reference sets, following the implementation in their original papers~\cite{pleiss2020identifying,swayamdipta2020dataset,zhu2022detecting}.

\subsection{Captioning}

\subsubsection*{Pre-trained or Supervised}

\textbf{LLaVA}~\cite{liu2024visual}: We prompt LLaVA (v1.6-vicuna-13b) with the following prompt: \texttt{The proposed caption for this image is "\{\}". Is this caption correct? Only answer with "Yes" or "No".} We examine the probability distribution over the first non-special token, and find the likelihood of the token with the highest probability. If the corresponding token in lower case starts with ``yes'', we return $1 - $ this probability as the mislabel score. Otherwise, we return the probability.

\textbf{VDC} \cite{zhu2024vdc}: As the original VDC paper does not explore the captioning setting, we make the following modifications to adapt it to our setup:
\begin{itemize}
    \item As we only utilize open-source models in our method, to preserve fairness, we also implement VDC with open-source models. Specifically, we use Llama-3.1-8B-Instruct~\cite{dubey2024llama} for the LLM in the Visual Question Generation (VQG) and Visual Answer Evaluation (VAE) stages (note that the VDC paper uses the OpenAI API), and InstructBLIP-Vicuna-7b~\cite{dai2023instructblip} as the VLLM in the Visual Question Answering (VQA) stage (as in the VDC paper).
    \item In the VQG stage, instead of generating specific questions for each class, we generate six specific questions for each caption. We slightly modify the VQG prompt (Table 8 in the VDC paper) to adapt it to captioning, and omit providing the label set, as the set of all possible captions is infinite. We keep the two general questions used in the VDC paper.
\end{itemize}

\textbf{CapFilt (oracle-like)}: We generate predictions using a pre-trained model trained to distinguish between high-quality MSCOCO and noisy synthetic captions~\cite{li2022blip}. This forms an oracle-like,  supervised baseline. We use a pre-trained checkpoint (pre-trained on the train split of MSCOCO)\footnote{\url{https://huggingface.co/Salesforce/blip-itm-base-coco}} for producing prediction scores.

\subsubsection*{Unsupervised}

\textbf{Datamap}: We compute the causal language modeling loss across training epochs and compute the product of the mean and variance in loss across epochs. That is, we expect captioning loss for instances with label errors to be consistently high. We train captioning models for 3 epochs, with LoRA rank set to 4, and a maximum length of corresponding to maximum model length\footnote{This is longer than captions in the train sets of all datasets except the medical dataset.} for the finetuning task. 

\textbf{Confident Learning}: We adapt this approach for dual-modality datasets, such as image-text pairs, by clustering text embeddings to serve as class labels for noise detection.

\subsubsection*{Downstream-task Unaware}

\textbf{Deep KNN}: We cluster captions similar to confident learning, adapting classification baseline.

\textbf{CLIP Similarity}: This is the same setup as classification.

\textbf{Discrepancy}: This is the same setup as classification.

\section{Compute Setup}
\label{sec:compute_setup}
We run our experiments on a shared Slurm cluster. Most experiments used one RTX A6000 with 48 GB VRAM, 10 CPU cores of Intel Xeon Ice Lake Platinum 8368, and 50 GB RAM.

\section{Hyperparameters in Label Error Detection}
The hyperparameters in each case were selected based on the validation set F1-score. Note that \oursfixed does not require hyperparameter tuning. Baseline code is included in the supplementary material. For SimiFeat-V and -R, we use the official open-sourced implementation directly.
\label{sec:hparams}
\subsection{Classification}

The search space for each method:
\begin{enumerate}
    \item AUM, Datamap: learning rate $\in \{5e-5, 5e-6\}$, training for epochs $\in \{5,10\}$\footnote{Note that we experiment with training for fewer epochs to avoid memorization, following~\cite{pleiss2020identifying}.}
    \item Confident learning: learning rate $\in \{5e-6, 5e-5\}$, upto 30 epochs with early stopping with a patience of 10.
    \item CLIP Sim.: cosine distance metric, no other hyperparameters
    \item CLIP Zero shot: distance metric
    \item Discrepancy: $k \in \{1,2 , 5, 10, 15, 20, 30, 50\}$
    \item deep k-NN: $k$, cosine distance metric
    \item Simifeat: we set $k=10$ following the original paper \cite{zhu2022detecting}.
\end{enumerate}

\subsection{Captioning}
For most baselines requiring a class index--obtained by clustering captions--we set the number of clusters to be 100.
\begin{enumerate}
    \item LLaVA: Small amount of prompt tuning. The optimal prompt selected was \texttt{The proposed caption for this image is ``\{\}''. Is this caption correct? Only answer with ``Yes'' or ``No''.'}
    \item Confident learning: learning rate $\in \{5e-6, 5e-5\}$, upto 30 epochs with early stopping with a patience of 10, number of clusters for captions\footnote{For \texttt{mimiccxr}, we use 10 clusters.}
     \item Discrepancy: $k \in \{1,2 , 5, 10, 15, 20, 30, 50\}$
    \item deep k-NN: $k \in \{1,2 , 5, 10, 15, 20, 30, 50\}$, number of text clusters
    \item VDC: no tunable hyperparameters
\end{enumerate}

\subsection{Our Method}
We search the following hyperparameters for our \oursopt:
\begin{enumerate}
    \item $k \in \{1,2 , 5, 10, 15, 20, 30, 50\}$
    \item Distance metric (either cosine or euclidean)
    \item $\beta, \gamma, \tau_{1, n}, \tau_{2, n}, \tau_{1, m}, \tau_{2, m}$: We take the hyperparameter set which achieves the best validation set F1 from these two strategies: (1) Using Scipy's \texttt{minimize} function, with initial guess $(1, 1, ..., 1)$, and with no explicit bounds. (2) Using a grid search with the following grid:
    \begin{itemize}
        \item $\beta \in \{0, 5, 10, 15, ..., 100\}$
         \item $\gamma \in \{0, 5, 10, 15, ..., 100\}$
         \item  $\tau_{1, n}, \tau_{2, n}, \tau_{1, m}, \tau_{2, m} \in \{0, 1, 5, 10\}$
    \end{itemize}
\end{enumerate}

\subsection{Optimal Hyperparameters}
Optimal hyperparameters for classification datasets can be found in Table \ref{tab:optim_hparams_clf}, and optimal hyperparameters for captioning datasets can be found in Table \ref{tab:optim_hparams_cap}.

\begin{table}[!htbp]
\caption{Optimal hyperparameters for methods shown in Table \ref{tab:clf_label_error_detection_full}. Note that Simifeat, VDC, CLIP Sim., and \oursfixed have no tunable hyperparameters.}
\centering
\begin{tabular}{@{}lllll@{}}
\toprule
                     & \texttt{cifar10}                                                                        & \texttt{cifar100}                                                                       & \texttt{miniImageNet}                                                                  & \texttt{stanfordCars}                                                                  \\ \midrule
\textbf{AUM}         & \begin{tabular}[c]{@{}l@{}}LR = 5E-6\\ Epochs = 5\end{tabular}                 & \begin{tabular}[c]{@{}l@{}}LR = 5E-5\\ Epochs = 5\end{tabular}                 & \begin{tabular}[c]{@{}l@{}}LR = 5E-6\\ Epochs = 10\end{tabular}               & \begin{tabular}[c]{@{}l@{}}LR = 5E-5\\ Epochs = 10\end{tabular}               \\ \midrule
\textbf{Datamap}     & \begin{tabular}[c]{@{}l@{}}LR = 5E-6\\ Epochs = 5\end{tabular}                 & \begin{tabular}[c]{@{}l@{}}LR = 5E-5\\ Epochs = 5\end{tabular}                 & \begin{tabular}[c]{@{}l@{}}LR = 5E-5\\ Epochs = 10\end{tabular}                & \begin{tabular}[c]{@{}l@{}}LR = 5E-5\\ Epochs = 10\end{tabular}               \\ \midrule
\textbf{Confident}   & \begin{tabular}[c]{@{}l@{}}LR=5e-06\\ Epochs=30 \\ Batch size=128\end{tabular} & \begin{tabular}[c]{@{}l@{}}LR=5e-06\\ Epochs=30 \\ Batch size=128\end{tabular} & \begin{tabular}[c]{@{}l@{}}LR=5e-05\\ Epochs=30\\ Batch size=128\end{tabular} & \begin{tabular}[c]{@{}l@{}}LR=5e-05\\ Epochs=30\\ Batch size=128\end{tabular} \\ \midrule
\textbf{CLIP Logits} & Cosine distance                                                                & Cosine distance                                                                & Cosine distance                                                               & Cosine distance                                                               \\ \midrule
\textbf{Discrepancy} & k=20                                                                           & k=50                                                                           & k=30                                                                          & k=20                                                                          \\ \midrule
\textbf{Deep k-NN}   & \begin{tabular}[c]{@{}l@{}}k=50\\ cosine distance\end{tabular}                 & \begin{tabular}[c]{@{}l@{}}k=20\\ cosine distance\end{tabular}                 & \begin{tabular}[c]{@{}l@{}}k=50\\ cosine distance\end{tabular}                & \begin{tabular}[c]{@{}l@{}}k=30\\ cosine distance\end{tabular}           \\ \midrule
\textbf{\oursopt}  &        \begin{tabular}[c]{@{}l@{}}k=50\\  cosine distance\\ $\beta=20$ \\ $\gamma = 35$ \\ $\tau_{1, n} = 0$ \\ $\tau_{2,n} = 5$ \\ $\tau_{1, m} = 0$ \\ $\tau_{2,m} = 5$\end{tabular}                                                                        &      \begin{tabular}[c]{@{}l@{}}k=20\\ cosine distance\\ $\beta=2.14$ \\ $\gamma = -0.024$ \\ $\tau_{1, n} = -1.71$ \\ $\tau_{2,n} = 4.85$ \\ $\tau_{1, m} = -0.068$ \\ $\tau_{2,m} = -0.019$\end{tabular}                                                                            &  \begin{tabular}[c]{@{}l@{}}
k=50\\ 
Euclidean distance\\ 
$\beta=0.664$ \\ 
$\gamma=0.395$ \\ 
$\tau_{1, n}=1.91$ \\ 
$\tau_{2, n}=1.04$ \\ 
$\tau_{1, m}=1.00$ \\ 
$\tau_{2, m}=1.35$
\end{tabular}                                                                             &     \begin{tabular}[c]{@{}l@{}}k=15\\ Euclidean distance\\ $\beta=0.631$ \\ $\gamma = 0.431$ \\ $\tau_{1, n} = 0.898$ \\ $\tau_{2,n} = -0.192$ \\ $\tau_{1, m} = 0.0$ \\ $\tau_{2,m} = -0.001$\end{tabular}
                                                                          \\ \bottomrule
\end{tabular}
\label{tab:optim_hparams_clf}
\end{table}

\begin{table}[!htbp]
\caption{Optimal hyperparameters for methods shown in Table \ref{tab:captioning}. Note that  LLaVA, VDC, CLIP Sim. and \oursfixed have no tunable hyperparameters.}
\centering
\begin{tabular}{@{}lllll@{}}
\toprule
                     & \texttt{flickr30k}                                                                                       & \texttt{mscoco}                                                                                           & \texttt{mmimdb}                                                                                         & \texttt{mimiccxr}                                                                                      \\ \midrule
\textbf{Datamap}     & \begin{tabular}[c]{@{}l@{}}Batch size = 16\\ Epochs = 3\\ LoRA rank = 4\end{tabular}            & \begin{tabular}[c]{@{}l@{}}Batch size = 16\\ Epochs = 3\\ LoRA rank = 4\end{tabular}             & \begin{tabular}[c]{@{}l@{}}Batch size = 16\\ Epochs = 3\\ LoRA rank = 4\end{tabular}           & \begin{tabular}[c]{@{}l@{}}Batch size = 16\\ Epochs = 3\\ LoRA rank = 4\end{tabular}          \\ \midrule
\textbf{Discrepancy} &      k=5                                                                                           &      k=10                                                                                            &      k=10                                                                                          &      k=10                                                                                           \\ \midrule
\textbf{Deep k-NN}   & \begin{tabular}[c]{@{}l@{}}k=50 \\  n\_cluster=100\end{tabular}                                & \begin{tabular}[c]{@{}l@{}}k=50\\ n\_cluster=100\end{tabular}                                   & \begin{tabular}[c]{@{}l@{}}k=20\\ n\_cluster=100\end{tabular}                                 & \begin{tabular}[c]{@{}l@{}}k=50\\ n\_cluster=100\end{tabular}                                \\ \midrule
\textbf{Confident}   & \begin{tabular}[c]{@{}l@{}}LR=5e-06\\ Epochs=30\\ Batch size=128\\ n\_cluster=10\end{tabular} & \begin{tabular}[c]{@{}l@{}}LR=5e-06\\ Epochs=30 \\ Batch size=128, \\ n\_cluster=100\end{tabular} & \begin{tabular}[c]{@{}l@{}}LR=5e-05\\ Epochs=30 \\ Batch size=128\\ n\_cluster=10\end{tabular} & \begin{tabular}[c]{@{}l@{}}LR=5e-06\\ Epochs=30 \\ Batch size=16\\ n\_cluster=10\end{tabular} \\ \midrule
\textbf{\oursopt}  &    \begin{tabular}[c]{@{}l@{}}k=30\\ cosine distance\\ $\beta=0.092$ \\ $\gamma = 0.177$ \\ $\tau_{1, n} = 0.274$ \\ $\tau_{2,n} = 0.074$ \\ $\tau_{1, m} = 0.072$ \\ $\tau_{2,m} = 0.0$\end{tabular}                                                                                              &   \begin{tabular}[c]{@{}l@{}}k=30\\ cosine distance\\ $\beta=5.324$ \\ $\gamma = 11.057$ \\ $\tau_{1, n} = 5.143$ \\ $\tau_{2,n} = 10.498$ \\ $\tau_{1, m} = 7.233$ \\ $\tau_{2,m} = 15.637$\end{tabular}                                                                                                  &      \begin{tabular}[c]{@{}l@{}}k=10\\ Euclidean distance\\ $\beta=1.001$ \\ $\gamma = 1.202$ \\ $\tau_{1, n} = 0.983$ \\ $\tau_{2,n} = 1.000$ \\ $\tau_{1, m} = 4.450$ \\ $\tau_{2,m} = 1.080$\end{tabular}                                                                                                          &   \begin{tabular}[c]{@{}l@{}}k=30\\ cosine distance\\ $\beta=5$ \\ $\gamma = 10$ \\ $\tau_{1, n} = 5$ \\ $\tau_{2,n} = 10$ \\ $\tau_{1, m} = 5$ \\ $\tau_{2,m} = 10$\end{tabular}                                                                                               \\ \bottomrule
\end{tabular}
\label{tab:optim_hparams_cap}
\end{table}

\section{Hyperparameters in Downstream Models}
\subsection{Classification}
We train a Vision Transformer (ViT)-based image classification ~\cite{dosovitskiy2020image}\footnote{\url{https://huggingface.co/google/vit-base-patch16-224}} model pre-trained on ImageNet-21k~\cite{ridnikimagenet} and fine-tuned on ImageNet 2012~\cite{russakovsky2015imagenet} with an additional linear layer. We add a linear layer above the last hidden state. We train for up to 30 epochs with an initial learning rate of and early stopping with a patience of 4. We tune learning rates on the fully noisy set, for learning rate in $\in \{5e-4, 1e-3, 1e-5\}$ (with cosine annealing learning rate scheduling).

\subsection{Captioning}

The hyperparameter tuning grid for the captioning model\footnote{\url{https://huggingface.co/microsoft/git-base}} are: learning rate of $1e-4$, batch size: 16, maximum number of epochs: 10. The model checkpoint from the epoch with the lowest validation loss is used for caption generation at test time. For text generation, we use beam search with 4 beams, following~\citet{wang2022git}. We use the AdamW optimizer~\cite{loshchilov2018decoupled}, with cosine scheduling for learning rate with 1000 warmup steps. We tune LoRA rank in \{4,16\} based on BLEU-4 scores on the validation set.  We also experimented with a lower learning rate of $1e-5$ on the fully noisy set, and observed higher performance (in terms of validation BLEU-4) with a higher learning rate of $1e-4$.

Here, for finding best epoch during training, we assume that validation loss is correlated with language model quality. We leave strategies such as self-critical training validation~\cite{rennie2017self} to future work. We verified that retaining the model from the last epoch leads to similar trends in results -- see Table~\ref{tab:captioning_downstream_results_extra} for models trained up to 10 epochs. We find that both \ours and the baseline perform within 1 point of each other on both datasets (similar to the results with loss-based checkpointing), and filtering out noisy data improves downstream captioning performance.

\section{Additional Experimental Results}
\label{app:add_results}

\subsection{Label Error Detection in Classification Settings}
\label{app:clf_label_error_detection.}
Full results on classification datasets using the noise types bolded in Table \ref{tab:datasets} (including AUPRC) can be found in Table \ref{tab:clf_label_error_detection_full}.

The performance of all baselines and our method on the two types of synthetic errors are shown in Table~\ref{tab:label_error_synth_clf}, all at a noise level of 40\% (comparable to the amount of error in the noisy CIFAR datasets).

\begin{table}[!tb]\centering
\caption{Label error detection performance across classification datasets, for the bolded noise types in Table \ref{tab:datasets}. We separate AUM, Datamap, and Confident learning, as they require training a classifier from scratch. Bold denotes best score within each training approach. }
\resizebox{0.8\textwidth}{!}{  
\begin{tabular}{lrcrrrrr}\toprule
\textbf{Dataset} &\textbf{Method} &\textbf{Training-free} &\textbf{AUROC (\%)} &\textbf{AUPRC (\%)} &\textbf{F1 (\%)} \\\midrule
\multirow{10}{*}{\texttt{cifar10}} &AUM & \multirow{3}{*}{\ding{55}} &\textbf{98.3} (0.1) &\textbf{97.9} (0.1) &\textbf{92.9} (0.1)\\
&Datamap && 98.2 (0.1) &97.6 (0.1) &92.2 (0.5)\\
&Confident && 89.6 (1.4) &86.1 (1.8) &88.2 (1.7)\\
\cmidrule{2-8}
&CLIP Logits & \multirow{7}{*}{\ding{51}} &95.5 (0.2) &93.9 (0.3) &86.2 (0.6)\\
&CLIP Sim. && 92.2 (0.2) & 90.2 (0.4) & 82.3 (0.3)\\
&Simifeat-V && 90.9 (0.1) &88.3 (0.4) &88.4 (0.5)\\
&Simifeat-R && 90.7 (0.2) &87.9 (0.4) &88.2 (0.3)\\
&Discrepancy && 77.1 (1.9) &70.4 (2.7) &66.4 (2.2)\\
&Deep k-NN && 97.8 (0.1) &96.5 (0.2) &91.4 (0.6)\\
&\oursfixed (Ours) && 97.7 (0.2) &96.8 (0.3) &90.9 (0.1)\\
&\oursopt (Ours) && \textbf{98.1} (0.0) &\textbf{97.4} (0.1) &\textbf{92.0} (0.2)\\
\midrule
\multirow{10}{*}{\texttt{cifar100}} &AUM & \multirow{3}{*}{\ding{55}} &\textbf{92.3} (0.3) &\textbf{90.0} (0.5) &\textbf{81.1} (0.3)\\
&Datamap && 91.8 (0.3) &89.5 (0.5) &80.8 (0.5)\\
&Confident && 78.6 (0.4) &68.8 (0.9) &73.7 (0.5)\\
\cmidrule{2-8}
&CLIP Logits & \multirow{7}{*}{\ding{51}} &84.9 (0.7) &80.3 (1.2) &72.0 (0.9)\\
&CLIP Sim. && 80.8 (0.9) & 75.2 (1.3) & 68.7 (1.1)\\
&Simifeat-V && 79.6 (0.2) &71.1 (0.5) &73.3 (0.3)\\
&Simifeat-R && 79.6 (0.2) &71.1 (0.5) &73.3 (0.3)\\
&Discrepancy && 66.0 (1.5) &57.4 (2.3) &58.9 (0.8)\\
&Deep k-NN && 87.4 (0.3) &77.9 (1.0) &75.7 (0.3)\\
&\oursfixed (Ours) && 88.9 (0.7) &84.6 (1.1) &75.4 (0.6)\\
&\oursopt (Ours) && \textbf{90.8} (0.0) &\textbf{87.4} (0.3) &\textbf{78.4} (0.0)\\
\midrule
\multirow{10}{*}{\texttt{miniImageNet}} &AUM & \multirow{3}{*}{\ding{55}} &83.1 (0.2) &73.2 (0.5) &68.3 (0.4)\\
&Datamap && \textbf{85.1} (0.3) &\textbf{70.1} (0.8) &\textbf{70.6} (0.2)\\
&Confident && 59.5 (0.7) &42.0 (0.8) &37.7 (1.5)\\
\cmidrule{2-8}
&CLIP Logits & \multirow{7}{*}{\ding{51}} &\textbf{90.0} (0.2) &\textbf{80.9} (0.5) &\textbf{77.1} (0.2)\\
&CLIP Sim. && 89.3 (0.2) &80.7 (0.3) &76.1 (0.4)\\
&Simifeat-V && 68.2 (0.3)	& 53.2 (0.6)	& 55.1 (0.6)

\\
&Simifeat-R && 68.1 (0.2)	&53.0 (0.2)	&54.8 (0.5)
\\

&Discrepancy && 79.5 (0.2) &65.9 (0.4) &64.0 (0.6)\\
&Deep k-NN && 83.2 (0.2) &70.9 (0.6) &75.2 (0.4)\\
&\oursfixed (Ours) && 89.5 (0.2) &81.5 (0.3) &74.7 (0.2)\\
&\oursopt (Ours) && \textbf{90.0} (0.4) &79.7 (3.1) &76.9 (0.2)\\
\midrule
\multirow{10}{*}{\texttt{stanfordCars}} &AUM & \multirow{3}{*}{\ding{55}} &70.4 (2.3) &\textbf{42.0} (1.0) &47.2 (3.1)\\
&Datamap && \textbf{72.2} (1.7) &39.5 (0.2) &\textbf{50.4} (2.1)\\
&Confident && 60.7 (0.3) &29.9 (0.2) &39.9 (0.6)\\
\cmidrule{2-8}
&CLIP Logits & \multirow{7}{*}{\ding{51}} &68.8 (0.1) &39.7 (0.9) &47.3 (0.5)\\
&CLIP Sim. && 69.8 (0.4) &40.7 (1.0) &46.6 (0.5)\\
&Simifeat-V && 63.4 (1.3) &	33.2 (1.6)	& 43.3 (1.5)

\\
&Simifeat-R && 63.6 (1.2) &	33.4 (1.1) &	43.5 (1.6)

\\
&Discrepancy && 65.7 (0.3) &33.3 (0.6) &44.3 (0.7)\\
&Deep k-NN && 71.5 (0.6) &41.7 (0.7) &49.1 (0.6)\\
&\oursfixed (Ours) && 72.6 (0.7) &\textbf{44.9} (1.4) &47.7 (2.0)\\
&\oursopt (Ours) && \textbf{73.1} (0.5) &40.5 (0.4) &\textbf{51.3} (0.5)\\
\bottomrule
\end{tabular}
}
\label{tab:clf_label_error_detection_full}
\end{table}

\begin{table}[!htp]\centering
\scriptsize
\caption{Label error detection performance on synthetic errors for classification datasets}
\begin{tabular}{lrrrrrrrr}\toprule
& & &\multicolumn{2}{c}{\textbf{AUROC}} &\multicolumn{2}{c}{\textbf{AUPRC}} &\multicolumn{2}{c}{\textbf{F1}} \\\cmidrule{4-9}
& & &\textbf{mean} &\textbf{std} &\textbf{mean} &\textbf{std} &\textbf{mean} &\textbf{std} \\\cmidrule{4-9}
\textbf{Dataset} &\textbf{Flip Type} &\textbf{Method} & & & & & & \\\midrule
\multirow{22}{*}{\texttt{cifar10}} & \multirow{11}{*}{\textbf{asymmetric}} & \textbf{AUM} & \textbf{93.7} & 0.6 & \textbf{86.6} & 0.6 & \textbf{86.6} & 1.1 \\
 &  & \textbf{Datamap} & 93.6 & 0.4 & 86.3 & 0.8 & 86.0 & 0.9 \\
 &  & \textbf{Confident} & 87.3 & 4.7 & 78.0 & 7.3 & 84.7 & 5.7 \\
  \cdashlinelr{3-9}
 &  & \textbf{CLIP Logits} & \textbf{98.8} & 0.2 & \textbf{97.9} & 0.3 & 93.2 & 0.4 \\
 &  & \textbf{CLIP Sim.} & 97.0 & 0.2 & 95.3 & 0.1 & 89.4 & 0.5 \\
 &  & \textbf{Simifeat-V} & 69.8 & 0.5 & 57.7 & 0.7 & 60.5 & 0.7 \\
 &  & \textbf{Simifeat-R} & 70.2 & 0.7 & 58.6 & 1.0 & 61.0 & 1.0 \\
 &  & \textbf{Discrepancy} & 63.2 & 3.3 & 51.3 & 1.4 & 58.5 & 1.2 \\
 &  & \textbf{Deep k-NN} & 85.2 & 0.7 & 66.2 & 0.9 & 79.5 & 0.8 \\
 &  & \textbf{\oursfixed} & 97.5 & 0.2 & 94.8 & 0.6 & 90.2 & 0.8 \\
 &  & \textbf{\oursopt} & \textbf{98.8} & 0.2 & 97.8 & 0.5 & \textbf{93.9} & 0.3 \\
\cmidrule{2-9}
 & \multirow{11}{*}{\textbf{symmetric}} & \textbf{AUM} & \textbf{99.8} & 0.0 & \textbf{99.7} & 0.0 & \textbf{98.1} & 0.3 \\
 &  & \textbf{Datamap} & \textbf{99.8} & 0.0 & \textbf{99.7} & 0.0 & 98.0 & 0.2 \\
 &  & \textbf{Confident} & 97.5 & 0.2 & 94.8 & 0.7 & 96.9 & 0.4 \\
   \cdashlinelr{3-9}
 &  & \textbf{CLIP Logits} & 98.5 & 0.0 & 97.9 & 0.1 & 92.2 & 0.1 \\
 &  & \textbf{CLIP Sim.} & 97.1 & 0.1 & 95.9 & 0.2 & 89.5 & 0.1 \\
 &  & \textbf{Simifeat-V} & 96.5 & 0.0 & 93.9 & 0.2 & 94.2 & 0.3 \\
 &  & \textbf{Simifeat-R} & 96.4 & 0.2 & 93.7 & 0.4 & 94.2 & 0.3 \\
 &  & \textbf{Discrepancy} & 84.5 & 2.3 & 78.2 & 3.0 & 73.2 & 1.1 \\
 &  & \textbf{Deep k-NN} & 99.2 & 0.1 & 98.1 & 0.2 & 96.1 & 0.3 \\
 &  & \textbf{\oursfixed} & 99.5 & 0.1 & 99.2 & 0.1 & 95.9 & 0.3 \\
 &  & \textbf{\oursopt} & \textbf{99.6} & 0.1 & \textbf{99.4} & 0.1 & \textbf{96.8} & 0.2 \\
\midrule
\multirow{22}{*}{\texttt{cifar100}} & \multirow{11}{*}{\textbf{asymmetric}} & \textbf{AUM} & \textbf{82.4} & 2.0 & \textbf{67.5} & 2.6 & \textbf{75.2} & 1.5 \\
 &  & \textbf{Datamap} & 74.4 & 1.0 & 59.3 & 1.2 & 66.1 & 1.5 \\
 &  & \textbf{Confident} & 71.7 & 2.5 & 56.6 & 3.0 & 67.3 & 2.7 \\
   \cdashlinelr{3-9}
 &  & \textbf{CLIP Logits} & 96.6 & 0.3 & 94.8 & 0.5 & 88.3 & 0.7 \\
 &  & \textbf{CLIP Sim.} & 95.1 & 0.4 & 93.1 & 0.5 & 85.6 & 0.6 \\
 &  & \textbf{Simifeat-V} & 65.4 & 1.4 & 52.4 & 1.7 & 57.1 & 1.9 \\
 &  & \textbf{Simifeat-R} & 65.4 & 1.2 & 52.8 & 1.6 & 57.0 & 1.7 \\
 &  & \textbf{Discrepancy} & 62.0 & 0.6 & 50.8 & 1.2 & 58.1 & 0.9 \\
 &  & \textbf{Deep k-NN} & 62.6 & 0.3 & 47.7 & 0.8 & 63.2 & 0.5 \\
 &  & \textbf{\oursfixed} & 94.9 & 0.3 & 92.1 & 0.4 & 84.3 & 0.4 \\
 &  & \textbf{\oursopt} & \textbf{96.7} & 0.3 & \textbf{95.3} & 0.4 & \textbf{88.4} & 0.5 \\
\cmidrule{2-9}
 & \multirow{11}{*}{\textbf{symmetric}} & \textbf{AUM} & \textbf{99.2} & 0.3 & \textbf{98.9} & 0.4 & \textbf{95.2} & 1.1 \\
 &  & \textbf{Datamap} & \textbf{99.2} & 0.4 & 98.8 & 0.7 & 95.0 & 1.5 \\
 &  & \textbf{Confident} & 87.8 & 0.6 & 77.8 & 1.0 & 85.2 & 0.5 \\
   \cdashlinelr{3-9}
 &  & \textbf{CLIP Logits} & 96.8 & 0.1 & 95.2 & 0.3 & 89.1 & 0.4 \\
 &  & \textbf{CLIP Sim.} & 95.5 & 0.2 & 93.5 & 0.3 & 86.6 & 0.9 \\
 &  & \textbf{Simifeat-V} & 91.1 & 0.5 & 84.9 & 1.0 & 84.5 & 0.9 \\
 &  & \textbf{Simifeat-R} & 90.9 & 0.6 & 84.4 & 1.7 & 84.6 & 0.8 \\
 &  & \textbf{Discrepancy} & 82.8 & 0.9 & 75.8 & 1.0 & 70.7 & 0.5 \\
 &  & \textbf{Deep k-NN} & 96.7 & 0.1 & 91.7 & 0.3 & 91.0 & 0.4 \\
 &  & \textbf{\oursfixed} & 98.4 & 0.1 & 97.7 & 0.2 & 92.0 & 0.1 \\
 &  & \textbf{\oursopt} & \textbf{99.0} & 0.0 & \textbf{98.7} & 0.1 & \textbf{94.1} & 0.1 \\
\bottomrule
\end{tabular}
\label{tab:label_error_synth_clf}
\end{table}

\subsection{Label Error Detection in Captioning Settings}
\label{sec:captioning_synth}
Full results on classification datasets using the noise types bolded in Table \ref{tab:datasets} (including AUPRC) can be found in Table \ref{tab:captioning_full}.

Results on the remaining synthetic noise types (at 40\%) can be found in: \texttt{flickr30k}: Table~\ref{tab:flickr_label_error}, \texttt{mscoco}: Table~\ref{tab:mscoco_label_error}, \texttt{mmimdb}: Table~\ref{tab:mmimdb_label_error}, and \texttt{mimic-cxr}: Table~\ref{tab:mimiccxr_label_error}.
Across all datasets and noising types, we find that our model outperforms other non-oracle/supervised baselines.

\begin{table}[!htbp]\centering
\caption{Label error detection performance on captioning datasets, for the bolded noise types in Table \ref{tab:datasets}. }
\resizebox{0.7\textwidth}{!}{  
\begin{tabular}{lrrrrr}\toprule
\textbf{Dataset} &\textbf{Method} &\textbf{AUROC (\%)} &\textbf{AUPRC (\%)} &\textbf{F1 (\%)}  \\\midrule
\multirow{8}{*}{\texttt{flickr30k}} &LLaVA & 79.3 (0.8) &58.5 (0.2) &65.0 (1.1)  \\
 &Datamap & 52.7 (1.5) &37.9 (1.4) &50.4 (1.8)  \\
 &Discrepancy & 73.0 (0.6) &59.2 (1.8) &59.0 (0.3)  \\
 &VDC & 92.9 (1.1) &87.2 (0.3) &81.1 (1.6)  \\
 &Deep k-NN & 71.1 (0.4) &52.0 (1.0) &59.2 (0.8)  \\
 &Confident & 63.1 (0.9) &42.1 (1.2) &54.0 (0.9)  \\
 &CLIP Sim. & \textbf{94.8} (0.5) &\textbf{92.8} (0.5) &\textbf{84.2} (0.9)  \\
 &\oursfixed (Ours) & 93.6 (0.2) &92.0 (0.2) &-  \\
 &\oursopt (Ours) & 94.5 (0.2) &\textbf{92.8} (0.3) &83.6 (1.4)  \\
 \cdashlinelr{2-6}
 &CapFilt (Oracle) &98.6 (0.1) &98.1 (0.1) &93.1 (0.7) \\
\midrule
\multirow{8}{*}{\texttt{mscoco}} &LLaVA & 80.3 (0.1) &63.4 (0.3) &74.9 (0.3)  \\
 &Datamap & 68.9 (0.8) &60.3 (0.0) &60.3 (1.2)  \\
 &Discrepancy & 72.7 (0.3) &67.2 (0.4) &62.5 (0.3)  \\
 &VDC & 94.1 (0.2) &91.8 (0.2) &86.3 (0.4)  \\
 &Deep k-NN & 76.6 (0.4) &70.3 (0.6) &67.5 (0.8)  \\
 &Confident & 71.5 (0.5) &56.4 (0.5) &66.5 (0.5)  \\
 &CLIP Sim. & 93.8 (0.2) &93.0 (0.4) &84.5 (0.4)  \\
 &\oursfixed (Ours) & 92.0 (0.1) &91.8 (0.3) &-  \\
 &\oursopt (Ours) & \textbf{95.6} (0.2) &\textbf{94.6} (0.3) &\textbf{87.0} (0.2)  \\
 \cdashlinelr{2-6}
 &CapFilt (Oracle) &99.3 (0.0) &99.1 (0.0) &95.4 (0.4) \\
\midrule
\multirow{8}{*}{\texttt{mmimdb}} &LLaVA & 58.4 (0.2) &46.4 (0.2) &58.5 (0.1)  \\
 &Discrepancy & 57.8 (0.4) &46.1 (0.9) &57.4 (0.2)  \\
 &VDC & 80.5 (0.3) &67.1 (0.3) &69.3 (0.6)  \\
 &Datamap & 54.0 (0.3) &43.3 (0.4) &57.2 (0.1)  \\
 &deep k-NN & 61.2 (0.4) &47.2 (0.5) &58.3 (0.4)  \\
 &Confident & 52.8 (1.1) &41.4 (0.6) &51.8 (1.8)  \\
 &CLIP Sim. & 85.1 (0.3) &77.8 (0.7) &72.7 (0.6)  \\
 &\oursfixed (Ours) & 84.3 (0.3) &77.7 (0.8) &-  \\
 &\oursopt (Ours) & \textbf{86.0 }(0.1) &\textbf{79.4} (0.6) &\textbf{73.5} (0.3)  \\
 \cdashlinelr{2-6}
 &CapFilt &82.7 (0.7) &73.3 (1.2) &71.3 (0.3) \\
\midrule
\multirow{8}{*}{\texttt{mimiccxr}} &LLaVA & 53.9 (0.5) &42.7 (0.7) &57.0 (0.1)  \\
 &Datamap & 50.2 (1.2) &40.2 (1.0) &57.0 (0.1)  \\
 &Discrepancy & 60.0 (0.7) &50.2 (0.5) &57.2 (0.1)  \\
 &VDC & 50.8 (0.4) &40.3 (0.2) &57.0 (0.1)  \\
 &deep k-NN & 62.9 (0.4) &48.0 (0.3) &59.2 (0.1)  \\
 &Confident & 61.8 (0.3) &47.0 (0.2) &58.1 (0.6)  \\
 &CLIP Sim. & 64.1 (0.4) &51.7 (0.5) &59.2 (0.0)  \\
 &\oursfixed (Ours) & 66.3 (0.4) &54.6 (0.5) &-  \\
 &\oursopt (Ours) & \textbf{70.4 }(1.6) &\textbf{60.4} (1.6) &\textbf{61.1} (0.8)  \\
 \cdashlinelr{2-6}
 &CapFilt &49.7 (0.3) &40.0 (0.2) &57.0 (0.0) \\
\bottomrule
\end{tabular}
}
\label{tab:captioning_full}
\end{table}

\begin{table}[!htp]\centering
\caption{\texttt{flickr30k}: Label Error Detection}\label{tab:flickr_label_error}
\scriptsize
\begin{tabular}{lrrrrrrrr}
\toprule
\textbf{Dataset}& \textbf{Noise Type}& \textbf{Method}&\multicolumn{2}{c}{\textbf{AUROC}} &\multicolumn{2}{c}{\textbf{AUPRC}} &\multicolumn{2}{c}{\textbf{F1}} \\
\cmidrule{4-9}
& & &\textbf{mean} &\textbf{std} &\textbf{mean} &\textbf{std} &\textbf{mean} &\textbf{std} \\
\midrule
\multirow{20}{*}{\texttt{flickr30k}}
&\multirow{10}{*}{\textbf{noun}} &\textbf{LLaVA} &79.3 &0.8 &58.5 &0.2 &65.0 &1.1 \\
& &\textbf{Datamap} &52.7 &1.5 &37.9 &1.4 &50.4 &1.8 \\
& &\textbf{Discrepancy} &73.0 &0.6 &59.2 &1.8 &59.0 &0.3 \\
& &\textbf{VDC} &92.9 &1.1 &87.2 &0.3 &81.1 &1.6 \\
& &\textbf{Deep kNN} &71.1 &0.4 &52.0 &1.0 &59.2 &0.8 \\
& &\textbf{Confident} &63.1 &0.9 &42.1 &1.2 &54.0 &0.9 \\
& &\textbf{CLIP Sim.} &94.8 &0.5 &92.8 &0.5 &84.2 &0.9 \\
& &\textbf{\oursfixed} &93.6 &0.2 &92.0 &0.2 &83.4 &0.6 \\
& &\textbf{\oursopt} &94.5 &0.2 &92.8 &0.3 &83.6 &1.4 \\
\cdashlinelr{3-9}
& &\textbf{CapFilt} &98.6 &0.1 &98.1 &0.1 &93.1 &0.7 \\
\cmidrule{3-9}
&\multirow{10}{*}{\textbf{random}} &\textbf{LLaVA} &81.3 &1.0 &65.6 &1.4 &72.2 &1.1 \\
& &\textbf{Datamap} &68.2 &0.4 &61.1 &1.5 &58.9 &0.5 \\
& &\textbf{Discrepancy} &83.8 &1.2 &75.3 &2.3 &72.2 &1.4 \\
& &\textbf{VDC} &98.2 &0.2 &96.4 &0.3 &92.1 &0.6 \\
& &\textbf{Deep kNN} &81.1 &1.6 &65.3 &1.7 &72.7 &1.2 \\
& &\textbf{Confident} &71.2 &0.6 &55.3 &0.5 &67.1 &0.8 \\
& &\textbf{CLIP Sim.} &99.5 &0.1 &99.3 &0.1 &95.7 &0.5 \\
& &\textbf{\oursfixed} &99.4 &0.2 &99.3 &0.2 &96.3 &0.7 \\
& &\textbf{\oursopt} &99.5 &0.2 &99.4 &0.3 &96.3 &1.0 \\
\cdashlinelr{3-9}
& &\textbf{CapFilt} &99.9 &0.0 &99.8 &0.0 &97.9 &0.2 \\
\bottomrule
\end{tabular}
\end{table}

\begin{table}[!htp]\centering
\caption{\texttt{msccoco}: Label Error Detection}\label{tab:mscoco_label_error}
\scriptsize
\begin{tabular}{lrrrrrrrr}
\toprule
\textbf{Dataset}& \textbf{Noise Type}& \textbf{Method}&\multicolumn{2}{c}{\textbf{AUROC}} &\multicolumn{2}{c}{\textbf{AUPRC}} &\multicolumn{2}{c}{\textbf{F1}} \\\cmidrule{4-9}
& & &\textbf{mean} &\textbf{std} &\textbf{mean} &\textbf{std} &\textbf{mean} &\textbf{std} \\\midrule
\multirow{30}{*}{\texttt{mscoco}} &\multirow{10}{*}{\textbf{cat}} &\textbf{LLaVA} &80.3 &0.1 &63.4 &0.3 &74.9 &0.3 \\
&\textbf{} &\textbf{Datamap} &68.9 &0.8 &60.3 &0.0 &60.3 &1.2 \\
&\textbf{} &\textbf{Discrepancy} &72.7 &0.3 &67.2 &0.4 &62.5 &0.3 \\
&\textbf{} &\textbf{VDC} &94.1 &0.2 &91.8 &0.2 &86.3 &0.4 \\
&\textbf{} &\textbf{Deep kNN} &76.6 &0.4 &70.3 &0.6 &67.5 &0.8 \\
&\textbf{} &\textbf{Confident} &71.5 &0.5 &56.4 &0.5 &66.5 &0.5 \\
&\textbf{} &\textbf{CLIP Sim.} &93.8 &0.2 &93.0 &0.4 &84.5 &0.4 \\
&\textbf{} &\textbf{\oursfixed} &92.0 &0.1 &91.8 &0.3 &82.8 &0.4 \\
&\textbf{} &\textbf{\oursopt} &95.6 &0.2 &94.6 &0.3 &87.0 &0.2 \\
\cdashlinelr{3-9}
&\textbf{} &\textbf{CapFilt} &99.3 &0.0 &99.1 &0.0 &95.4 &0.4 \\

\cmidrule{3-9}
&\multirow{10}{*}{\textbf{noun}} &\textbf{LLaVA} &79.4 &0.2 &61.3 &0.3 &72.6 &0.2 \\
&\textbf{} &\textbf{Datamap} &62.1 &0.7 &50.0 &0.3 &56.2 &0.2 \\
&\textbf{} &\textbf{Discrepancy} &72.4 &0.6 &64.0 &0.4 &60.0 &0.9 \\
&\textbf{} &\textbf{VDC} &91.9 &0.5 &88.3 &0.7 &82.7 &0.5 \\
&\textbf{} &\textbf{Deep kNN} &75.7 &1.3 &66.8 &1.4 &65.8 &1.3 \\
&\textbf{} &\textbf{Confident} &69.8 &1.1 &52.8 &1.5 &63.3 &1.3 \\
&\textbf{} &\textbf{CLIP Sim.} &92.1 &0.2 &90.5 &0.2 &80.8 &0.6 \\
&\textbf{} &\textbf{\oursfixed} &90.4 &0.5 &89.5 &0.4 &80.2 &0.5 \\
&\textbf{} &\textbf{\oursopt} &92.9 &0.5 &91.5 &0.5 &82.3 &0.7 \\
\cdashlinelr{3-9}
&\textbf{} &\textbf{CapFilt} &98.7 &0.2 &98.4 &0.2 &93.6 &0.5 \\

\cmidrule{3-9}
&\multirow{10}{*}{\textbf{random}} &\textbf{LLaVA} &82.6 &0.3 &65.1 &0.6 &76.7 &0.2 \\
&\textbf{} &\textbf{Datamap} &78.8 &0.5 &71.5 &0.8 &67.0 &0.6 \\
&\textbf{} &\textbf{Discrepancy} &90.8 &0.4 &84.2 &0.6 &80.1 &0.9 \\
&\textbf{} &\textbf{VDC} &99.1 &0.2 &98.3 &0.2 &95.8 &0.3 \\
&\textbf{} &\textbf{Deep kNN} &94.5 &0.4 &88.4 &0.7 &87.9 &0.7 \\
&\textbf{} &\textbf{Confident} &85.0 &0.8 &71.4 &1.1 &81.8 &0.9 \\
&\textbf{} &\textbf{CLIP Sim.} &99.5 &0.1 &99.4 &0.1 &97.1 &0.1 \\
&\textbf{} &\textbf{\oursfixed} &99.5 &0.2 &99.4 &0.1 &97.3 &0.2 \\
&\textbf{} &\textbf{\oursopt} &99.6 &0.1 &99.5 &0.1 &97.5 &0.1 \\
\cdashlinelr{3-9}
&\textbf{} &\textbf{CapFilt} &99.9 &0.0 &99.9 &0.0 &98.9 &0.1 \\
\bottomrule
\end{tabular}
\end{table}

\begin{table}[!htp]\centering
\caption{\texttt{mmimdb}: Label Error Detection}\label{tab:mmimdb_label_error}
\scriptsize
\begin{tabular}{lrrrrrrrr}
\toprule
\textbf{Dataset}& \textbf{Noise Type}& \textbf{Method}&\multicolumn{2}{c}{\textbf{AUROC}} &\multicolumn{2}{c}{\textbf{AUPRC}} &\multicolumn{2}{c}{\textbf{F1}} \\
\cmidrule{4-9}
& & &\textbf{mean} &\textbf{std} &\textbf{mean} &\textbf{std} &\textbf{mean} &\textbf{std} \\
\midrule
\multirow{30}{*}{\texttt{mmimdb}} & \multirow{10}{*}{\textbf{cat}} & \textbf{LLaVA} & 58.4 & 0.2 & 46.4 & 0.2 & 58.5 & 0.1 \\
 &  & \textbf{Datamap}      & 54.0 & 0.3 & 43.3 & 0.4 & 57.2 & 0.1 \\
 &  & \textbf{Discrepancy}  & 57.8 & 0.4 & 46.1 & 0.9 & 57.4 & 0.2 \\
 &  & \textbf{VDC}          & 80.5 & 0.3 & 67.1 & 0.3 & 69.3 & 0.6 \\
 &  & \textbf{deep k-nn}    & 61.2 & 0.4 & 47.2 & 0.5 & 58.3 & 0.4 \\
 &  & \textbf{Confident}    & 52.8 & 1.1 & 41.4 & 0.6 & 51.8 & 1.8 \\
 &  & \textbf{CLIP Sim.}    & 85.1 & 0.3 & 77.8 & 0.7 & 72.7 & 0.6 \\
 &  & \textbf{\oursfixed}   & 84.3 & 0.3 & 77.7 & 0.8 & 69.9 & 0.8 \\
 &  & \textbf{\oursopt}     & 86.0 & 0.1 & 79.4 & 0.6 & 73.5 & 0.3 \\
 \cdashlinelr{3-9}
 &  & \textbf{CapFilt}      & 82.7 & 0.7 & 73.3 & 1.2 & 71.3 & 0.3 \\
\cmidrule{2-9}
 & \multirow{10}{*}{\textbf{noun}} & \textbf{LLaVA} & 59.1 & 0.3 & 44.2 & 0.6 & 55.2 & 0.2 \\
 &  & \textbf{Datamap}      & 47.5 & 0.5 & 35.0 & 0.4 & 54.0 & 0.3 \\
 &  & \textbf{Discrepancy}  & 58.8 & 1.0 & 43.3 & 1.3 & 54.6 & 0.7 \\
 &  & \textbf{VDC}          & 79.0 & 0.2 & 62.3 & 0.3 & 65.9 & 0.4 \\
 &  & \textbf{deep k-nn}    & 61.4 & 0.1 & 44.2 & 0.3 & 55.8 & 0.3 \\
 &  & \textbf{Confident}    & 53.7 & 1.3 & 38.8 & 0.8 & 48.6 & 1.9 \\
 &  & \textbf{CLIP Sim.}    & 82.8 & 0.4 & 72.8 & 0.5 & 68.4 & 1.1 \\
 &  & \textbf{\oursfixed}   & 82.1 & 0.4 & 72.7 & 0.6 & 65.2 & 1.1 \\
 &  & \textbf{\oursopt}     & 84.2 & 0.5 & 75.9 & 0.5 & 69.5 & 0.4 \\
 \cdashlinelr{3-9}
 &  & \textbf{CapFilt}      & 79.9 & 0.1 & 66.2 & 0.4 & 67.1 & 0.3 \\
\cmidrule{2-9}
 & \multirow{10}{*}{\textbf{random}} & \textbf{LLaVA} & 58.5 & 0.8 & 46.7 & 0.5 & 58.5 & 0.1 \\
 &  & \textbf{Datamap}      & 54.3 & 0.9 & 43.6 & 1.1 & 57.2 & 0.1 \\
 &  & \textbf{Discrepancy}  & 60.0 & 0.7 & 47.4 & 0.8 & 58.0 & 0.5 \\
 &  & \textbf{VDC}          & 82.7 & 0.2 & 69.6 & 0.4 & 72.1 & 0.4 \\
 &  & \textbf{deep k-nn}    & 64.0 & 0.2 & 49.0 & 0.1 & 60.4 & 0.1 \\
 &  & \textbf{Confident}    & 52.0 & 1.0 & 41.0 & 0.5 & 52.1 & 3.1 \\
 &  & \textbf{CLIP Sim.}    & 88.1 & 0.1 & 82.0 & 0.2 & 75.7 & 0.4 \\
 &  & \textbf{\oursfixed}   & 87.6 & 0.1 & 81.9 & 0.3 & 73.8 & 0.2 \\
 &  & \textbf{\oursopt}     & 89.3 & 0.9 & 84.2 & 1.3 & 77.0 & 1.2 \\
 \cdashlinelr{3-9}
 &  & \textbf{CapFilt}      & 84.9 & 0.4 & 76.4 & 0.7 & 73.1 & 0.3 \\
\bottomrule
\end{tabular}
\end{table}

\begin{table}[!htp]\centering
\caption{\texttt{mimiccxr}: Label Error Detection}\label{tab:mimiccxr_label_error}
\scriptsize
\begin{tabular}{lrrrrrrrr}
\toprule
\textbf{Dataset} & \textbf{Noise Type} & \textbf{Method} & \multicolumn{2}{c}{\textbf{AUROC}} & \multicolumn{2}{c}{\textbf{AUPRC}} & \multicolumn{2}{c}{\textbf{F1}} \\
\cmidrule{4-9}
& & & \textbf{mean} & \textbf{std} & \textbf{mean} & \textbf{std} & \textbf{mean} & \textbf{std} \\
\midrule
\multirow{20}{*}{\texttt{mimiccxr}} & \multirow{10}{*}{\textbf{cat}} & \textbf{LLaVA}        & 53.9 & 0.5 & 42.7 & 0.7 & 57.0 & 0.1 \\
& & \textbf{Datamap}       & 50.2 & 1.2 & 40.2 & 1.0 & 57.0 & 0.1 \\
& & \textbf{Discrepancy}   & 60.0 & 0.7 & 50.2 & 0.5 & 57.2 & 0.1 \\
& & \textbf{VDC}           & 50.8 & 0.4 & 40.3 & 0.2 & 57.0 & 0.1 \\
& & \textbf{Deep k‑NN}     & 62.9 & 0.4 & 48.0 & 0.3 & 59.2 & 0.1 \\
& & \textbf{Confident}     & 61.8 & 0.3 & 47.0 & 0.2 & 58.1 & 0.6 \\
& & \textbf{CLIP Sim.}     & 64.1 & 0.4 & 51.7 & 0.5 & 59.2 & 0.0 \\
& & \textbf{\oursfixed}    & 66.3 & 0.4 & 54.6 & 0.5 & 55.5 & 0.3 \\
& & \textbf{\oursopt}      & 70.4 & 1.6 & 60.4 & 1.6 & 61.1 & 0.8 \\
 \cdashlinelr{3-9}
& & \textbf{CapFilt}       & 49.7 & 0.3 & 40.0 & 0.2 & 57.0 & 0.0 \\
\cmidrule{3-9}
& \multirow{10}{*}{\textbf{random}} & \textbf{LLaVA}        & 50.8 & 0.4 & 40.6 & 0.2 & 57.1 & 0.0 \\
& & \textbf{Datamap}       & 51.3 & 0.4 & 40.9 & 0.6 & 57.1 & 0.0 \\
& & \textbf{Discrepancy}   & 62.5 & 0.5 & 52.2 & 1.0 & 57.2 & 0.5 \\
& & \textbf{VDC}           & 52.4 & 0.9 & 41.6 & 0.7 & 57.2 & 0.1 \\
& & \textbf{Deep k‑NN}     & 66.6 & 0.6 & 53.8 & 1.1 & 59.4 & 0.2 \\
& & \textbf{Confident}     & 65.0 & 1.0 & 49.5 & 0.7 & 61.6 & 1.1 \\
& & \textbf{CLIP Sim.}     & 66.8 & 0.8 & 54.4 & 0.9 & 60.1 & 0.4 \\
& & \textbf{\oursfixed}    & 69.5 & 0.7 & 57.8 & 1.0 & 57.7 & 1.2 \\
& & \textbf{\oursopt}      & 73.7 & 1.7 & 64.1 & 2.2 & 63.5 & 0.8 \\
 \cdashlinelr{3-9}
& & \textbf{CapFilt}       & 50.6 & 0.4 & 40.1 & 0.5 & 57.1 & 0.0 \\
\bottomrule
\end{tabular}
\end{table}

\subsection{Varying Noise Level}
\label{sec:vary_noise_app}

\begin{figure*}[t]
  \centering
  \begin{subfigure}{0.24\textwidth}
  \includegraphics[width=1.0\linewidth]{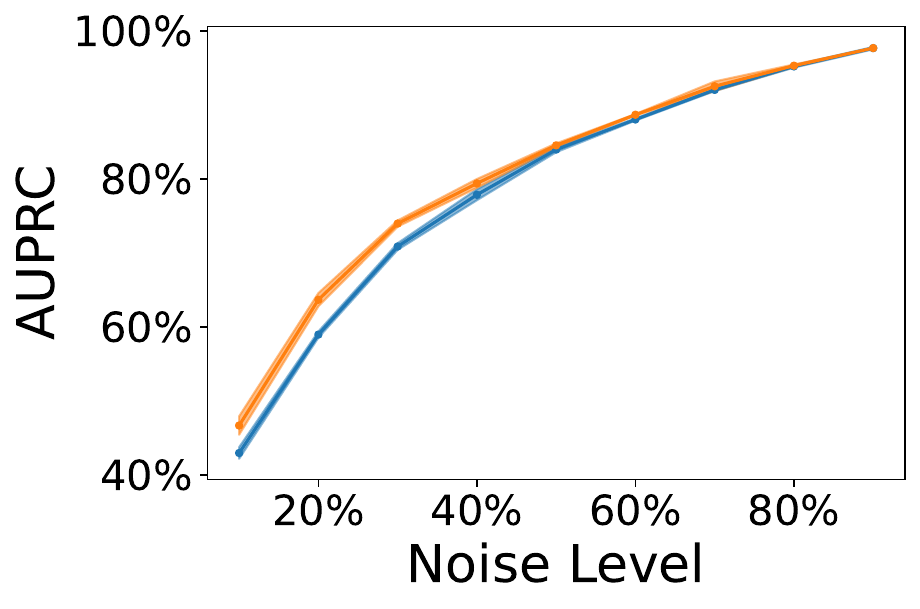}
    \caption{AUPRC on \texttt{mmimdb}} 
  \end{subfigure}%
     \hfill
   \begin{subfigure}{0.24\textwidth}
  \includegraphics[width=1.0\linewidth]{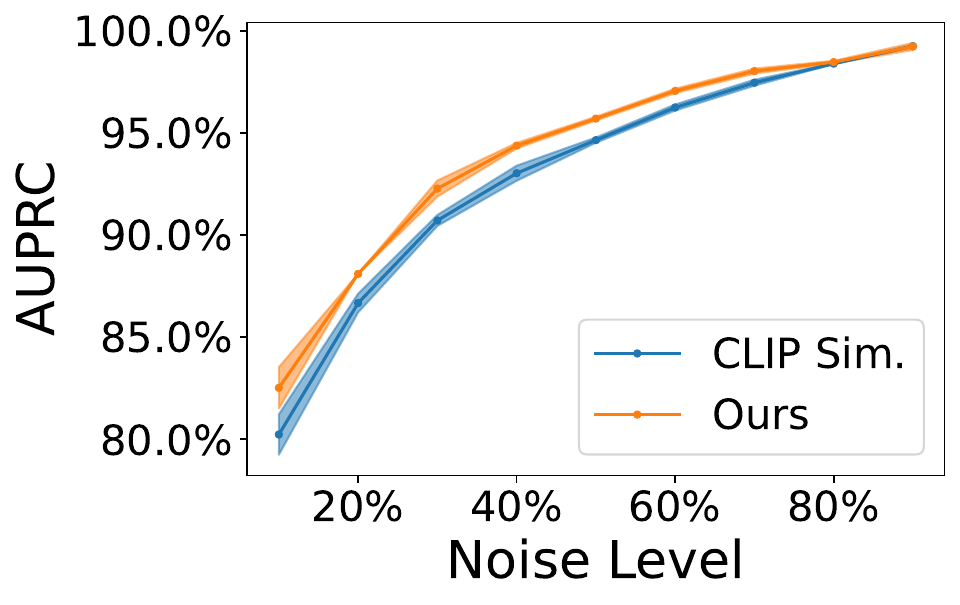}
    \caption{AUPRC on \texttt{mscoco}} 
  \end{subfigure}%
  \hfill
   \begin{subfigure}{0.24\textwidth}
  \includegraphics[width=1.0\linewidth]{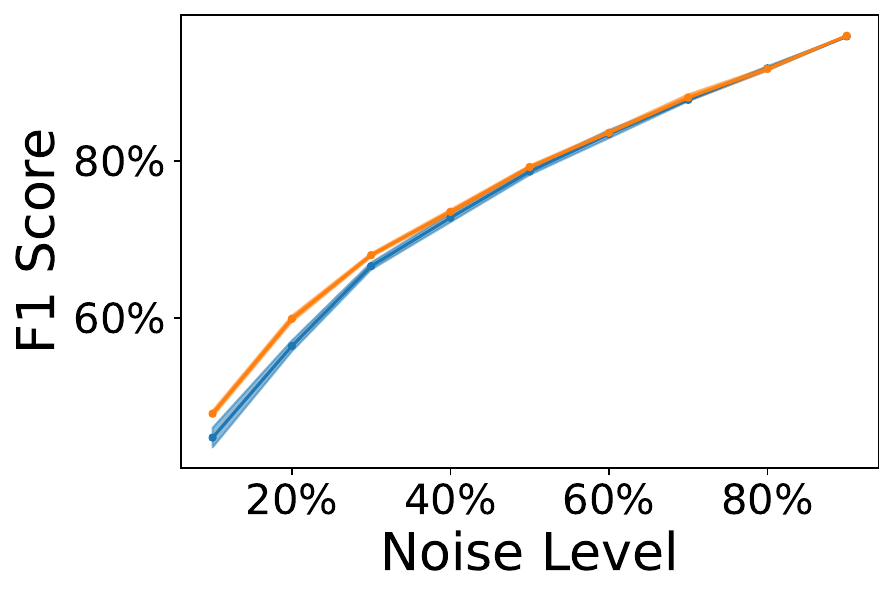}
    \caption{F1 on \texttt{mmimdb}} 
  \end{subfigure}%
   \hfill
   \begin{subfigure}{0.24\textwidth}
  \includegraphics[width=1.0\linewidth]{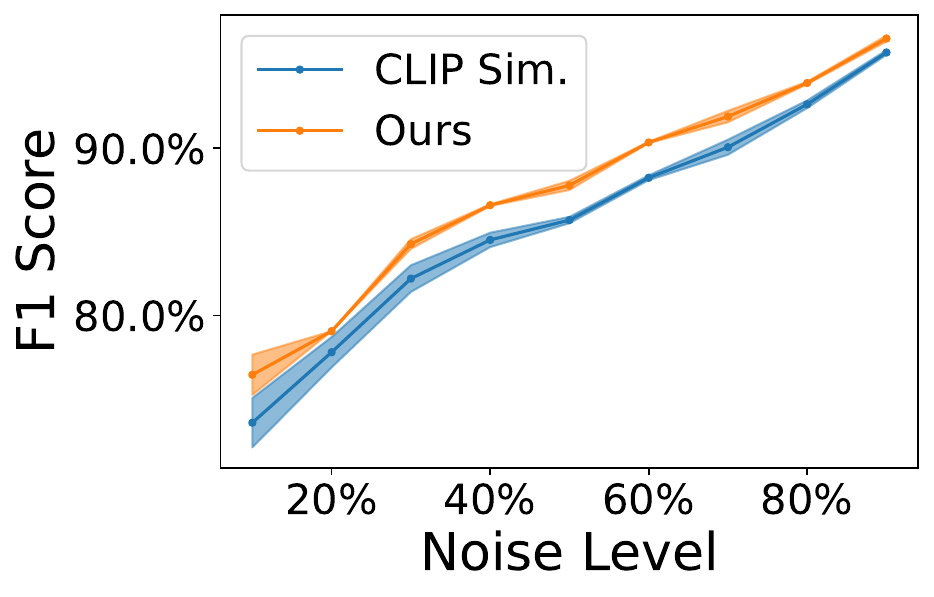}
    \caption{F1 on \texttt{mscoco}}
  \end{subfigure}%
  \\
   \begin{subfigure}{0.24\textwidth}
  \includegraphics[width=1.0\linewidth]{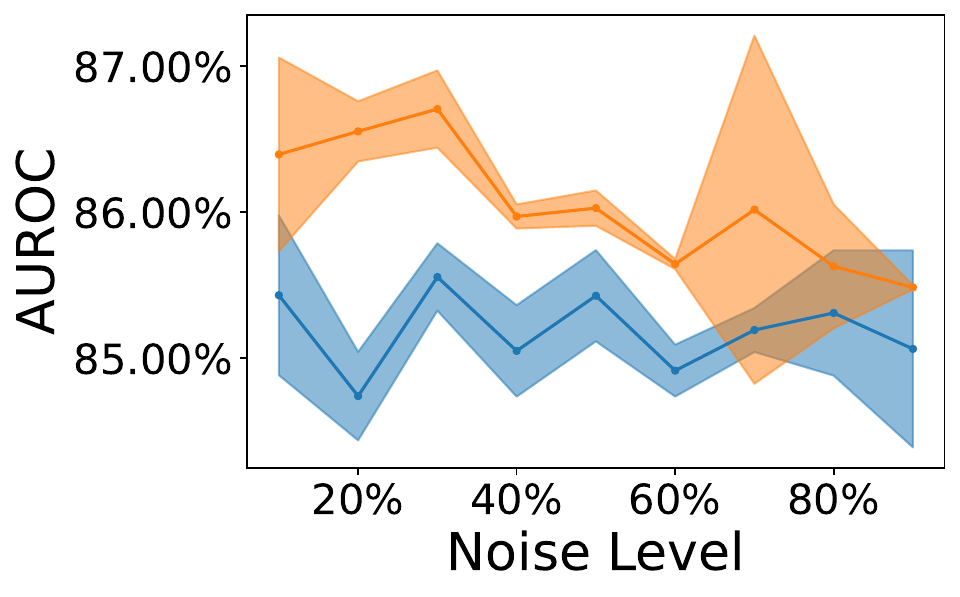}
    \caption{AUROC on \texttt{mmimdb}} 
  \end{subfigure}%
   \begin{subfigure}{0.24\textwidth}
  \includegraphics[width=1.0\linewidth]{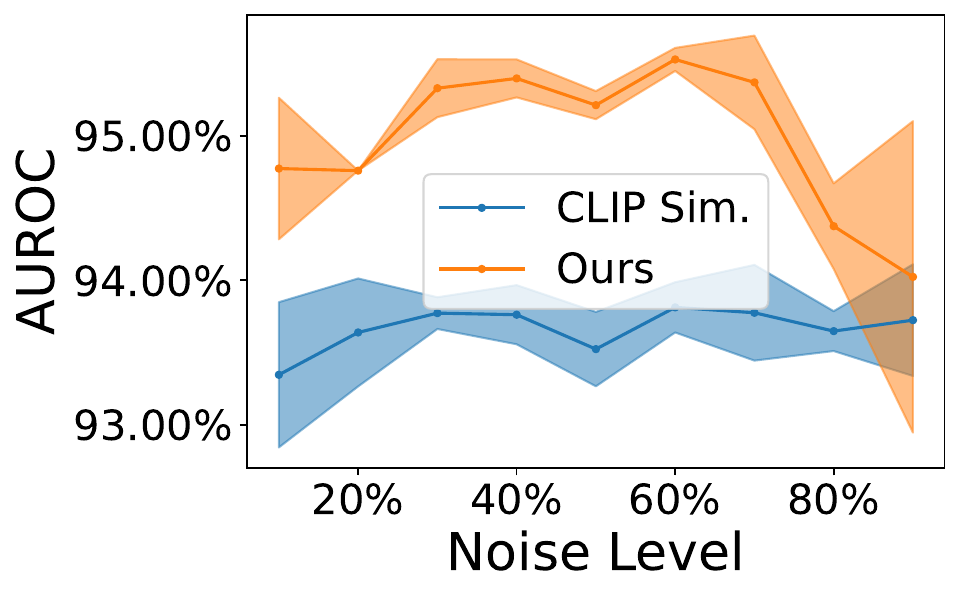}
    \caption{AUROC on \texttt{mscoco}} 
  \end{subfigure}%
  \caption{Test-set performance of \oursopt compared to the CLIP Similarity for varying levels of the synthetic noise. }
  \label{fig:vary_noise}
\end{figure*}

We show the AUROC for varying noise levels in Figure \ref{fig:vary_noise}.

\subsection{Robustness to Hyperparameters}
\label{sec:robustness_to_hparams}

\begin{figure}[htbp]
\begin{subfigure}{0.4\linewidth}
        \centering  
        \includegraphics[width=1.0\linewidth]{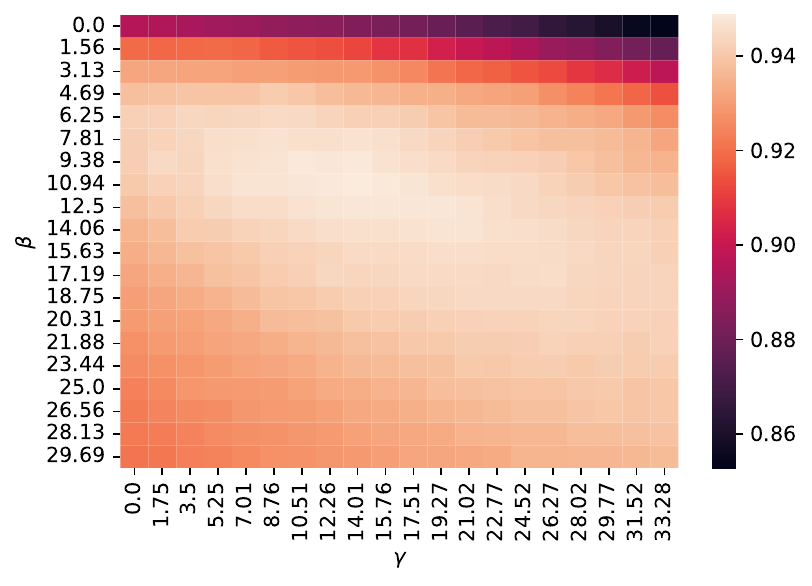}
\caption{\texttt{cifar10}, asymmetric noise}
\end{subfigure}%
\hfill
\begin{subfigure}{0.4\linewidth}
        \centering  
        \includegraphics[width=1.0\linewidth]{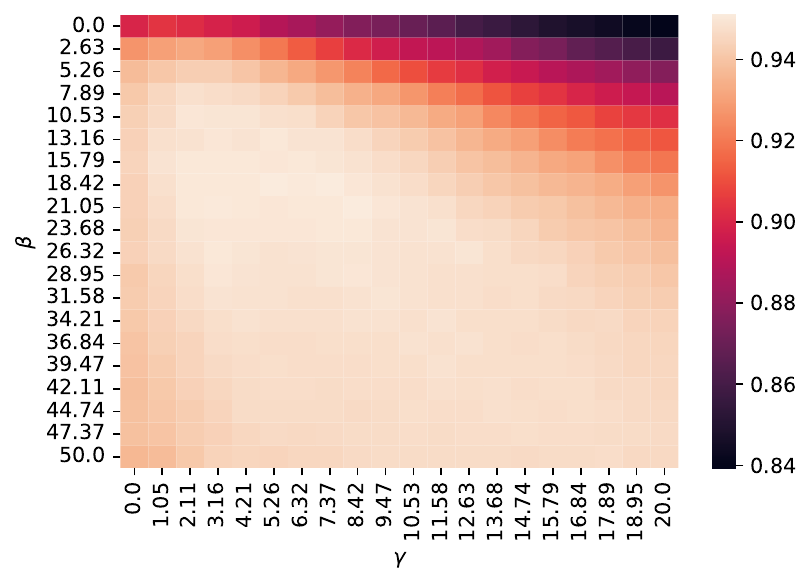}
\caption{\texttt{cifar10}, symmetric noise}
\end{subfigure} \\
\begin{subfigure}{0.4\linewidth}
        \centering  
        \includegraphics[width=1.0\linewidth]{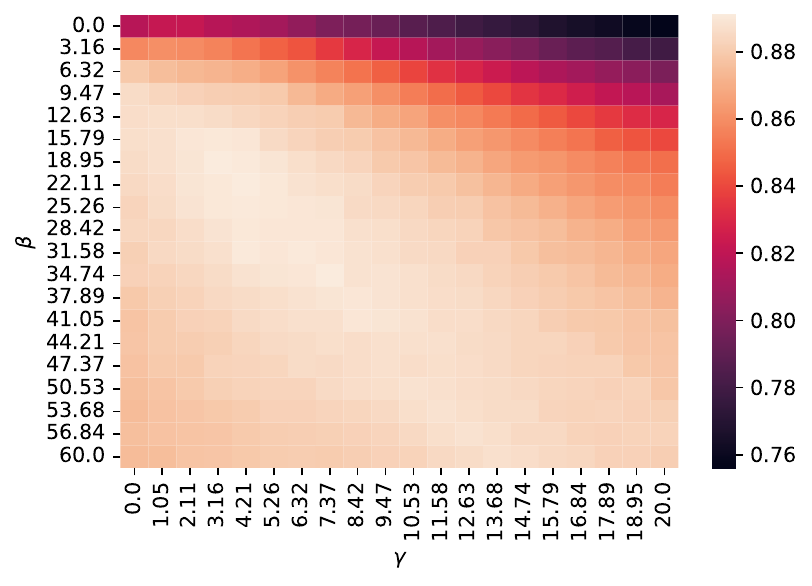}
\caption{\texttt{cifar10}, real noise}
\end{subfigure}%
\hfill
\begin{subfigure}{0.4\linewidth}
        \centering  
        \includegraphics[width=1.0\linewidth]{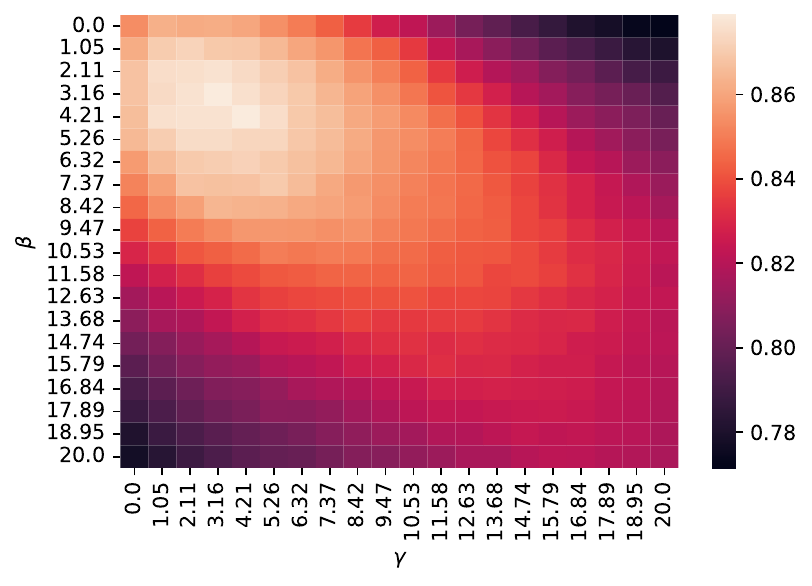}
\caption{\texttt{cifar100}, asymmetric noise}
\end{subfigure} \\
\begin{subfigure}{0.4\linewidth}
        \centering  
        \includegraphics[width=1.0\linewidth]{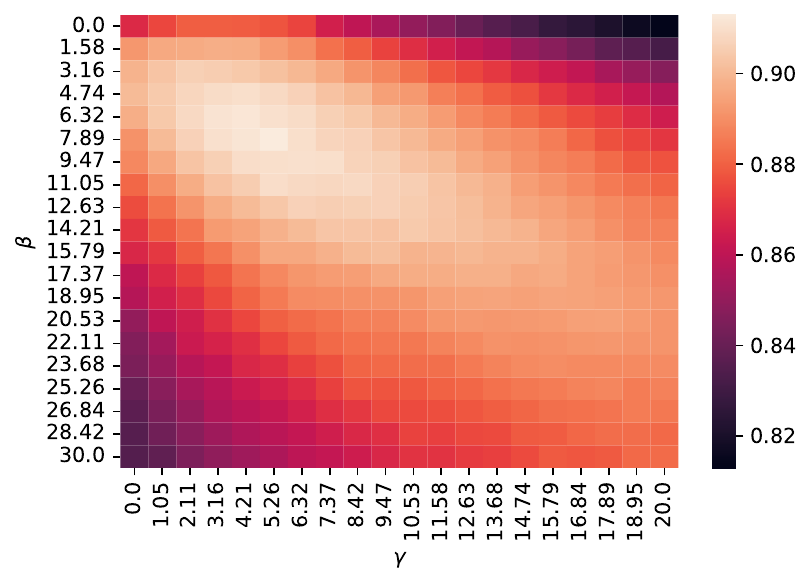}
\caption{\texttt{cifar100}, symmetric noise}
\end{subfigure}%
\hfill
\begin{subfigure}{0.4\linewidth}
        \centering  
        \includegraphics[width=1.0\linewidth]{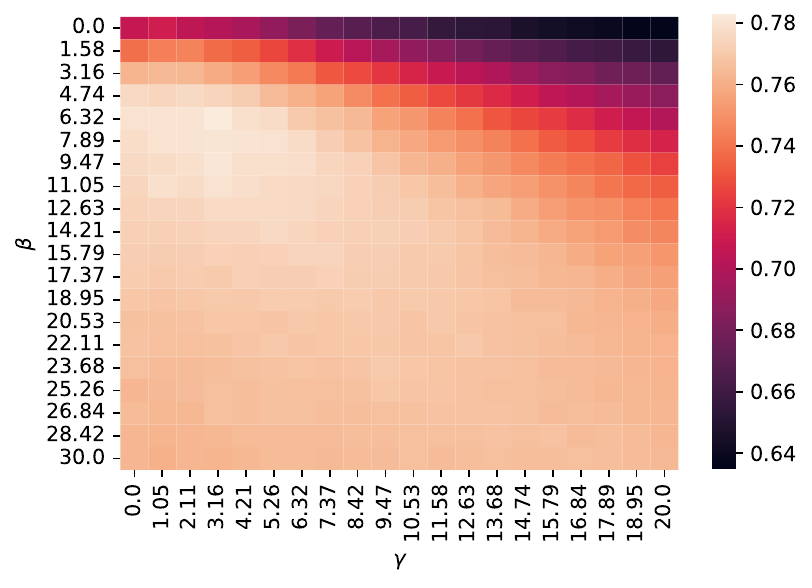}
\caption{\texttt{cifar100}, real noise}
\end{subfigure} \\
\begin{subfigure}{0.4\linewidth}
        \centering  
        \includegraphics[width=1.0\linewidth]{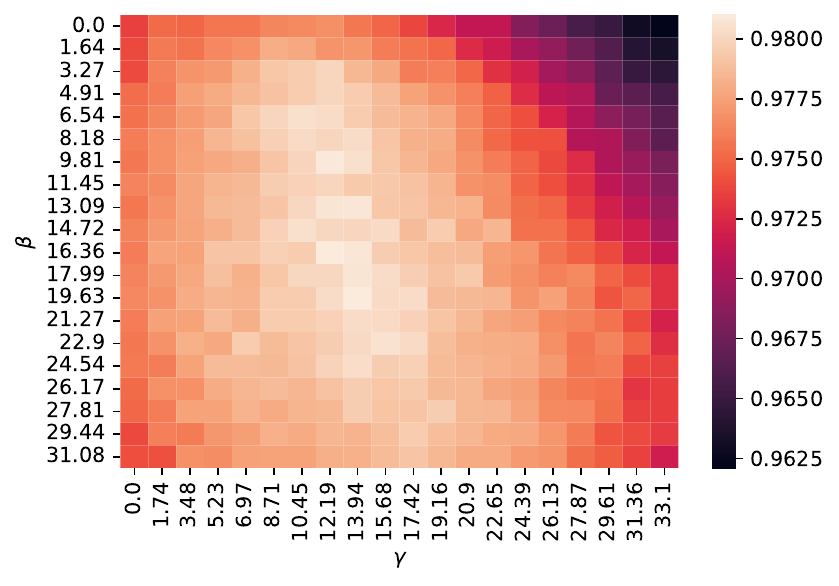}
\caption{\texttt{mscoco}, random noise}
\end{subfigure}
\hfill
\begin{subfigure}{0.4\linewidth}
        \centering  
        \includegraphics[width=1.0\linewidth]{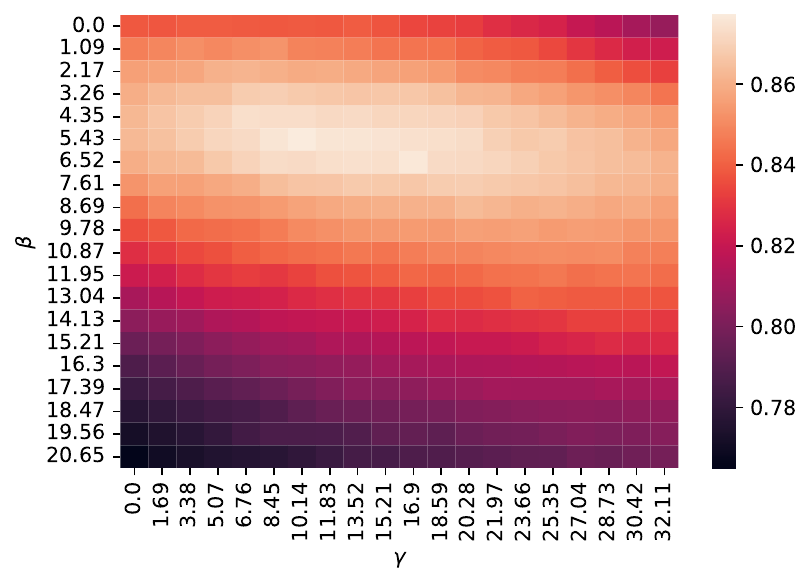}
\caption{\texttt{mscoco}, cat noise}
\end{subfigure}
\end{figure}

\begin{figure} \ContinuedFloat %
\begin{subfigure}{0.4\linewidth}
        \centering  
        \includegraphics[width=1.0\linewidth]{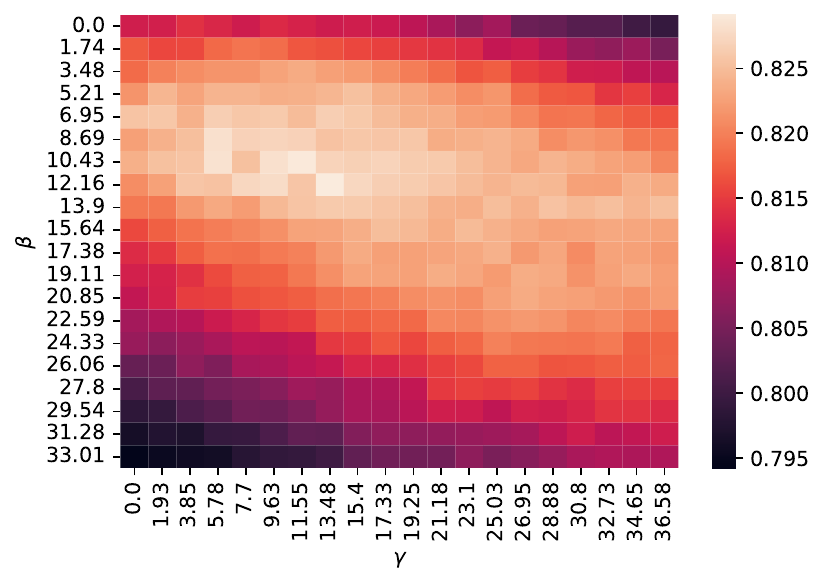}
\caption{\texttt{mscoco}, noun noise}
\end{subfigure}
\hfill
\begin{subfigure}{0.4\linewidth}
        \centering  
        \includegraphics[width=1.0\linewidth]{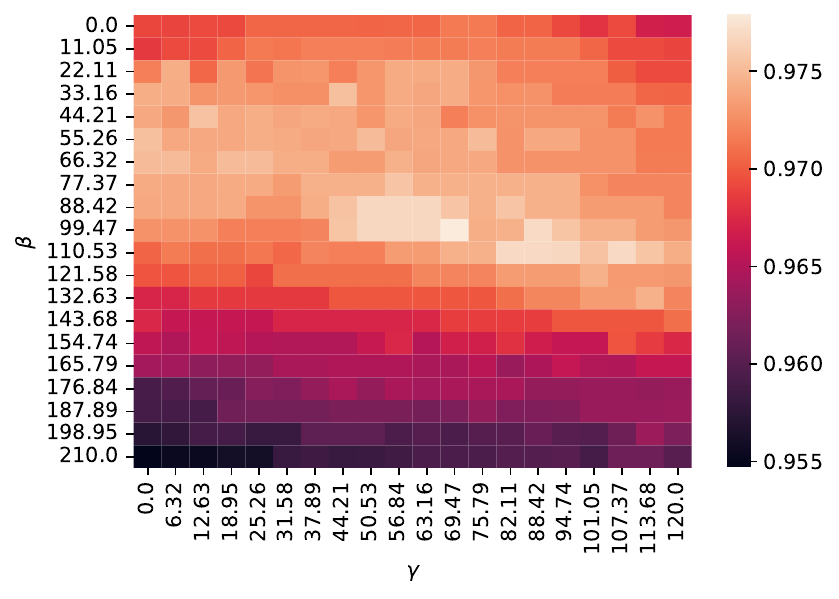}
\caption{\texttt{flickr30k}, random noise}
\end{subfigure} \\
\begin{subfigure}{0.4\linewidth}
        \centering  
        \includegraphics[width=1.0\linewidth]{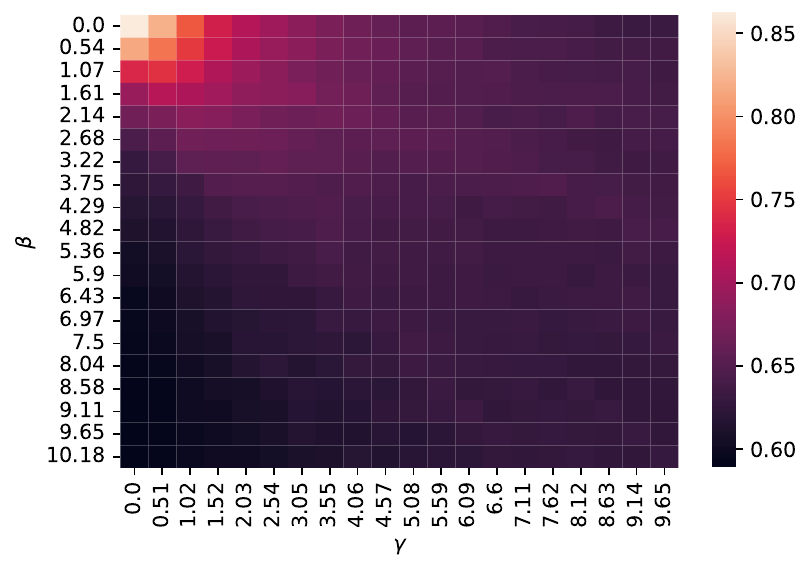}
\caption{\texttt{flickr30k}, noun noise}
\end{subfigure}
\hfill
\begin{subfigure}{0.4\linewidth}
        \centering  
        \includegraphics[width=1.0\linewidth]{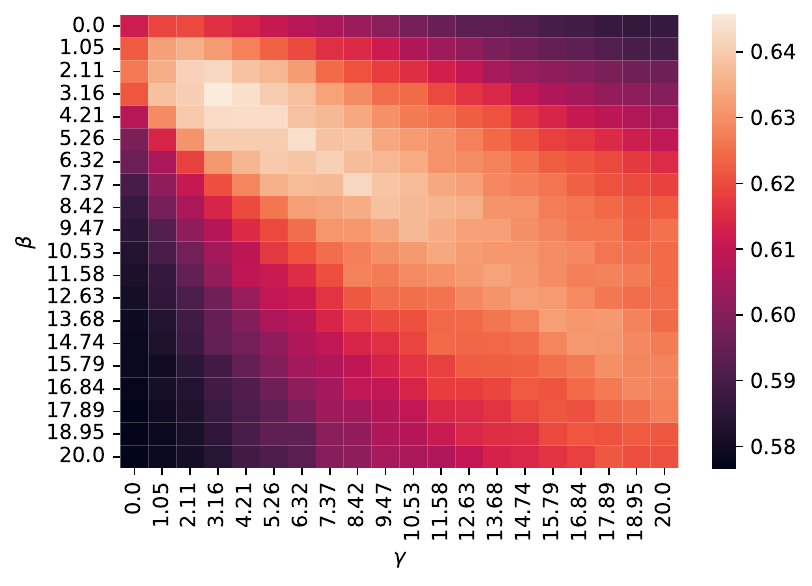}
\caption{\texttt{mimiccxr}, random noise} 
\end{subfigure} \\
\begin{subfigure}{0.4\linewidth}
        \centering  
        \includegraphics[width=1.0\linewidth]{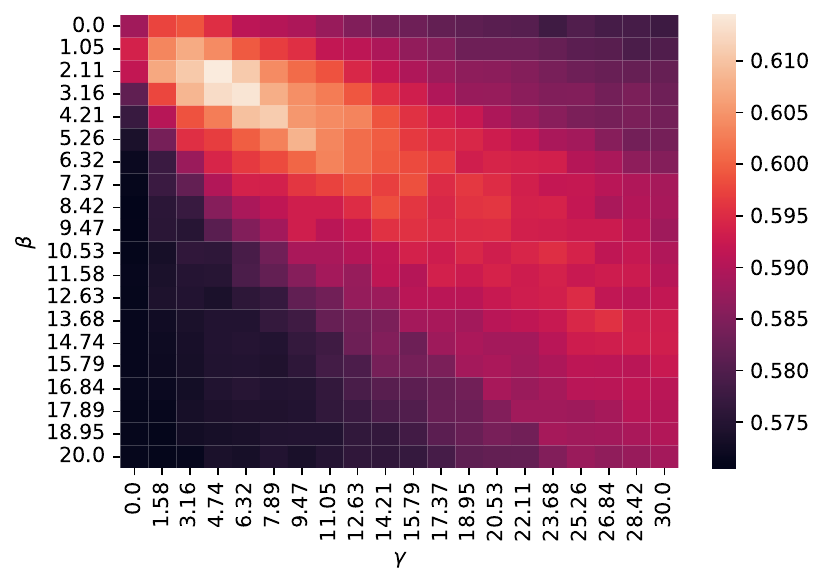}
\caption{\texttt{mimiccxr}, cat noise}
\end{subfigure}
\hfill
\begin{subfigure}{0.4\linewidth}
        \centering  
        \includegraphics[width=1.0\linewidth]{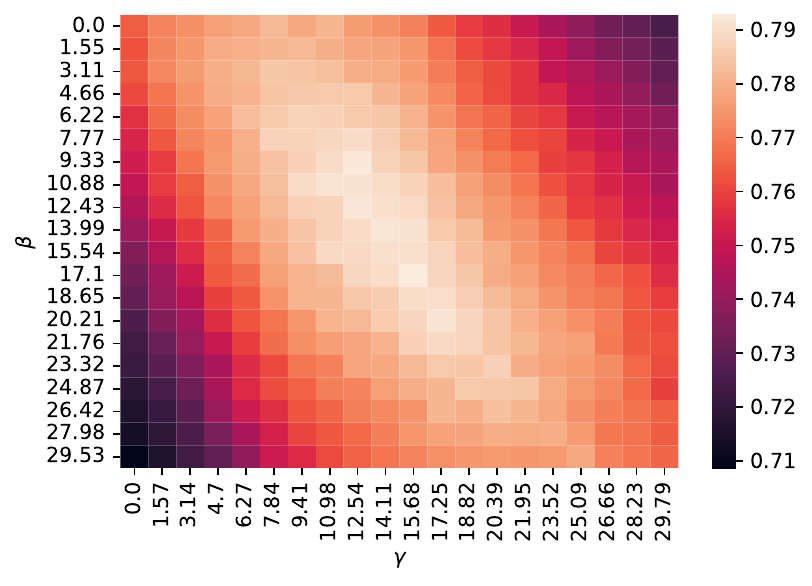}
\caption{\texttt{mmimdb}, random noise} 
\end{subfigure} \\
\begin{subfigure}{0.4\linewidth}
        \centering  
        \includegraphics[width=1.0\linewidth]{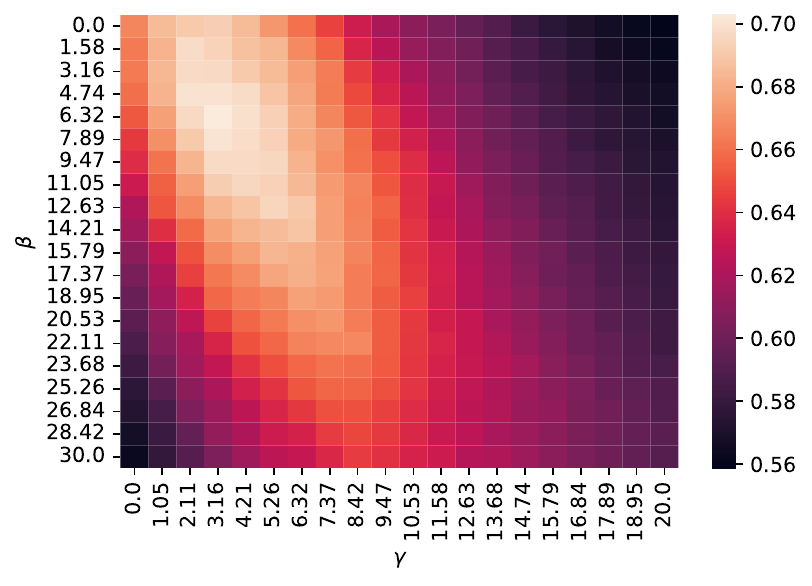}
\caption{\texttt{mmimdb}, noun noise}
\end{subfigure}
\hfill
\begin{subfigure}{0.4\linewidth}
        \centering  
        \includegraphics[width=1.0\linewidth]{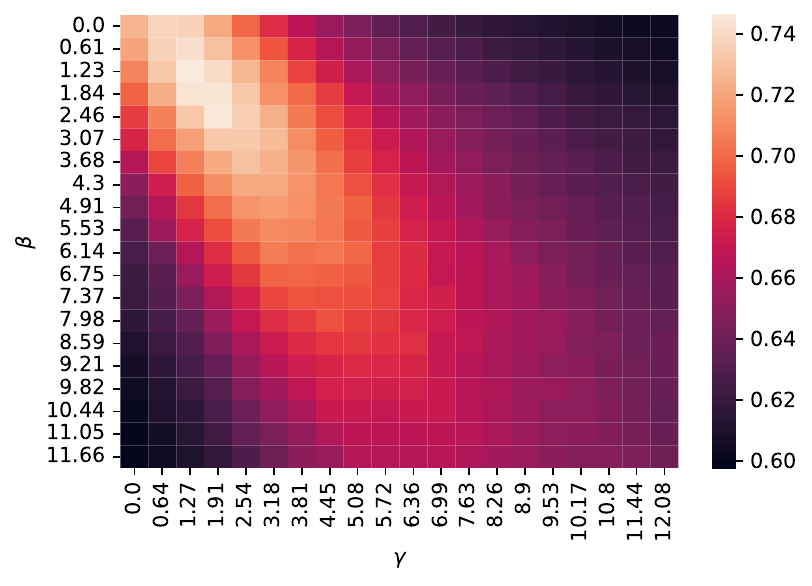}
\caption{\texttt{mmimdb}, cat noise} 
\end{subfigure}
\caption{F1 of our method for varying $\beta$ and $\gamma$, keeping all other hyperparameters their fixed optimal values.}
\label{fig:huge_heatmaps}
\end{figure}

We show the test-set F1 of \ours for varying $\beta$ and $\gamma$, keeping all other hyperparameters at their fixed optimal values, in Figure \ref{fig:huge_heatmaps}. In Table \ref{tab:no_hparams_still_good}, we show the performance of \ours when hyperparameters are fixed (at $k = 30$, cosine distance, $\beta = \gamma = 5$, $\tau_{1, n} = \tau_{1, m} = 0.1$, and $\tau_{2, n} = \tau_{2, m} = 5$) versus when they are optimized using a labeled validation set. Note that F1 is not computed as it requires external information to select a threshold.

\begin{table}[]
\caption{We show the AUROC and AUPRC of \ours when we search for the optimal hyperparameters using a labeled validation set (\oursopt) and when we use fixed hyperparameters (\oursfixed:  $k = 30$, cosine distance, $\beta = \gamma = 5$, $\tau_{1, n} = \tau_{1, m} = 0.1$, and $\tau_{2, n} = \tau_{2, m} = 5$). The mean gap in AUROC is -1.6 (1.3), and the mean gap in AUPRC is -1.6 (2.4). Note that F1 is not computed as it requires external information to select a threshold. }
\centering
\resizebox{0.8\textwidth}{!}{ 
\begin{tabular}{@{}rrrrrrrr@{}}
\toprule
\multicolumn{1}{l}{}                & \multicolumn{1}{l}{} & \multicolumn{3}{c}{\textbf{AUROC}}               & \multicolumn{3}{c}{\textbf{AUPRC}}                 \\ \cmidrule(l){3-5} \cmidrule(l){6-8}
\textbf{Dataset}                    & \textbf{Noise Type}  & \textbf{\oursopt} & \textbf{\oursfixed} & \textbf{Gap} & \textbf{\oursopt} & \textbf{\oursfixed}  & \textbf{Gap} \\ \midrule
\multirow{3}{*}{\texttt{cifar10}}   & \textbf{asymmetric}  & 98.8 (0.2)       & 97.5 (0.2)     & -1.4 (0.1)   & 97.8 (0.5)       & 94.8 (0.6)     & -3.0 (0.1)   \\
                                    & \textbf{real}        & 98.1 (0.0)       & 97.7 (0.2)     & -0.5 (0.2)   & 97.4 (0.1)       & 96.8 (0.3)     & -0.5 (0.2)   \\
                                    & \textbf{symmetric}   & 99.6 (0.1)       & 99.5 (0.1)     & -0.2 (0.1)   & 99.4 (0.1)       & 99.2 (0.1)     & -0.2 (0.1)   \\ \midrule
\multirow{3}{*}{\texttt{cifar100}}  & \textbf{asymmetric}  & 96.7 (0.3)       & 94.9 (0.3)     & -1.9 (0.1)   & 95.3 (0.4)       & 92.1 (0.4)     & -3.2 (0.1)   \\
                                    & \textbf{real}        & 90.8 (0.0)       & 88.9 (0.7)     & -1.8 (0.7)   & 87.4 (0.3)       & 84.6 (1.1)     & -2.8 (0.9)   \\
                                    & \textbf{symmetric}   & 99.0 (0.0)       & 98.4 (0.1)     & -0.7 (0.1)   & 98.7 (0.1)       & 97.7 (0.2)     & -1.0 (0.1)   \\ \midrule
\multirow{1}{*}{\texttt{miniImageNet}}  & \textbf{human}      & 90.0 (0.4)               & 89.5 (0.2)            & -0.5 (0.2)           & 79.7 (3.0)               & 81.5 (0.3)             &    +1.8 (2.7)          \\ \midrule
\multirow{1}{*}{\texttt{StanfordCars}}  & \textbf{human}      & 73.1 (0.5)               &     72.6 (0.7)        & -0.5 (0.5)         & 40.5 (0.4)              &     44.9 (1.4)         & +4.4 (1.0)           \\ \midrule
\multirow{2}{*}{\texttt{flickr30k}} & \textbf{noun}        & 94.5 (0.2)       & 93.6 (0.2)     & -0.9 (0.3)   & 92.8 (0.3)       & 92.0 (0.2)     & -0.8 (0.1)   \\
                                    & \textbf{random}      & 99.5 (0.2)       & 99.4 (0.2)     & -0.0 (0.1)   & 99.4 (0.3)       & 99.3 (0.2)     & -0.1 (0.2)   \\ \midrule
\multirow{2}{*}{\texttt{mimiccxr}}  & \textbf{cat}         & 70.4 (1.6)       & 66.3 (0.4)     & -4.1 (1.5)   & 60.4 (1.6)       & 54.6 (0.5)     & -5.8 (1.5)   \\
                                    & \textbf{random}      & 73.7 (1.7)       & 69.5 (0.7)     & -4.1 (1.5)   & 64.1 (2.2)       & 57.8 (1.0)     & -6.3 (1.7)   \\ \midrule
\multirow{3}{*}{\texttt{mmimdb}}    & \textbf{cat}         & 86.0 (0.1)       & 84.3 (0.3)     & -1.6 (0.3)   & 79.4 (0.6)       & 77.7 (0.8)     & -1.7 (0.2)   \\
                                    & \textbf{noun}        & 84.2 (0.5)       & 82.1 (0.4)     & -2.1 (0.6)   & 75.9 (0.5)       & 72.7 (0.6)     & -3.2 (0.5)   \\
                                    & \textbf{random}      & 89.3 (0.9)       & 87.6 (0.1)     & -1.6 (0.8)   & 84.2 (1.3)       & 81.9 (0.3)     & -2.3 (1.2)   \\ \midrule
\multirow{3}{*}{\texttt{mscoco}}    & \textbf{cat}         & 95.6 (0.2)       & 92.0 (0.1)     & -3.6 (0.1)   & 94.6 (0.3)       & 91.8 (0.3)     & -2.8 (0.1)   \\
                                    & \textbf{noun}        & 92.9 (0.5)       & 90.4 (0.5)     & -2.5 (0.2)   & 91.5 (0.5)       & 89.5 (0.4)     & -2.0 (0.3)   \\
                                    & \textbf{random}      & 99.6 (0.1)       & 99.5 (0.2)     & -0.1 (0.0)   & 99.5 (0.1)       & 99.4 (0.1)     & -0.1 (0.0)   \\ \bottomrule
\end{tabular}

}
\label{tab:no_hparams_still_good}
\end{table}

\subsection{Ablations of our Method}
Ablations of our method can be found in Table \ref{tab:ablation} and Table \ref{tab:ablation_rebuttal}.

\subsection{Runtime Comparison}
We compare the runtime of \ours with baselines in Table \ref{tab:runtime}.

\begin{table}[t!]\centering
\caption{Downstream captioning performance when removing 40\% samples with highest mislabel scores, and models are trained without early stopping for 10 epochs. We find that filtering noisy data with \oursopt improves captioning.}
\begin{adjustbox}{width = 0.4\linewidth}
\begin{tabular}{lcrrr}\toprule
\textbf{Dataset}  & \textbf{Method} & \textbf{B@4} &\textbf{CIDER} &\textbf{ROUGE}\\\midrule

\multirow{4}{*}{\texttt{flickr30k}} & No Filtering &28.0 &	49.5	& 65.1\\
 & CLIP Sim.& 29.1 & 50.5 &	71.4\\
&\oursopt & 29.5 & 50.9	& 72.1
\\ \cdashlinelr{2-5}
  &Clean& 30.8 &	51.9 &	74.6\\

\midrule
\multirow{4}{*}{\texttt{mscoco}} & No Filtering &35.0 & 56.3 &	116.7\\
 & CLIP Sim.&38.1 &	58.5 &	126.9	\\

&\oursopt&37.9 &	58.4 &	126.5	\\

\cdashlinelr{2-5}
  &Clean&38.2	 & 58.5	 & 127.7	\\
\bottomrule
\end{tabular}
\end{adjustbox}
\label{tab:captioning_downstream_results_extra}
\end{table}

\begin{table}[htbp!]
\caption{Performance of our method after ablating various components. We find that mislabel detection performance almost decreases monotonically as we remove additional components, with the exception of two metrics on \texttt{mmimdb} where one ablation is statistically comparable to the original method. }
\centering
\resizebox{0.9\textwidth}{!}{  
\begin{tabular}{@{}lrrrrrr@{}}
\toprule
                   & \multicolumn{3}{c}{\texttt{mmimdb}}                                  & \multicolumn{3}{c}{\texttt{mscoco}}                                  \\ 
\cmidrule(l){2-4} \cmidrule(l){5-7} 
                   & \textbf{AUROC} & \textbf{AUPRC} & \textbf{F1}  & \textbf{AUROC} & \textbf{AUPRC} & \textbf{F1} \\ 
\cmidrule(l){2-4} \cmidrule(l){5-7} 
\oursopt (Ours)     & 86.0 (0.1)     & 79.4 (0.6)     & 73.5 (0.3)     & \textbf{95.6} (0.2) & \textbf{94.6} (0.3) & \textbf{87.0} (0.2) \\
$- \tau_1$          & 85.3 (0.3)     & 78.2 (1.1)     & 72.9 (0.5)     & 94.6 (0.3)          & 93.8 (0.4)          & 85.2 (0.5)          \\
$- \tau_2$          & 85.6 (0.1)     & 78.6 (0.5)     & 73.3 (0.3)     & 94.7 (0.2)          & 93.8 (0.5)          & 85.4 (0.8)          \\
$- \tau_1, \tau_2$  & 85.4 (0.2)     & 78.1 (0.7)     & 73.0 (0.7)     & 94.7 (0.3)          & 93.8 (0.5)          & 85.3 (0.9)          \\
$- s_n$             & \textbf{86.1} (0.2) & \textbf{79.6} (0.6) & \textbf{73.7} (0.2) & 94.6 (0.3)          & 93.6 (0.5)          & 84.7 (0.7)          \\
$- s_m$             & 85.3 (0.2)     & 77.9 (0.7)     & 73.1 (0.4)     & 94.9 (0.2)          & 94.0 (0.4)          & 86.5 (0.6)          \\
$- s_n, s_m$ (CLIP Sim.) & 85.1 (0.3) & 77.8 (0.7) & 72.7 (0.6) & 93.8 (0.2)          & 93.0 (0.4)          & 84.5 (0.4)          \\ 
\bottomrule
\end{tabular}

\label{tab:ablation}
}
\end{table}

\begin{table}[!htbp]
\caption{AUROC of label error detection for each component of our score. We find that $d_{mm}$ is the most critical component of the score. Of the two nearest neighbor terms, we find that $s_n$ (nearest image neighbors) is the more important term for most datasets.}
\centering
\resizebox{1.0\textwidth}{!}{
\begin{tabular}{@{}lrrrrrrrr@{}}
\toprule
                      & \texttt{cifar10}    & \texttt{cifar100}   & \texttt{miniImageNet} & \texttt{stanfordCars} & \texttt{flickr30k}  & \texttt{mscoco}     & \texttt{mmimdb}     & \texttt{mimiccxr} \\ \midrule
$d_{mm}$ (CLIP Sim.)           &  92.2 (0.2) & 80.8 (0.1)  & 89.3 (0.2)  & 69.8 (0.4)  & 94.8 (0.5)  & 93.8 (0.2) & 85.1 (0.3)  & 64.1 (0.4)  \\
$s_m$                       & 80.2 (1.1) & 65.4 (2.0) & 80.8 (0.3) & 66.1 (0.6) & 75.9 (3.0) & 75.9 (0.2) & 60.4 (0.7) & 59.7 (0.5) \\
$s_n$                        & 98.1 (0.0) & 88.4 (0.1) & 84.3 (0.2) & 73.0 (0.6) & 69.9 (2.3) & 76.3 (0.8) & 53.7 (0.2) & 57.7 (0.7) \\
$d_{mm} + s_m$                & 92.5 (0.5) & 81.3 (1.1) & 89.6 (0.2) & 70.0 (0.6) & \textbf{95.0} (0.5) & 94.6 (0.3) & \textbf{86.0} (0.2) & 64.5 (0.6) \\
$s_n + s_m$                 & 98.0 (0.2) & 88.8 (0.2) & 84.5 (0.4) & 73.0 (0.7) & 83.2 (0.5) & 86.1 (0.6) & 67.5 (1.0) & 64.7 (1.0) \\
$d_{mm} + s_n$                & \textbf{98.2} (0.1) & 90.7 (0.2) & \textbf{90.0} (0.4) & \textbf{73.1} (0.5) & 94.9 (0.3) & 94.9 (0.2) & 85.3 (0.2) & 64.4 (2.0) \\
$d_{mm} + s_n + s_m$ (\ours) & 98.1 (0.0)  & \textbf{90.8} (0.0) & \textbf{90.0} (0.4)  & \textbf{73.1} (0.5)  & 94.5 (0.2)  & \textbf{95.6} (0.2) & \textbf{86.0} (0.0) & \textbf{70.4} (1.6)  \\ \bottomrule
\end{tabular}

\label{tab:ablation_rebuttal}
}
\end{table}

\begin{table}[!htbp]
\caption{Average ($\sim
$) per-sample runtime (milliseconds) of each method for label error detection. Standard deviation across 3 random data seeds are shown in parentheses. Experiments were run using job scheduling with GTX A6000 or A100 GPUs.}
\centering
\resizebox{1.0\textwidth}{!}{
\begin{tabular}{lrrrrrrrr}
\toprule
 & \texttt{cifar10} & \texttt{cifar100} & \texttt{miniImageNet} & \texttt{stanfordCars} & \texttt{mscoco} & \texttt{flickr30k} & \texttt{mimiccxr} & \texttt{mmimdb} \\
\midrule
\ours & 10.1 (0.5) & 9.6 (0.5) & 7.8 (1.6) & 11.0 (2.0) & 18.8 (1.8) & 35.9 (1.2) & 52.2 (2.7) & 21.1 (1.4) \\
CLIP Sim. & 2.9 (0.0) & 2.9 (0.0) & 6.5 (0.1) & 6.8 (0.1) & 	11.5 (0.1) & 9.8 (0.0) & 17.1 (0.0) & 24.7 (1.0) \\
Deep kNN & 7.0 (1.4) & 6.3 (1.8) & 7.5 (1.8) & 5.4 (0.4) & 50.5 (1.4) & 63.3 (3.4) & 337.6 (13.9) & 29.6 (2.0) \\
Datamap & 19.0 (0.7) & 19.3 (0.5) & 39.2 (1.4) & 41.2 (2.7) & 35.4 (0.9) & 35.2 (1.6) & 45.5 (0.2) & 67.5 (1.2) \\
CaptFilt & - & - & - & - & 13.5 (0.1) & 14.8 (0.0) & 29.6 (0.0) & 35.3  (1.6) \\

VDC & - & - & - & - & 4460 (880) & 5160 (503) & 7932 (3.4) & 4672 (357) \\
\bottomrule
\end{tabular}
}
\label{tab:runtime}
\end{table}

\FloatBarrier

\subsection{Varying Validation Set Size}
\label{app:vary_val_size}

\begin{figure}[htbp!]
\begin{subfigure}{0.43\linewidth}
        \centering  
        \includegraphics[width=1.0\linewidth]{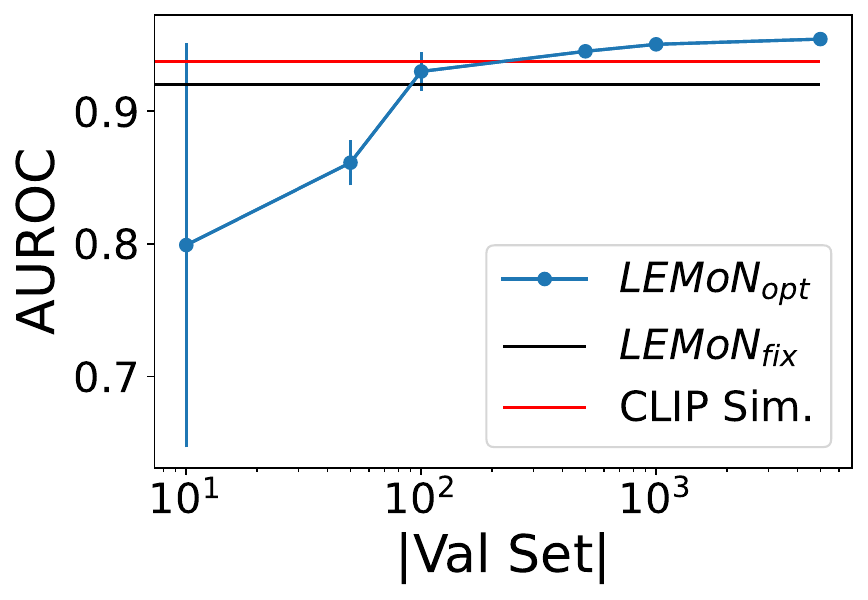}
\caption{\texttt{mscoco}, cat noise}
\end{subfigure}%
\hfill
\begin{subfigure}{0.43\linewidth}
        \centering  
        \includegraphics[width=1.0\linewidth]{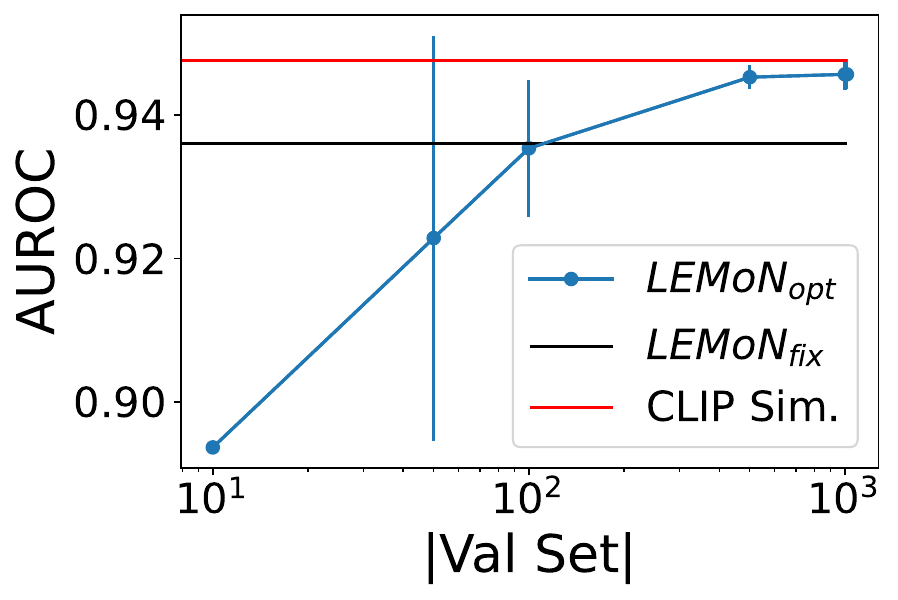}
\caption{\texttt{flickr30k}, cat noise}
\end{subfigure} \\
\begin{subfigure}{0.43\linewidth}
        \centering  
        \includegraphics[width=1.0\linewidth]{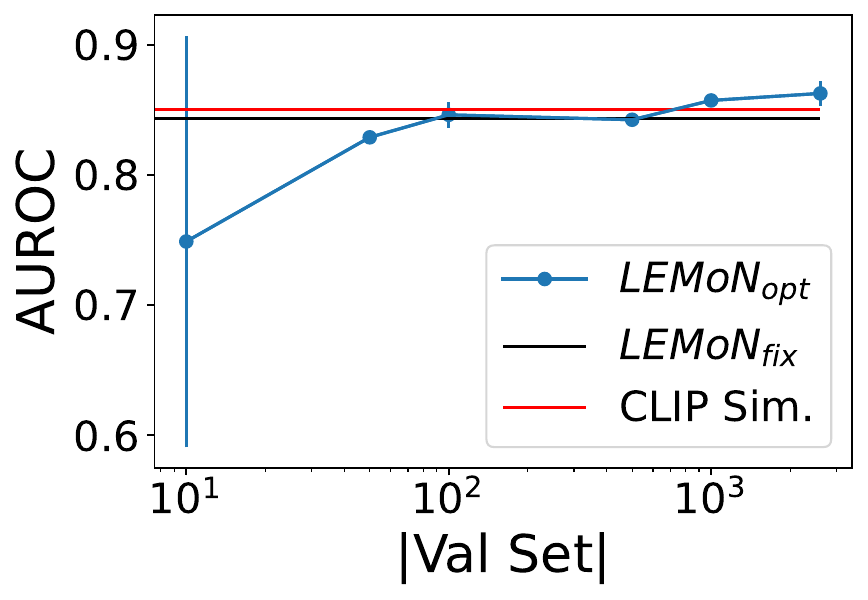}
\caption{\texttt{mmimdb}, noun noise}
\end{subfigure}%
\hfill
\begin{subfigure}{0.43\linewidth}
        \centering  
        \includegraphics[width=1.0\linewidth]{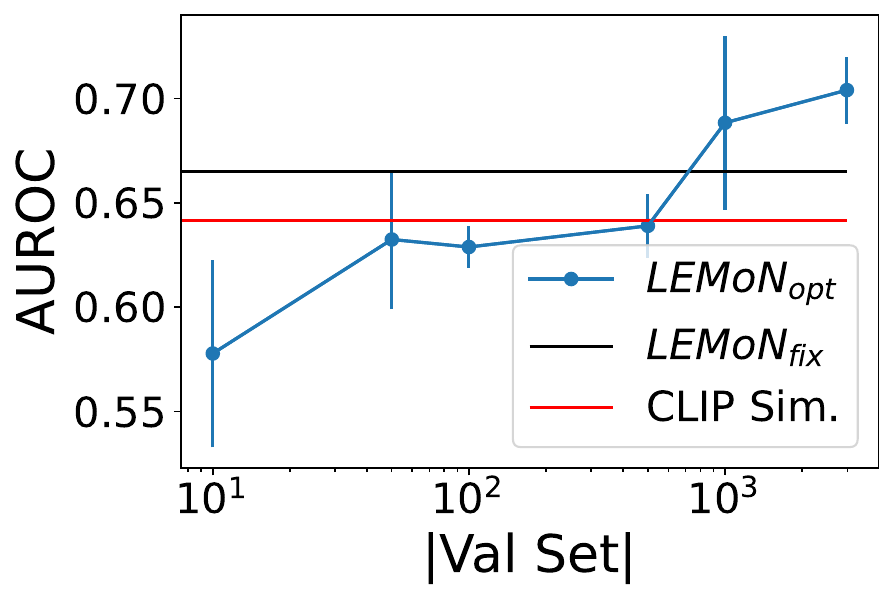}
\caption{\texttt{mimiccxr}, cat noise}
\end{subfigure}%
\caption{Test-set AUROC of mislabel detection with varying size of the labeled validation set for \oursopt. Note that \oursfixed and CLIP Sim. do not have any hyperparameters and as such do not rely on a labeled validation set. \label{fig:vary_val_size}}
\end{figure}

In Figure \ref{fig:vary_val_size}, we examine the effect of varying validation set size (by random subsampling) on \oursopt.

\subsection{Empirical Comparison with \citet{thomas2022emphasizing}}
\label{app:r1_score_table}
In Table \ref{tab:r1_baseline_full}, we compare the performance of \oursopt against the four individual scores and two combined scores proposed in \citet{thomas2022emphasizing}, using the datasets and noise types shown in Table \ref{tab:datasets}. For the $\text{Comb-Val}$ strategy, as there are four terms, we sweep over weights in $\{1, 2, 3, 4, 5\}^4$, selecting the best combination using a labeled validation set, identically to \ours. For the $\text{Comb-Stat}$ strategy, we use the mean and standard deviations, as in Equation (8) in \citet{thomas2022emphasizing}. We find that none of the combined scores significantly outperform $\Upsilon_X^{DIS}$. This is because in both combination strategies, a non-zero weight is placed on the other terms, which essentially adds noise to the final score without contributing any signal.

\begin{table}[htbp!]
\centering
\caption{Comparison of label error detection performance of LEMoN versus baselines from \citet{thomas2022emphasizing}.}
\resizebox{\textwidth}{!}{%
\begin{tabular}{llrrrrrrr}
\toprule
\textbf{Dataset} & \textbf{Metric} 
& $\Upsilon_{X}^{DIS}$ 
& $\Upsilon_{Y}^{DIS}$ 
& $\Upsilon_{X}^{DIV}$ 
& $\Upsilon_{Y}^{DIV}$ 
& \text{Comb-Val} 
& \text{Comb-Stat} 
& \oursopt \\
\midrule
\multirow{3}{*}{\texttt{cifar10}} & AUROC & 77.1 (1.9) & 48.2 (1.2) & 50.3 (3.5) & 45.0 (1.9) & 59.6 (19.5) & 59.6 (19.5) & \textbf{98.1 (0.0)} \\
& AUPRC & 70.4 (2.7) & 41.2 (1.1) & 41.6 (1.6) & 38.9 (2.1) & 51.6 (20.5) & 51.6 (20.5) & \textbf{97.4 (0.1)} \\
& F1 & 66.4 (2.2) & 58.4 (0.8) & 58.4 (0.8) & 58.4 (0.8) & 62.9 (8.0) & 62.9 (8.0) & \textbf{92.0 (0.2)} \\
\midrule
\multirow{3}{*}{\texttt{cifar100}} & AUROC & 66.0 (1.5) & 49.7 (0.9) & 51.1 (1.1) & 49.9 (1.9) & 51.5 (6.2) & 51.8 (7.0) & \textbf{90.8 (0.0)} \\
& AUPRC & 57.4 (2.3) & 40.0 (1.3) & 42.8 (1.6) & 40.9 (1.1) & 41.9 (5.6) & 42.3 (6.3) & \textbf{87.4 (0.3)} \\
& F1 & 58.9 (0.8) & 57.2 (0.3) & 57.3 (0.2) & 57.1 (0.2) & 57.8 (0.7) & 57.8 (0.7) & \textbf{78.4 (0.0)} \\
\midrule
\multirow{3}{*}{\texttt{miniImageNet}} & AUROC & 79.5 (0.3) & 47.4 (0.5) & 64.6 (0.2) & 48.3 (1.1) & 75.4 (0.2) & 76.6 (0.3) & \textbf{90.0 (0.4)} \\
& AUPRC & 65.9 (0.4) & 32.5 (0.0) & 46.3 (0.2) & 33.5 (0.2) & 59.8 (0.2) & 61.7 (0.3) & \textbf{79.7 (3.1)} \\
& F1 & 64.0 (0.1) & 50.9 (0.1) & 53.7 (0.3) & 50.9 (0.1) & 60.5 (0.4) & 61.2 (0.6) & \textbf{76.9 (0.2)} \\
\midrule
\multirow{3}{*}{\texttt{stanfordCars}} & AUROC & 65.7 (0.3) & 50.8 (1.1) & 51.9 (0.9) & 50.1 (0.5) & 62.0 (0.1) & 64.1 (0.1) & \textbf{73.1 (0.5)} \\
& AUPRC & 33.3 (0.6) & 23.3 (0.7) & 23.3 (0.4) & 23.4 (0.2) & 30.1 (0.2) & 31.9 (0.4) & \textbf{40.5 (0.4)} \\
& F1 & 44.3 (0.7) & 38.0 (0.1) & 38.0 (0.5) & 38.2 (0.2) & 41.4 (0.1) & 43.3 (0.2) & \textbf{51.3 (0.5)} \\
\midrule
\multirow{3}{*}{\texttt{flickr30k}} & AUROC & 73.0 (0.6) & 53.3 (1.4) & 49.9 (2.9) & 52.9 (0.2) & 63.9 (0.6) & 70.5 (0.2) & \textbf{94.5 (0.2)} \\
& AUPRC & 59.2 (1.8) & 37.1 (1.8) & 32.8 (1.9) & 37.0 (0.8) & 47.2 (2.1) & 53.7 (2.7) & \textbf{92.8 (0.3)} \\
& F1 & 59.0 (0.3) & 50.9 (0.3) & 50.8 (0.4) & 50.8 (0.3) & 59.4 (3.7) & 62.6 (3.3) & \textbf{83.6 (1.4)} \\
\midrule
\multirow{3}{*}{\texttt{mimiccxr}} & AUROC & 60.0 (0.7) & 49.6 (0.4) & 49.7 (1.1) & 49.1 (1.3) & 51.6 (3.1) & 52.5 (4.7) & \textbf{70.4 (1.6)} \\
& AUPRC & 50.2 (0.5) & 39.3 (0.5) & 39.8 (0.1) & 39.4 (1.0) & 40.5 (2.7) & 42.2 (4.8) & \textbf{60.4 (1.6)} \\
& F1 & 57.2 (0.1) & 57.0 (0.0) & 57.0 (0.1) & 57.0 (0.1) & 57.2 (0.1) & 57.3 (0.1) & \textbf{61.1 (0.8)} \\
\midrule
\multirow{3}{*}{\texttt{mmimdb}} & AUROC & 57.8 (0.4) & 50.1 (0.4) & 48.6 (0.4) & 50.4 (0.3) & 53.4 (2.0) & 54.7 (2.6) & \textbf{86.0 (0.1)} \\
& AUPRC & 46.1 (0.9) & 40.2 (0.6) & 38.9 (0.5) & 40.2 (0.5) & 41.0 (3.1) & 42.0 (3.6) & \textbf{79.4 (0.6)} \\
& F1 & 57.4 (0.2) & 57.1 (0.0) & 57.1 (0.0) & 57.1 (0.0) & 56.5 (1.8) & 56.6 (2.0) & \textbf{73.5 (0.3)} \\
\midrule
\multirow{3}{*}{\texttt{mscoco}} & AUROC & 72.7 (0.3) & 48.5 (0.8) & 52.9 (0.8) & 48.8 (0.2) & 49.7 (0.5) & 50.1 (0.7) & \textbf{95.6 (0.2)} \\
& AUPRC & 67.2 (0.4) & 39.1 (0.5) & 42.3 (1.0) & 39.4 (0.0) & 38.9 (0.3) & 39.5 (0.2) & \textbf{94.6 (0.3)} \\
& F1 & 62.5 (0.3) & 57.0 (0.2) & 57.1 (0.0) & 57.0 (0.2) & 57.3 (0.1) & 57.3 (0.1) & \textbf{87.0 (0.2)} \\
\bottomrule
\end{tabular}
}
\label{tab:r1_baseline_full}
\end{table}

\subsection{Real-World Web Scale Corpus (CC3M)}
\label{app:cc3m_clip}

We conduct an experiment of \oursfixed on CC3M~\cite{changpinyo2021conceptual}, a large web-scraped dataset of images and annotations, where we demonstrate the utility of \ours filtered data on CLIP pretraining. We download CC3M, which contains 2.9 million valid URLs to image-caption pairs. We then pretrain a CLIP model (ViT-B/16) from scratch on this dataset for 20 epochs, with a batch size of 128, and using a cyclic learning rate scheduler with a learning rate of $10^{-4}$.

We then use this CLIP model as the basis to compute distances for \oursfixed, using the reasonable hyperparameters from the main paper: $k = 30$, cosine distance, $\tau_{1, n} = \tau_{1, m} = 0.1$, and $\tau_{2, n} = \tau_{2, m} = 5$. We then select the 1 million samples with the lowest mislabel scores, filtering out the 1.9 million samples most suspected to be mislabels. We pre-train another CLIP model from scratch on this subset using the same architecture and setup as above. We evaluate the resulting model on zero-shot classification using the VTAB benchmark~\cite{zhai2019visual}, and compare it with CLIP models trained using data filtered to 1 million examples using the CLIP similarity baseline, and the original unfiltered model.

In Table \ref{tab:vtab}, we find that \oursfixed marginally outperforms the CLIP similarity baseline on average zero-shot accuracy, though both underperform pretraining on the full corpus. Similar results can be found for few-shot linear probing (Table~\ref{tab:vtab_few}) and full finetuning (Table~\ref{tab:vtab_full}). One likely explanation of this is that although a large proportion of images in the CC3M dataset are technically ``mislabelled'' in that the caption is not a precisely correct description of the image, some substrings of these noisy captions may, on aggregate, contain useful word associations which the model learns, and thus may be useful to downstream tasks. 

\begin{figure}
    \centering
    \includegraphics[width=0.9\linewidth]{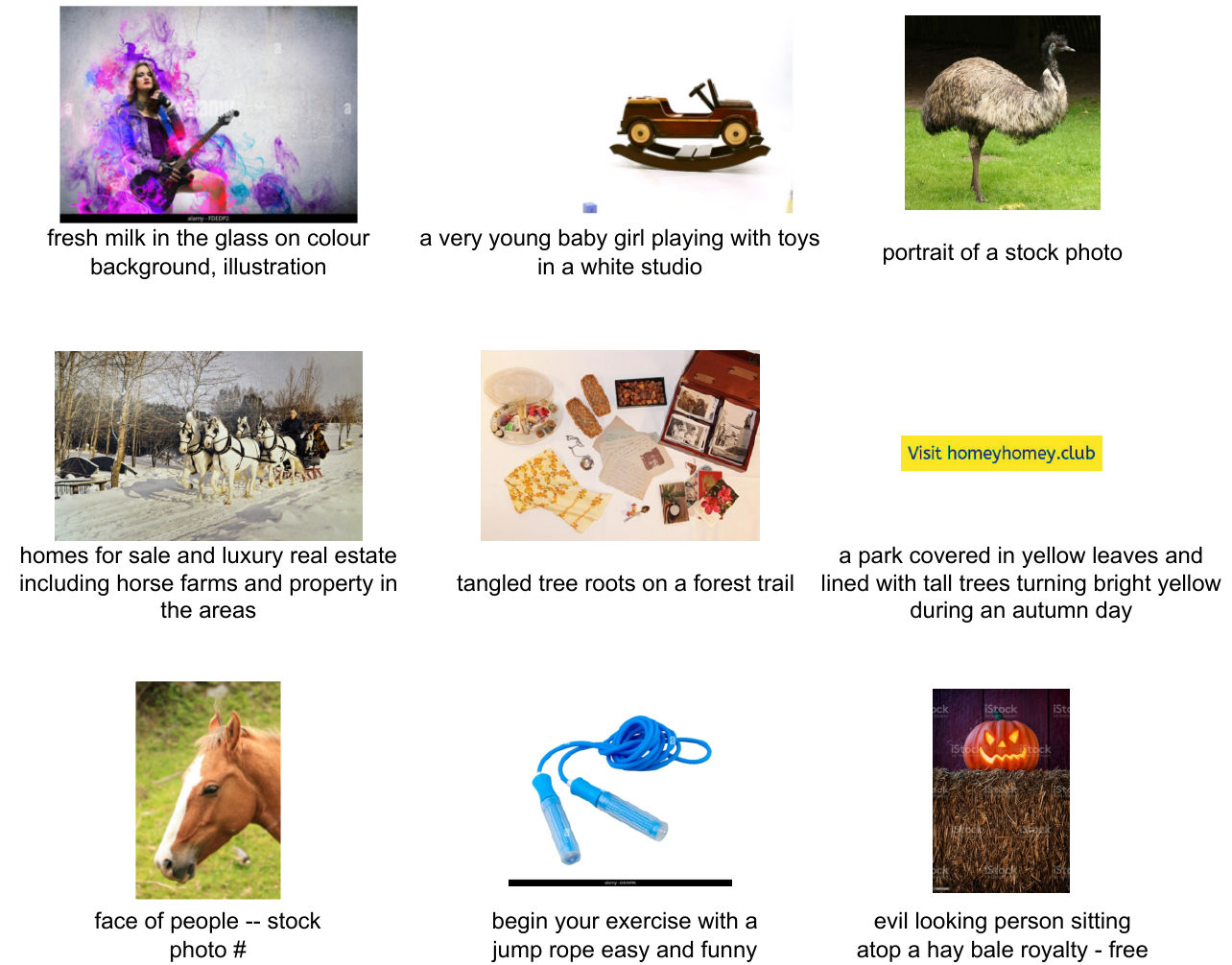}
    \caption{Sample images and captions from CC3M which have been identified as mislabeled by \oursfixed.}
    \label{fig:cc3m_ex}
\end{figure}

We examine images of images selected to be mislabels by our method in Figure \ref{fig:cc3m_ex}. We find that our method identifies images that are completely mislabeled -- one cause of which is images changing after they have been indexed. In addition, our method also identifies samples which are ambiguous or imprecise.

\subsection{Real-World Web Scale Corpus (Datacomp)}
\label{app:datacomp}
We conduct an experiment of \oursfixed on Datacomp \cite{gadre2024datacomp}. We use the small dataset from the filtering track, which originally consisted of 12.8M images. As these images are accessed directly from the web, only 9.96M images were able to be downloaded as of 2024/11/14. We apply \oursfixed to this dataset using OpenAI CLIP ViT-L/14 embeddings provided by Datacomp. We select the 3.5M images with lowest mislabel scores, and use the default hyperparameters from Datacomp to train a CLIP model, and evaluate it on the same 38 zero-shot classification datasets. We compare with filtering using only the CLIP score (equivalent to CLIP Sim.) to the same number of images. In Table \ref{tab:datacomp}, we find that given the available images, \oursfixed outperforms the baseline on average, and on three of four individual evaluations. However, neither method outperforms the scores reported in the original paper due to their dataset being larger.

\begin{table}[]
\caption{Performance of each method on the Datacomp \cite{gadre2024datacomp} small benchmark from the filtering track. As of 2024/11/14, only 9.96M images (``Data Available'') out of 12.8M are accessible. We compare the performance of \oursopt versus the CLIP score baseline after filtering to 3.5M images. } 
\centering
\resizebox{1.0\textwidth}{!}{ 
\begin{tabular}{@{}llrrrrr@{}}
\toprule
                                                                                  & Method                   & \multicolumn{1}{l}{ImageNet} & \multicolumn{1}{l}{ImageNet Dist. Shifts} & \multicolumn{1}{l}{VTAB} & \multicolumn{1}{l}{Retrieval} & \multicolumn{1}{l}{Avg (38 datasets)} \\ \midrule
\multirow{2}{*}{\begin{tabular}[c]{@{}l@{}}Data Available\\ (9.96M Samples)\end{tabular}} & \oursfixed                    & \textbf{0.045}               & \textbf{0.053}                            & \textbf{0.188}           & 0.116                & \textbf{0.168}                        \\
                                                                                  & CLIP score               & 0.043                        & 0.049                                    & 0.177                    & \textbf{0.119}                         & 0.160                                 \\ \midrule
\multirow{7}{*}{\begin{tabular}[c]{@{}l@{}}From \citet{gadre2024datacomp}\\ (12.8M Samples)\end{tabular}}     & No filtering             & 0.025                        & 0.033                                     & 0.145                    & 0.114                         & 0.132                                 \\
                                                                                  & Basic filtering          & 0.038                        & 0.043                                     & 0.150                    & 0.118                         & 0.142                                 \\
                                                                                  & Text-based               & 0.046                        & 0.052                                     & 0.169                    & \textbf{0.125}                         & 0.157                                 \\
                                                                                  & Image-based              & 0.043                        & 0.047                                     & 0.178                    & 0.121                         & 0.159                                 \\
                                                                                  & LAION-2B filtering       & 0.031                        & 0.040                                     & 0.136                    & 0.092                         & 0.133                                 \\
                                                                                  & CLIP score               & \textbf{0.051}                        & \textbf{0.055}                                     & \textbf{0.190}                    & 0.119                         & \textbf{0.173}                                 \\
                                                                                  & Image-based + CLIP score & 0.039                        & 0.045                                     & 0.162                    & 0.094                         & 0.144                                 \\ \bottomrule
\end{tabular}
}
\label{tab:datacomp}
\end{table}

\begin{table}[htbp!]
\centering
\caption{{Zero-shot accuracy (\%) of various CLIP models on the VTAB benchmark \cite{zhai2019visual}. CLIP models (ViT-B/16) are pretrained from scratch on a subset of CC3M~\cite{changpinyo2021conceptual} which has been filtered to 1 million samples using \oursfixed and the CLIP similarity baseline, using a version of CLIP pretrained on the entire dataset. }}
\begin{tabular}{lrr|r}
\toprule
\textbf{}                        & \textbf{CLIP Sim.} & \textbf{\oursfixed}               & \textbf{Unfiltered} \\ \midrule
caltech101                       & 28.25              & \textbf{28.99} & 51.43               \\
cifar100                         & \textbf{11.02}     & 6.79           & 18.65               \\
clevr\_closest\_object\_distance & 18.11              & \textbf{22.58} & 25.76               \\
clevr\_count\_all                & \textbf{12.98}     & 12.65          & 12.05               \\
dmlab                            & 14.78              & \textbf{16.22} & 16.62               \\
dsprites\_label\_orientation     & \textbf{2.44}      & 1.34           & 1.98                \\
dsprites\_label\_x\_position     & 3.06               & \textbf{3.20}  & 3.13                \\
dsprites\_label\_y\_position     & \textbf{3.11}      & 2.89           & 3.20                \\
dtd                              & \textbf{6.60}      & 3.94          & 12.34               \\
eurosat                          & 14.37              & \textbf{22.07} & 9.93                \\
flowers                          & \textbf{6.11}      & 5.19         & 6.83                \\
food101                          & 4.94               & \textbf{5.31}  & 9.02                \\
pets                             & \textbf{7.63}      & 4.69           & 8.23                \\
sun397                           & 13.89              & \textbf{14.22} & 24.02               \\
svhn                             & 7.80               & \textbf{12.35} & 8.00                \\ \midrule
\textbf{Average}                 & 10.34              & \textbf{10.83}                      & 14.08               \\ \bottomrule
\end{tabular}
\label{tab:vtab}
\end{table}

\begin{table*}[htbp!]
\centering
\caption{{Few-shot linear-probing accuracy (\%) using 5 samples per class of various CLIP models on the VTAB benchmark. }}
\begin{tabular}{lrr|r}
\toprule
\textbf{}                        & \textbf{CLIP Sim.}         & \textbf{\oursfixed}         & \textbf{Unfiltered} \\ \midrule
caltech101                       & \textbf{53.50}             & 53.45                       & 60.82               \\
cifar100                         & \textbf{18.66}             & 17.11                       & 23.59               \\
clevr\_closest\_object\_distance & 23.39                      & \textbf{25.48}              & 25.97               \\
clevr\_count\_all                & 19.19                      & \textbf{20.43}              & 23.22               \\
dmlab                            & 18.43                      & \textbf{21.01}              & 19.72               \\
dsprites\_label\_orientation     & \textbf{9.11}              & 8.40                        & 9.91                \\
dsprites\_label\_x\_position     & 3.78                       & \textbf{4.52}               & 4.20                \\
dsprites\_label\_y\_position     & 6.93                       & \textbf{7.90}               & 8.20                \\
dtd                              & 28.88                      & \textbf{29.84}              & 38.78               \\
eurosat                          & \textbf{66.17}             & 65.20                       & 64.76               \\
flowers                          & \textbf{57.31}             & 54.53                       & 62.35               \\
food101                          & 11.94                      & \textbf{13.50}              & 17.89               \\
pets                             & \textbf{18.92}             & 17.12                       & 24.42               \\
sun397                           & 21.43                      & \textbf{21.60}              & 32.53               \\
svhn                             & 12.05                      & \textbf{12.60}              & 12.99               \\ \midrule
\textbf{Average}                 & 24.65                      & \textbf{24.85}              & 28.62               \\ 
\bottomrule
\end{tabular}
\label{tab:vtab_few}
\end{table*}

\begin{table*}[htbp!]
\centering
\caption{{Full finetuning linear-probing accuracy (\%) using 5 samples per class of various CLIP models on the VTAB benchmark. }}
\begin{tabular}{lrr|r}
\toprule
\textbf{}                        & \textbf{CLIP Sim.}         & \textbf{\oursfixed}         & \textbf{Unfiltered} \\ 
\midrule
caltech101                       & 65.91                      & \textbf{65.94}              & 74.80               \\
cifar100                         & \textbf{49.74}             & 48.53                       & 56.54               \\
clevr\_closest\_object\_distance & 52.35                      & \textbf{55.83}              & 53.74               \\
clevr\_count\_all                & \textbf{55.03}             & 54.05                       & 57.99               \\
dmlab                            & 39.10                      & \textbf{39.31}              & 43.00               \\
dsprites\_label\_orientation     & 38.58                      & \textbf{38.65}              & 42.75               \\
dsprites\_label\_x\_position     & 27.08                      & \textbf{36.61}              & 32.78               \\
dsprites\_label\_y\_position     & 46.31                      & \textbf{47.63}              & 49.56               \\
dtd                              & \textbf{44.20}             & 42.07                       & 51.91               \\
eurosat                          & \textbf{92.70}             & 92.09                       & 93.00               \\
flowers                          & \textbf{65.72}             & 63.65                       & 72.45               \\
food101                          & 47.75                      & \textbf{48.62}              & 55.23               \\
pets                             & \textbf{40.86}             & 40.69                       & 51.57               \\
sun397                           & \textbf{49.58}             & 49.45                       & 57.84               \\
svhn                             & 38.08                      & \textbf{39.25}              & 42.74               \\ 
\midrule
\textbf{Average}                 & 50.20                      & \textbf{50.83}              & 55.73               \\ 
\bottomrule
\end{tabular}
\label{tab:vtab_full}
\end{table*}

\subsection{Hyperparameters Used for Real-World}
\label{sec:opt_hparams_real_app}

We show the hyperparameters used for the real-world experiment in Table \ref{tab:real_world_hparams}. We use $k=30$, cosine distance, and these hyperparameters, which originate from a hyperparameter search on synthetically noised data.  We note that \texttt{flickr30k} has some negative hyperparameters, which we attribute to overfitting to a relatively small validation set during hyperparameter selection.

\begin{table}[htbp!]
\caption{Hyperparameters used for the real-world experiment. We use $k=30$, cosine distance, and the hyperparameters below, which originate from a hyperparameter search on synthetically noised data.}
\centering
\begin{tabular}{@{}lrrrrrr@{}}
\toprule
          & $\beta$  &  $\gamma$  &  $\tau_{1,n}$  & $\tau_{2,n}$ &  $\tau_{1,m}$ &  $\tau_{2,m}$ \\ \midrule
\texttt{cifar10}   & 20    & 10     & 0      & 5      & 0      & 5      \\
\texttt{cifar100}  & 15    & 0      & 0      & 5      & 0      & 0      \\
\texttt{mscoco}    & 5.324 & 11.057 & 5.143  & 10.498 & 7.233  & 15.637 \\
\texttt{mmimdb}    & 15    & 5      & 5      & 10     & 5      & 10     \\
\texttt{flickr30k} & 0.092 & -0.177 & -0.274 & -0.074 & -0.072 & 0.000  \\
\texttt{mimiccxr}  & 5     & 10     & 5      & 10     & 5      & 10     \\ \bottomrule
\end{tabular} \label{tab:real_world_hparams}
\end{table}

\subsection{Examples of Detected Real Label Errors}
\begin{figure}
    \centering
    \includegraphics[width=1.0\textwidth]{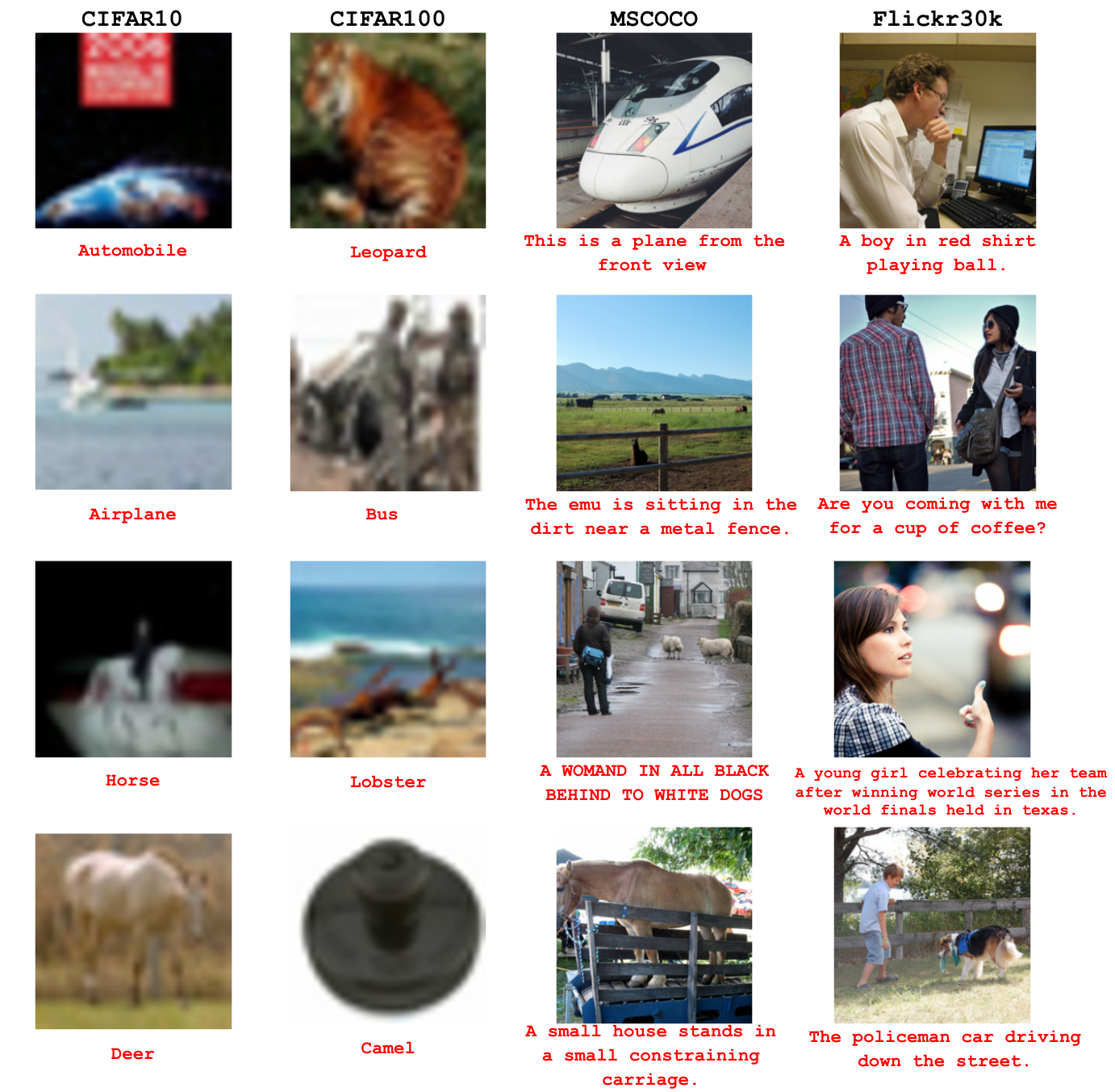}
    \caption{Example images in each dataset identified by our method to be mislabels, and labeled as errors by a human annotator.}
    \label{fig:teaser_full}
\end{figure}

We show additional examples of label errors in Figure~\ref{fig:teaser_full}.

\FloatBarrier

\subsection{Comparison with Northcutt et al., 2021 \cite{northcutt2021pervasive}}
\label{app:cl_real_world}

\begin{figure}[!htbp]
    \centering
    \includegraphics[width=0.9\linewidth]{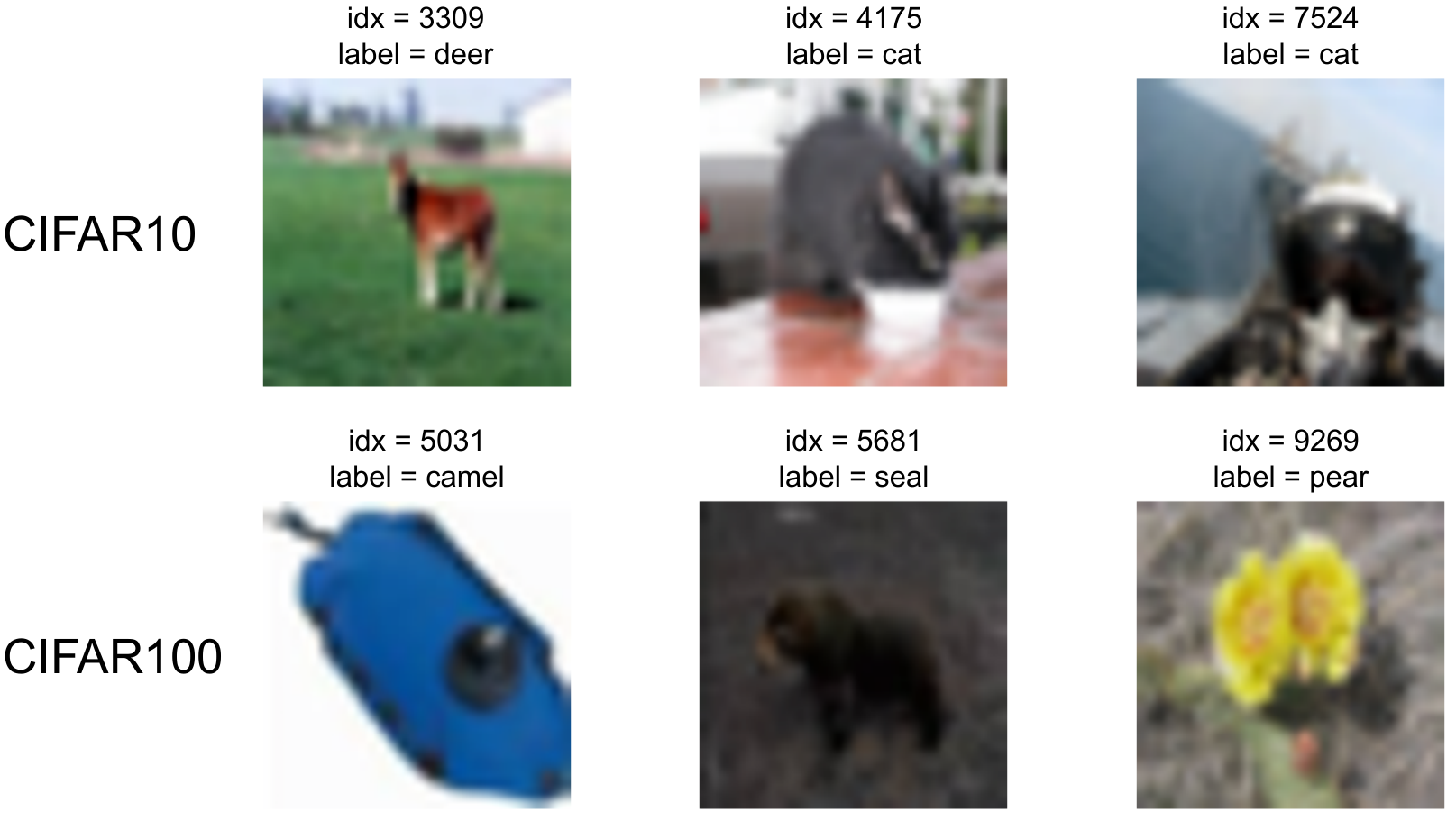}
    \caption{Demonstrative examples of mislabeled samples in \texttt{cifar10} and \texttt{cifar100} which have been identified by our method in the top |CL Set|, but was not identified by confident learning in Northcutt et al., 2021 \cite{northcutt2021pervasive} and thus was not a part of their error set. }
    \label{fig:cl_missed}
\end{figure}

In Northcutt et al., 2021 \cite{northcutt2021pervasive}, the authors utilized confident learning~\cite{northcutt2021confident} to identify suspected errors in the test sets of \texttt{cifar10} and \texttt{cifar100}. They then obtained 5 human labels for each suspected error using Amazon Mechanical Turk, and confirmed the image to be a mislabel if at least 3 of 5 workers stated so. This amounts to 54 confirmed mislabels in \texttt{cifar10} (out of 221 suspected), and 585 confirmed mislabels in \texttt{cifar100} (out of 1650 suspected). In this section, we compare the performance of \oursfixed versus the CLIP similarity baseline on this set. As this set is a subset of the images identified to be mislabels by confident learning, we are not able to compare our model performance with confident learning itself. In addition, this presents a pessimistic view (lower bound) of the performance of our method, as there are many images identified by \ours which \textit{are} mislabeled, but were not selected by confident learning in \cite{northcutt2021pervasive}. We demonstrate examples of these images in Figure \ref{fig:cl_missed}.

In Table \ref{tab:compare_with_cl}, we compare the performance of \oursfixed with the CLIP similarity baseline on the error set from  Northcutt et al., 2021 \cite{northcutt2021pervasive}. First, we compute the mean ranking of all error set samples as ranked by each method, out of 10,000 test-set samples. We find that our method ranks error set samples higher on average than the baseline, though the variance is large. Next, we subset to the top |CL Set| ranked samples for each method, and compute the percentage of which are actually in the error set. We note that this precision metric is upper bounded by the precision of the reference method (confident learning). Again, we find that \oursfixed outperforms the baseline, and is able to identify more actual label errors than CLIP similarity at this threshold.

\begin{table}[!htbp]
\caption{Comparison of \oursfixed (Ours) with the CLIP similarity baseline on the human labeled error set from Northcutt et al., 2021 \cite{northcutt2021pervasive}. In this prior work, the authors used confident learning to identify $|$CL Set$|$ candidate label errors in \texttt{cifar10} and \texttt{cifar100}, $|$Error Set$|$ of which are confirmed to be mislabels by Mechanical Turkers. Mean Ranking denotes the average ranking of all error set samples as ranked by each method. Precision @ Top $|$CL Set$|$ involves taking the top $|$CL Set$|$ samples as ranked by each method, and computing the percentage of which are in the error set. Note that each dataset's test set consists of 10,000 samples. Numbers in parentheses represent one standard deviation.}
\resizebox{1.0\textwidth}{!}{  
\begin{tabular}{@{}lrrrrrrr@{}}
\toprule
                  & \multicolumn{1}{l}{}   & \multicolumn{1}{l}{}       & \multicolumn{2}{c}{\textbf{Mean Ranking}} & \multicolumn{3}{c}{\textbf{Precision @ Top $|$CL Set$|$}} \\ \cmidrule(l){4-8} 
\textbf{Dataset}  & \textbf{$|$CL Set$|$} & \textbf{$|$Error Set$|$} & \textbf{\oursfixed}    & \textbf{CLIP Sim.}    & \textbf{Oracle}   & \textbf{\oursfixed}   & \textbf{CLIP Sim.}  \\ \midrule
\texttt{cifar10}  & 275                    & 54                         & 1269.7 (1905.1)   & 2681.0 (2507.1)  & 19.64\%           & 6.55\%                & 1.45\%         \\
\texttt{cifar100} & 2235                   & 585                        & 2357.5 (1981.5)   & 3642.1 (2719.5)  & 26.17\%           & 14.41\%               & 10.16\%        \\ \bottomrule
\end{tabular}
}
\label{tab:compare_with_cl}
\end{table}

\subsection{Downstream Classification with Label Error Detection-based Filtering}
\label{app:downstream_clf_acc}
Here, we show the impact of filtering out different proportions of the training data based on label error predictions, and obtaining test performance. 
\subsubsection{Average accuracy}
\begin{figure}[!htbp]
    \centering
    \includegraphics[width=0.8\linewidth]{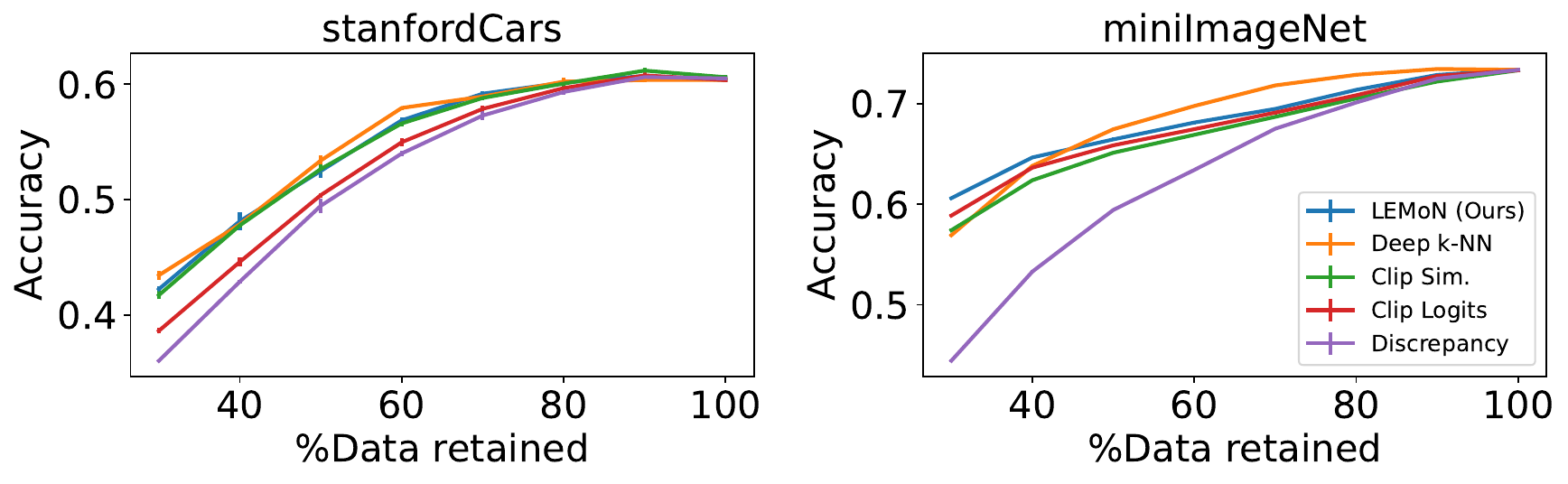}
    \caption{Downstream accuracy on \texttt{stanfordCars},
    and \texttt{miniImageNet}.}
    \label{fig:downstream_classification_all}
\end{figure}

We observe that the gap in performance is low between \oursopt and the best method (less than 0.5\% on \texttt{miniImageNet} and \texttt{stanfordCars}) in terms of downstream accuracy (see Figure~\ref{fig:downstream_classification_all}).

\subsection{Area Under Test Error vs \% Data Retained Curve}

We compute the area under the test error (i.e., 1-accuracy) vs \% data retained curve in Table~\ref{tab:auc_downstream_filter_clf}. Note that the minimum data retained is 30\% (i.e., the minimum amount of data required for training the downstream model). 

On \texttt{cifar10}, we observe that \ours performs the best in terms of AUC (i.e., lowest test error). On \texttt{stanfordCars} and \texttt{miniImageNet}, Deep k-NN performs better. However, the gap in performance is low between \oursopt and the best method on all datasets.

\begin{table}[!htp]\centering
\caption{Area under the curve: test error vs \% data retained for all four classification datasets. Lower is better, and bold denotes best method.}
\scriptsize
\begin{tabular}{lrrrrr}
\toprule
\textbf{Method}& \texttt{cifar10} &\texttt{cifar100} &\texttt{stanfordCars}&\texttt{miniImageNet} \\
\midrule
CLIP Sim. &4.58 &\textbf{16.89} &31.18 &22.88 \\
CLIP Logits &4.54 &16.90 &32.22 &22.42 \\
Discrepancy &5.98 &20.22 &32.81 &25.48 \\
Deep k-NN &4.37 &17.54 & \textbf{30.94} &\textbf{21.57} \\
Ours &\textbf{4.23} & 17.02 &31.12 &22.01 \\
\bottomrule
\end{tabular}
\label{tab:auc_downstream_filter_clf}
\end{table}

\subsection{Out-of-Domain Robustness}
\label{ref:ood_robust}
We report the test performance on an Out-of-Domain (OOD) dataset CIFAR-10C~\cite{hendrycks2018benchmarking}, when models are trained and validated on the \texttt{cifar10} noisy train set. The CIFAR-10C dataset contains 19 corruptions applied to the \texttt{cifar10} test set, with varying severity of corruption. Then, robustness is measured as the average test top-1 class accuracy performance on the CIFAR-10C dataset (across all corruption types and severities), following prior work~\cite{diffenderfer2021winning}. We observe that our method outperforms the CLIP similarity baseline on robustness, when percentage of data retained is less than 60\%. 

\begin{figure}[!htbp]
    \centering
    \includegraphics[width=0.4\linewidth]{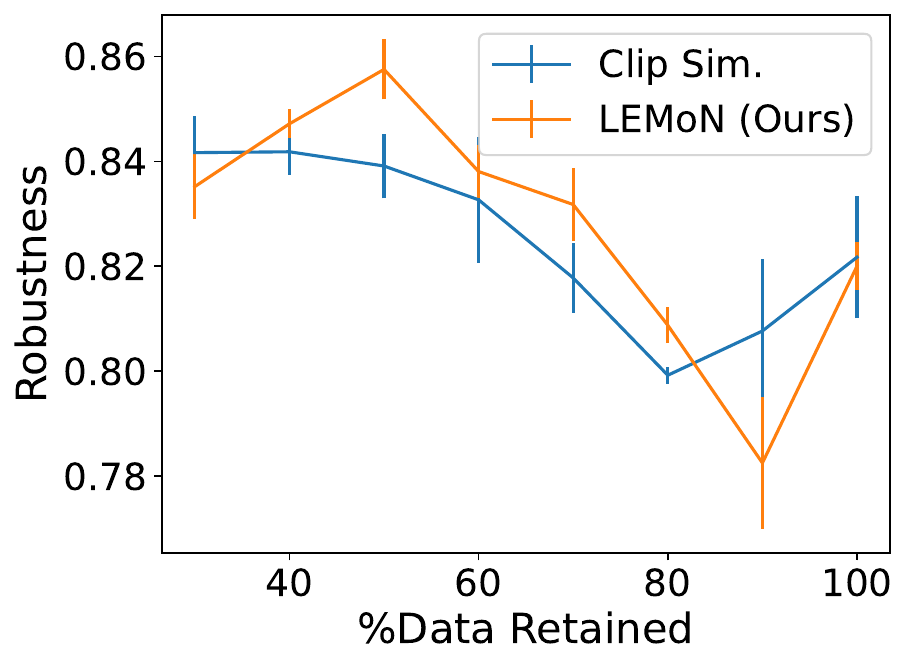}
    \caption{Downstream accuracy on CIFAR-10C, averaged across all corruption types.}
    \label{fig:robustness_cifar10c}
\end{figure}

\end{document}